\documentclass[11pt]{article}

\usepackage[final]{acl}

\usepackage{times}
\usepackage{latexsym}
\usepackage{xcolor}
\usepackage{colortbl}
\usepackage{subcaption}
\usepackage{CJKutf8}
\usepackage{url}

\usepackage[utf8]{inputenc}

\usepackage[T2A, T1]{fontenc}
\usepackage[main=english, russian]{babel}

\usepackage{microtype}
\usepackage{url}
\usepackage{multirow}

\usepackage{times}
\usepackage{latexsym}
\usepackage{enumitem}
\usepackage{listings}
\usepackage{url}
\usepackage[T1]{fontenc}
\usepackage[utf8]{inputenc}

\usepackage{microtype}

\usepackage{inconsolata}

\usepackage{longtable}
\usepackage{graphicx}

\usepackage{caption}
\usepackage{subcaption}

\usepackage{colortbl}
\usepackage{makecell}
\usepackage{multirow}
\usepackage{supertabular}
\usepackage{placeins}

\usepackage{subfiles}
\usepackage{enumitem}

\usepackage{algorithm}
\usepackage{algpseudocode}
\usepackage{amsmath}

\usepackage{cleveref}

\setlist[itemize]{leftmargin=1em}  % smaller left margin = more text on one line

\usepackage{booktabs}

\usepackage{tcolorbox}
\newcounter{promptno}[section]
\newlength\mystoreparindent
\newenvironment{prompt}[1][]
{
  \setlength{\mystoreparindent}{\the\parindent}
  \setlength{\parindent}{0pt}
  \refstepcounter{promptno}
  \par\medskip
  \noindent
  \begin{tcolorbox}[left=1pt,right=1pt]
  \textsc{{template \small\thesubsection.\thepromptno}}\\
  \small
  \tt
}{
  \end{tcolorbox}
  \setlength{\parindent}{\mystoreparindent}
  \medskip
}

\newcounter{utterance}
\usepackage{amssymb}

\DeclareUnicodeCharacter{25A2}{$\square$}
\usepackage{color}

\newcolumntype{T}{>{\ttfamily}l}

\tcbuselibrary{skins}

\definecolor{a-colour}{RGB}{135,74,175}
\definecolor{b-colour}{RGB}{243,145,9}
\definecolor{gm-colour}{RGB}{128,128,128}
\definecolor{wordlegreen}{rgb}{0.4,0.79,0.17}
\definecolor{wordlered}{rgb}{0.86,0.31,0.29}
\definecolor{wordleyellow}{rgb}{1,1,0}

\def \dcolwidth {3cm}
\def \skipcolwidth {0.8cm}
\def \arccorner {3mm}
\def \disttocorner {6pt}

\newtcolorbox[use counter=nbubbles]{a-gm}[1]{
    colback=a-colour!10!white,
    colframe=a-colour,
    fonttitle=\bfseries\tiny,
    fontupper=\footnotesize,
    title={#1},
    sharp corners=west,
    arc=\arccorner,
    width=\dcolwidth,
    left skip=0cm,
    top=0pt,
    bottom=0pt,
    left=0pt,
    right=\disttocorner,
    before={\vspace{-0.1cm}},
    boxrule=0.5pt,
    enhanced,
    attach boxed title to top left={yshift=-0.1mm},
    boxed title style={size=small,colback=a-colour},
    overlay unbroken and first = {
    \node[text width=0.4cm,draw=black,line width=0.1mm,align=center,gray] at (-0.5,0.2) {\footnotesize \thetcbcounter};
  }
}

\newtcolorbox[use counter=nbubbles]{gm-a}[1]{
    colback=gm-colour!5!white,
    colframe=gm-colour,
    fonttitle=\bfseries\tiny,
    fontupper=\footnotesize,
    title={#1},
    sharp corners=east,
    arc=\arccorner,
    width=\dcolwidth + \skipcolwidth,
    left skip=\skipcolwidth,
    top=0pt,
    bottom=0pt,
    left=\disttocorner,
    right=0pt,
    before={\vspace{-0.1cm}},
    boxrule=0.5pt,
    halign title=flush right,
    enhanced,
    attach boxed title to top right={yshift=-0.1mm},
    boxed title style={size=small,colback=gm-colour},
    overlay unbroken and first = {
    \node[anchor=north east,text width=0.4cm,draw=black,line width=0.1mm,align=center,gray] at (-0.9,0.55) {\footnotesize\thetcbcounter};}
}

\newtcolorbox[use counter=nbubbles]{b-gm}[1]{
    colback=b-colour!10!white,
    colframe=b-colour,
    fonttitle=\bfseries\tiny,
    fontupper=\footnotesize,
    title={#1},
    sharp corners=east,
    arc=\arccorner,
    width=\dcolwidth + \skipcolwidth + \dcolwidth +  \skipcolwidth,
    left skip=\dcolwidth + \skipcolwidth + \skipcolwidth,
    top=0pt,
    bottom=0pt,
    left=\disttocorner,
    right=0pt,
    before={\vspace{-0.1cm}},
    boxrule=0.5pt,
    halign title=flush right,
    enhanced,
    attach boxed title to top right={yshift=-0.1mm},
    boxed title style={size=small,colback=b-colour},
    overlay unbroken and first = {
    \node[anchor=north east,text width=0.4cm,draw=black,line width=0.1mm,align=center,gray] at (-4.7,0.55) {\footnotesize\thetcbcounter};}
}

\newtcolorbox[use counter=nbubbles]{gm-b}[1]{
    colback=gm-colour!5!white,
    colframe=gm-colour,
    fonttitle=\bfseries\tiny,
    fontupper=\footnotesize,
    title={#1},
    sharp corners=west,
    arc=\arccorner,
    width=\dcolwidth + \skipcolwidth + \dcolwidth,
    left skip=\dcolwidth + \skipcolwidth,
    top=0pt,
    bottom=0pt,
    left=0pt,
    right=\disttocorner,
    before={\vspace{-0.1cm}},
    boxrule=0.5pt,
    enhanced,
    attach boxed title to top left={yshift=-0.1mm},
    boxed title style={size=small,colback=gm-colour},
    overlay unbroken and first = {
    \node[anchor=north east,text width=0.4cm,draw=black,line width=0.1mm,align=center,gray] at (-3.9,0.55) {\footnotesize\thetcbcounter};}
}

\newtcolorbox[use counter=nbubbles]{gm-gm}[1]{
    colback=gm-colour!5!white,
    colframe=gm-colour,
    fonttitle=\bfseries\tiny,
    fontupper=\footnotesize,
    title={#1},
    sharp corners,
    width=\dcolwidth + \dcolwidth + \skipcolwidth + \skipcolwidth,
    leftright skip=\dcolwidth,
    top=0pt,
    bottom=0pt,
    left=0pt,
    right=0pt,
    before={\vspace{-0.1cm}},
    boxrule=0.5pt,
    halign title=center,
    enhanced,
    attach boxed title to top center={yshift=-0.1mm},
    boxed title style={size=small,colback=gm-colour},
    overlay unbroken and first = {
    \node[anchor=north east,text width=0.4cm,draw=black,line width=0.1mm,align=center,gray] at (-3.1,0.55) {\footnotesize\thetcbcounter};}
}

\usepackage{inconsolata}

\usepackage{graphicx}

\title{The Price of Thought: A Multilingual Analysis of Reasoning, Performance, and Cost of Negotiation in Large Language Models}

\author{%
Sherzod Hakimov${^\mathbf{1}}$, 
Roland Bernard${^\mathbf{3}}$, 
Tim Leiber${^\mathbf{1}}$,
Karl Osswald${^\mathbf{1}}$,
Kristina Richert${^\mathbf{1}}$,\\
\textbf{Ruilin Yang}${^\mathbf{1}}$,
\textbf{Raffaella Bernardi}${^\mathbf{3}}$,
\textbf{David Schlangen}${^\mathbf{1,2}}$\\$^{\mathbf{1}}$Computational Linguistics, Department of Linguistics\\
University of Potsdam, Germany\\
$^{\mathbf{2}}$German Research Center for Artificial Intelligence (DFKI), Berlin, Germany\\
$^{\mathbf{3}}$Free University of Bozen-Bolzano, Italy\\
{\texttt{\{firstname.lastname\}@uni-potsdam.de}}, 
{\texttt{\{firstname.lastname\}@unibz.it}}
}

\begin{document}
\maketitle
\begin{abstract}
Negotiation is a fundamental challenge for AI agents, as it requires an ability to reason strategically, model opponents, and balance cooperation with competition. We present the first comprehensive study that systematically evaluates how explicit reasoning training affects the negotiation abilities of both commercial and open-weight large language models, comparing these models to their vanilla counterparts across three languages. Using a self-play setup across three diverse dialogue games, we analyse trade-offs between performance and cost, the language consistency of reasoning processes, and the nature of strategic adaptation exhibited by models.
Our findings show that enabling reasoning---that is, scaling test time compute---significantly improves negotiation outcomes by enhancing collaboration and helping models overcome task complexities, but comes at a substantial computational cost: reasoning improves GPT-5's performance by 31.4 \% while increasing its cost by nearly 400 \%. Most critically, we uncover a significant multilingual reasoning distinction: open-weight models consistently switch to English for their internal reasoning steps, even when negotiating in German or Italian (and thus possibly impacting potential explainability gains through the disclosure of reasoning traces), while a leading commercial model maintains language consistency between reasoning and final output.
\end{abstract}

\section{Introduction}

Negotiation is a key aspect of human social and economic behaviour that poses significant challenges for AI systems. The problem is complex, as agents must produce coherent language while performing strategic reasoning, modelling the opponents' preferences and goals, and managing the tension between cooperation and competition to maximise their own outcomes. 
With LLMs increasingly being deployed as autonomous agents in domains ranging from e-commerce to resource distribution, assessing their negotiation capabilities has become essential. Negotiation requires strategic decision-making, which fundamentally depends on reasoning, and provides a strong test-bed for measuring agent competence through interactive, multi-turn scenarios with objective measures that go beyond static evaluation. We therefore study LLM reasoning abilities in negotiation settings.

Studies show that LLMs often deviate from optimal play with unreliable performance, where even top-performing systems can lose to weaker opponents and fail in cooperative scenarios \cite{DBLP:conf/iclr/DavidsonVK024,DBLP:journals/corr/abs-2411-05990,DBLP:conf/icml/0001CYTJ024,DBLP:conf/emnlp/KwonWKCLG24}. These models can acquire deceptive tactics, e.g., expressing false interest in low-value items for later concessions~\cite{lewis-etal-2017-deal}, express desperation to improve outcomes~\cite{DBLP:conf/icml/0001CYTJ024}, perform much worse as buyers than sellers \cite{DBLP:conf/acl/Xia0RMZ0024}, or even take economic risks that lead to budget violations and overpayment~\cite{DBLP:journals/corr/abs-2506-00073}. Previous studies have also investigated theory-of-mind reasoning in negotiation tasks~\cite{DBLP:conf/emnlp/KwonWKCLG24}, and their performance in interactive multi-agent negotiation games and the effect of Chain-of-Thought reasoning~\cite{ DBLP:conf/emnlp/ChanJYDF0L0WS24, abdelnabi2024llmdeliberation}.

Despite this growing body of research, two critical gaps remain unexplored. First, no previous work has systematically investigated how vanilla LLMs specifically trained to reason perform on negotiation tasks and their effects in terms of both performance and computational cost, a fundamental oversight given that reasoning is central to strategic decision-making. Second, the field has been constrained by monolingual analyses based solely on English, leaving multilingual negotiation capabilities completely unexplored. To address these significant limitations, we conduct the first comprehensive study examining the effects of reasoning on negotiation tasks across three languages: English, German, and Italian. We implement three distinct dialogue games that require bargaining skills, collaboration, and strategic preference management~\cite{VonNeumann+Morgenstern:1944,nash1953two,fisher_getting_2011}, testing both commercial and open-weight LLMs in a self-play mode where both sides of the games are played by instances of the same model. We systematically analyse the impact of turning on and off the reasoning mode of models. We also investigate whether models retain their language consistency in their reasoning traces or if they switch to a dominant language in their training data (mainly English). It is important to study, as the reasoning traces provide essential cues for the explainability of model decisions. We target three fundamental research questions:

\noindent\textbf{RQ1}: \textit{What is the computational and performance trade-off between reasoning overhead and negotiation effectiveness across tasks and languages?}

\noindent\textbf{RQ2}: \textit{To what extent do models maintain language consistency in their reasoning processes when performing multilingual negotiation tasks?}

\noindent\textbf{RQ3}: \textit{Do models demonstrate strategic adaptation over multiple turns, or merely surface-level pattern matching creating an illusion of thinking?}

\section{Related Work}

The convergence of Large Language Models (LLMs) and game-theoretic frameworks represents a significant development in the evaluation and understanding of strategic capabilities in AI systems \cite{sun2025gametheorymeetslarge}, as we explain in the next two parts the state-of-the art methodologies in this field.

\textbf{Benchmarking strategic capabilities in LLMs}: systematic evaluation of LLM negotiation performance reveals substantial variance in strategic behaviour, with models frequently departing from optimal play and exhibiting asymmetric performance patterns~\cite{DBLP:conf/iclr/DavidsonVK024, DBLP:journals/corr/abs-2411-05990,DBLP:journals/corr/abs-2305-16867, pollo2025re}, or even some bigger models underperform against weaker opponents and encounter difficulties in cooperative bargaining scenarios. This has motivated the creation of comprehensive evaluation frameworks and experimental platforms designed to assess agent (initialised as LLMs) behaviour across various negotiation contexts, including resource allocation and pricing tasks \cite{DBLP:conf/icml/0001CYTJ024, DBLP:conf/acl/Xia0RMZ0024}.
Empirical studies reveal that language models can display interesting tactics, e.g. deceptive strategies such as expressing false interest in low-value items to create bargaining leverage in subsequent exchanges \cite{lewis-etal-2017-deal}. LLMs also exhibit susceptibility to cognitive biases, including anchoring effects and responsiveness to social manipulation tactics, where expressions of desperation or aggressive language can substantially affect negotiation outcomes \cite{DBLP:conf/icml/0001CYTJ024}. As these behaviours present practical risks, outcomes may result in budget constraint violations and acceptance of economically disadvantageous agreements~\cite{DBLP:journals/corr/abs-2506-00073}. A consistent finding across studies is the asymmetric difficulty in buyer versus seller roles, with agents systematically underperforming as buyers~\cite{DBLP:conf/acl/Xia0RMZ0024}. Our work focuses on studying the reasoning behind model choices in a multilingual setting \cite{DBLP:journals/corr/abs-2502-09457}.

\textbf{Improving strategic reasoning and decision-making}: Given the inconsistent strategic reasoning observed in current LLMs \cite{wong2025reasoningcapabilitieslargelanguage}, research efforts focus on enhancing decision-making through structured approaches such as integration of game-theoretic solvers with LLM dialogue~\cite{DBLP:journals/corr/abs-2402-01704}, utilising post-hoc CoT prompting~\cite{abdelnabi2024llmdeliberation}, structured reasoning workflows based on Dominant Strategy Search and Backward Induction~\cite{DBLP:journals/corr/abs-2411-05990}, or even learn by interacting based with reinforcement learning~\cite{DBLP:conf/iclr/CaoLLLTC18}. From a prompting perspective, Chain-of-Thought reasoning emerges as a critical factor in agent performance~\cite{DBLP:journals/corr/abs-2305-19165}, contributing to more consistent outcomes in integrative negotiation settings \cite{DBLP:journals/corr/abs-2503-06416}. Hybrid architectural approaches have been developed to constrain agent behaviour, such as employing deterministic offer generation modules to control pricing decisions while utilising LLMs for natural language dialogue production, resulting in substantial improvements in deal completion rates and profit margins \cite{DBLP:conf/acl/Xia0RMZ0024}. The focus of this paper is on evaluating the effect of reasoning on negotiation strategies without adding any additional component by simply using the vanilla LLMs that are trained specifically to exhibit reasoning capabilities.
%We do not focus on enhancing any particular aspect, and simply use the vanilla LLMs with reasoning capability.

\section{Evaluating Negotiation Abilities with Dialogue Games}

In this section, we provide details on how we implement three dialogue games to test negotiation abilities in LLMs. We define a dialogue game as a structured communication between two agents (LLMs) according to a given communication protocol aimed at achieving a defined goal. The goals in games are defined as the expected outcome, e.g., making a deal that agents attempt to maximise. The communication protocols dictate how messages are expected to be formatted. A negotiation task unfolds by agents starting to communicate with one another for the given goal, and the conversation continues until it reaches one of the defined stopping criteria: 1) the goal state is reached, 2) the maximum turn limit is reached, or 3) agents violate the communication protocol and the game-play is aborted. The orchestration of this message passing between agents, validation of communication protocols, and checking whether goal states are reached are done by the Game Master, a scripted entity in the dialogue. We use the \textit{clembench framework}~\cite{chalamalasetti-etal-2023-clembench}. The dialogue games are run in a self-play mode where both participants of the game are LLMs playing against or with one another. Next, we provide game details\footnote{\url{https://github.com/clembench/clembench}}.

\subsection{Deal or No Deal}

The Deal or No Deal (DoND) game is a two-player game~\cite{lewis-etal-2017-deal} designed to simulate a multi-issue bargaining scenario. \Cref{fig:dond_example} shows an example. Two players must communicate to reach a mutually beneficial agreement on a set of items, each holding different values for each player. The game focuses on evaluating negotiation skills, including the ability to express and understand preferences, as well as compromise (see Appendix~\ref{sec:appendix_dond} for prompts and other details).

\paragraph{Game Mechanics}
Two agents negotiate over a shared set of items, each having a different private value for item types. Players exchange free-form messages and then simultaneously make secret proposals using specific syntax. If proposals are compatible (enough items are requested for both), each player receives points based on the value of their requested items, as determined by their private function. If proposals conflict, both receive zero points. The Game Master strictly enforces rules without re-prompting, making adherence to rules critical for a successful play experience.

\begin{figure}[ht!]
  \centering
  {
\scriptsize
\setcounter{utterance}{1}
\setlength{\tabcolsep}{0pt}
\begin{supertabular}{c@{$\;$}|p{.15\linewidth}@{}p{.15\linewidth}p{.15\linewidth}p{.15\linewidth}p{.15\linewidth}p{.15\linewidth}}   
  \# & $\;$\textbf{Player A} & \multicolumn{4}{c}{\textbf{Game Master}} & $\;\:$\textbf{Player B}\\
  \hline

  \theutterance \stepcounter{utterance} & &
  \multicolumn{4}{p{0.6\linewidth}}{
    \cellcolor[rgb]{0.9,0.9,0.9}{
      \makecell[{{p{\linewidth}}}]{
        \tt {\tiny [A$\langle$GM$\rangle$B]}
        <GAME DESCRIPTION> ... \\
        \tt Available items: 1 book, 2 hats, 2 balls.
      }
    }
  } & \\

  \theutterance \stepcounter{utterance} & &
  \multicolumn{4}{p{0.6\linewidth}}{
    \cellcolor[rgb]{0.9,0.9,0.9}{
      \makecell[{{p{\linewidth}}}]{
        \tt {\tiny [A$\langle$GM]}
        \tt Player A value function: book: 0,\\ \tt hat: 1, ball: 4.
      }
    }
  }  \vspace{0.55em} & \\
 
  \theutterance \stepcounter{utterance} & &
  \multicolumn{4}{p{0.6\linewidth}}{
    \cellcolor[rgb]{0.9,0.9,0.9}{
      \makecell[{{p{\linewidth}}}]{
        \tt {\tiny [GM$\rangle$B]}
        \tt Player B value function: book: 6,\\ \tt hat: 1, ball: 1.
      }
    }
  } & \\

  \theutterance \stepcounter{utterance} &
  \multicolumn{4}{p{0.6\linewidth}}{
    \cellcolor[rgb]{0.8,1,0.9}{
      \makecell[{{p{\linewidth}}}]{
        \tt {\tiny [A$\rangle$B]}
        Hi! Here are my preferences: I value balls the most, then hats, and I don't value the book at all. How about you? What do you value most? 
      }
    } 
  } & & \\
 
  \theutterance \stepcounter{utterance} & & &
  \multicolumn{4}{p{0.6\linewidth}}{
    \cellcolor[rgb]{1,0.85,0.72}{
      \makecell[{{p{\linewidth}}}]{
        \tt {\tiny [A$\langle$B]}
        Thanks for sharing! I value the book most, while hats and balls are equally less important to me. Would you take both balls and one hat, and I'll take the book and the other hat? That way, we both get what we value most.
      }
    }
  } \\
 
  \theutterance \stepcounter{utterance} &
  \multicolumn{4}{p{0.6\linewidth}}{
    \cellcolor[rgb]{0.8,1,0.9}{
      \makecell[{{p{\linewidth}}}]{
        \tt {\tiny [A$\rangle$B]}
        That sounds fair! With this split, I'd get both balls and one hat, while you'd get the book and one hat, so we each get what we value most. Ready to submit this as my proposal unless you'd like to tweak anything.
      }
    }
  } & & \\
 
  \theutterance \stepcounter{utterance} & & &
  \multicolumn{4}{p{0.6\linewidth}}{
    \cellcolor[rgb]{1,0.85,0.72}{
      \makecell[{{p{\linewidth}}}]{
        \tt {\tiny [A$\langle$B]}
        I agree with the split. I'll take the book and one hat, you take both balls and the other hat.
      }
    } 
  } \\
 
  \theutterance \stepcounter{utterance} &
  \multicolumn{4}{p{0.6\linewidth}}{
    \cellcolor[rgb]{0.8,1,0.9}{
      \makecell[{{p{\linewidth}}}]{
        \tt {\tiny [A$\rangle$GM]}
        [Proposal: 1 hat, 2 balls] 
      }
    }
  } & & \\
 
  \theutterance \stepcounter{utterance} & &
  \multicolumn{4}{p{0.6\linewidth}}{
    \cellcolor[rgb]{0.9,0.9,0.9}{
      \makecell[{{p{\linewidth}}}]{
        \tt {\tiny [GM$\rangle$B]}  
        The other player has submitted a proposal. You must now submit your proposal.
      }
    }
  } & \\

  \theutterance \stepcounter{utterance} & & &
  \multicolumn{4}{p{0.6\linewidth}}{
    \cellcolor[rgb]{1,0.85,0.72}{
      \makecell[{{p{\linewidth}}}]{
        \tt {\tiny [GM$\langle$B]}  
        [Proposal: 1 book, 1 hat] 
      }
    }
  } \\
\end{supertabular}
}

  \caption{An example of a \textit{Deal or No Deal} episode that ends in a successful Pareto optimal agreement.}
  \label{fig:dond_example}
\end{figure}

\paragraph{Game Instances}
We generate instances for the game programmatically by randomly sampling 100 common nouns. Each instance was generated with a maximum turn limit of 5, and each game instance features a random set of between 3 and 5 different item types. We have two different sets of experiments. In the \textbf{semi-competitive} experiment, players are asked to maximise their own score (to balance cooperation, avoid zero-point conflicts). For the \textbf{cooperative} experiment, players are asked to maximise the sum of the scores of both players. This transforms the game into a purely cooperative task, in which two rational agents should consistently achieve the maximum score, assuming they can communicate and reason about the preferences perfectly. Each experiment includes 20 instances, with \textbf{total 40 instances}.

\subsection{Clean Up}

Clean Up is a two-player game focused on cooperative strategy development and object rearrangement. Object rearrangement has been a long-standing problem that requires modelling the given situation, the action space, and spatial reasoning \cite{DBLP:journals/corr/abs-2011-01975,DBLP:journals/ral/ZengWYZDCD24,DBLP:journals/access/KhanQCASR25}. Each player can only access its own grid and need to negotiate which items to rearrange so that the grids become identical at the end~\cite{DBLP:conf/sigdial/JeknicSK24}. A sample episode is shown in Figure~\ref{fig:clean_up_example}. We provide all prompts and other details in Appendix~\ref{sec:appendix_clean_up}.

\begin{figure}[ht!]
    \centering
    % \newcounter{utterance}

\scriptsize  \setcounter{utterance}{1}
\setlength{\tabcolsep}{0pt}
\begin{supertabular}{c@{$\;$}|p{.15\linewidth}@{}p{.15\linewidth}p{.15\linewidth}p{.15\linewidth}p{.15\linewidth}p{.15\linewidth}}   
\# & $\;$\textbf{Player A} & \multicolumn{4}{c}{\textbf{Game Master}} & $\;\:$\textbf{Player B}\\
    \hline

\theutterance \stepcounter{utterance}  
    & & \multicolumn{4}{p{0.66\linewidth}}{
        \cellcolor[rgb]{0.9,0.9,0.9}{
            \makecell[{{p{\linewidth}}}]{
                \texttt{\tiny{[A$\langle$GM]}}
                {<GAME DESCRIPTION> ...}\\
                \centering
                \includegraphics[height=60pt]{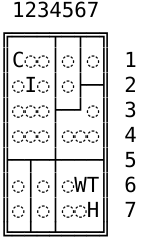}\\
                \raggedright
                \texttt{Please send a message to the other player.}
            }
        }
        \vspace{0.2cm}
    }
    & \\

\theutterance \stepcounter{utterance}  
    & \multicolumn{4}{p{0.6\linewidth}}{
        \cellcolor[rgb]{0.8,1,0.9}{
            \makecell[{{p{\linewidth}}}]{
                \texttt{\tiny{[A$\rangle$GM]}}
                \texttt{SAY: Hi! Let's coordinate our moves. [...]}
            }
        }
    }
    & & \\

\theutterance \stepcounter{utterance}  
    & & \multicolumn{4}{p{0.6\linewidth}}{
        \cellcolor[rgb]{0.9,0.9,0.9}{
            \makecell[{{p{\linewidth}}}]{
                \texttt{\tiny{[GM$\rangle$B]}}
                \centering{<GAME DESCRIPTION> ...}\\
                \centering
                \includegraphics[height=60pt]{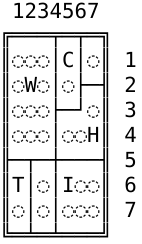}\\
                \raggedright
                \texttt{The other player started the game by sending this message:} \\
                \centering{<MESSAGE>}\\
                \raggedright
                \texttt{What is your first command?}
            }
        }
    }
    & \\

& \multicolumn{6}{c}{
\vdots
} \\

\theutterance \stepcounter{utterance}  
    & & & \multicolumn{4}{p{0.6\linewidth}}{
        \cellcolor[rgb]{1,0.85,0.72}{
            \makecell[{{p{\linewidth}}}]{
                \texttt{\tiny{[GM$\langle$B]}}
                \texttt{MOVE: C, (1,1)} \\
            }
        }
    } \\

\theutterance \stepcounter{utterance}  
    & & \multicolumn{4}{p{0.6\linewidth}}{
        \cellcolor[rgb]{0.9,0.9,0.9}{
            \makecell[{{p{\linewidth}}}]{
                \texttt{\tiny{[A$\langle$GM]}}
                Your message has been relayed to the other player.\\
                \centering <GAME STATE> \\
                \raggedright
                \texttt{The other player moved an object on their grid. \\What is your next command?}
            }
        }
    }
    & \\
    
\theutterance \stepcounter{utterance}  
    & \multicolumn{4}{p{0.6\linewidth}}{
        \cellcolor[rgb]{0.8,1,0.9}{
            \makecell[{{p{\linewidth}}}]{
                \texttt{\tiny{[A$\rangle$GM]}}
                \texttt{MOVE: I, (5,6)}
            }
        }
    }
    & & \\

& \multicolumn{6}{c}{
    \vdots
} \\
\theutterance \stepcounter{utterance}  
    & & & \multicolumn{4}{p{0.6\linewidth}}{
        \cellcolor[rgb]{1,0.85,0.72}{
            \makecell[{{p{\linewidth}}}]{
                \texttt{\tiny{[GM$\langle$B]}}
                \texttt{SAY: I think our grids are aligned now, finished?} \\
            }
        }
    } \\
\theutterance \stepcounter{utterance}  
    & \multicolumn{4}{p{0.6\linewidth}}{
        \cellcolor[rgb]{0.8,1,0.9}{
            \makecell[{{p{\linewidth}}}]{
                \texttt{\tiny{[A$\rangle$GM]}}
                \texttt{SAY: finished!}
            }
        }
    }
    & & \\
\end{supertabular}
    \caption{Example episode from Clean Up game to achieve a common goal configuration for a number of objects randomly placed on each player`s grid, and move them accordingly. Finally, both players have to agree the goal is reached to end the game.}
    
    \label{fig:clean_up_example}
\end{figure}

\paragraph{Game Mechanics} Both players are presented with $7\times7$ ASCII grids, with a number of randomly distributed objects in the form of capital letters placed on them. In each turn, a player can either send a message to their counterpart or move an object on their grid. The message and move text require certain formatting rules to follow. If a player does not follow the format, tries to move an object to a non-empty space or outside the grid bounds, or tries to move an object that doesn't exist, they receive a penalty and are re-prompted with information on the nature of their mistake. The game ends if (1) both players agree to end it, (2) round limit is exceeded, or (3) penalty limit is exceeded.

\paragraph{Game Instances}

\begin{figure}[ht]
    \centering
    % \newcounter{utterance}
{\scriptsize  \setcounter{utterance}{1}
\setlength{\tabcolsep}{0pt}
\begin{supertabular}{c@{$\;$}|p{.15\linewidth}@{}p{.15\linewidth}p{.15\linewidth}p{.15\linewidth}p{.15\linewidth}p{.15\linewidth}}   
\# & $\;$\textbf{Player A} & \multicolumn{4}{c}{\textbf{Game Master}} & $\;\:$\textbf{Player B}\\
    \hline

\theutterance \stepcounter{utterance}  
    & & \multicolumn{4}{p{0.66\linewidth}}{
        \cellcolor[rgb]{0.9,0.9,0.9}{
            \makecell[{{p{\linewidth}}}]{
                \texttt{\tiny{[A$\langle$GM]}}
                <GAME DESCRIPTION> ... \\
                \texttt{-Items and weights: Ball(2), Book(3), Magazine(1), Rope(4), Lamp(5)} \\
                \texttt{-Max Weight: 7} \\
                \texttt{-Your preferences: Ball=1, Book=2, Magazine=2, Rope=2, Lamp=1}
            }
        }
    }
    \vspace{0.05cm}
    & \\
    
\theutterance \stepcounter{utterance}  
    & \multicolumn{4}{p{0.60\linewidth}}{
        \cellcolor[rgb]{0.8,1,0.9}{
            \makecell[{{p{\linewidth}}}]{
                \texttt{\tiny{[A$\rangle$GM]}}
                \texttt{STRATEGIC REASONING: Magazine, Ball, Book have the best value-to-weight ratios while respecting the weight limit.}\\
                \texttt{PROPOSAL: Magazine, Ball, Book} \\
                \texttt{ARGUMENT: These items balance weight and value well for me. What do you think?}
            }
        }
        \vspace{0.2cm}
    }
    & & \\

\theutterance \stepcounter{utterance}  
    & & \multicolumn{4}{p{0.6\linewidth}}{
        \cellcolor[rgb]{0.9,0.9,0.9}{
            \makecell[{{p{\linewidth}}}]{
                \texttt{\tiny{[GM$\rangle$B]}}
                <GAME DESCRIPTION> ...\\
                \texttt{<Items, weights, weight limit>}\\
                \texttt{-Your preferences: Ball=1, Book=1, Magazine=2, Rope=4, Lamp=5} \\
                \centering{<PLAYER A Message without \texttt{STRATEGIC REASONING}>}
            }
        }
    }
    \vspace{0.09cm}
    & \\

\theutterance \stepcounter{utterance}  
    & & & \multicolumn{4}{p{0.6\linewidth}}{
        \cellcolor[rgb]{1,0.85,0.72}{
            \makecell[{{p{\linewidth}}}]{
                \texttt{\tiny{[GM$\langle$B]}}
                \texttt{STRATEGIC REASONING: I can increase the value of the other player's proposal for me by swapping Rope with Book. Let me propose this.}\\
                \texttt{PROPOSAL: Rope, Ball, Magazine}\\
                \texttt{ARGUMENT: I value Rope highly, but not Book. How about we swap them?}
            }
        }
    } 
    \vspace{0.15cm}
    \\

\theutterance \stepcounter{utterance}  
    & & \multicolumn{4}{p{0.6\linewidth}}{
        \cellcolor[rgb]{0.9,0.9,0.9}{
            \makecell[{{p{\linewidth}}}]{
                \texttt{\tiny{[A$\langle$GM]}}
                \centering{<PLAYER B Message without \texttt{STRATEGIC REASONING}>}
            }
        }
    }
    & \\

\theutterance \stepcounter{utterance}  
    & \multicolumn{4}{p{0.6\linewidth}}{
        \cellcolor[rgb]{0.8,1,0.9}{
            \makecell[{{p{\linewidth}}}]{
                \texttt{\tiny{[A$\rangle$GM]}}
                \texttt{STRATEGIC REASONING: I value Rope and Book equally. Weight of proposal is 4+2+1=7 which matches the limit. I should accept.}\\
                \texttt{AGREE: Rope, Ball, Magazine} \\
                \texttt{ARGUMENT: I am indifferent w.r.t. Rope and Book. Fair deal.} \\
            }
        }
    }
    & & \\
    
\end{supertabular}
}
    \caption{An example episode of the \textit{Air Balloon Survival} game. Two players must negotiate and argue for their preferred set of items. and must explicitly agree to a proposal made by the other.}
    \label{fig:air_balloon_example}
    
\end{figure}

All instances are created programmatically by creating different grids ($7\times7$), which contain obstacles in form of horizontal and vertical lines, branches, crossings, and corners. The maximum number of rounds is fixed to $4 \times n_{obj}$ (where $n_{obj}$ is the object count). 
% Players can collectively accumulate $2 \times n_{obj} + 2$ penalties. 
We create different sets of experiments by controlling two aspects: number of empty cells and objects. We have three difficulty levels. Of the 49 total cells, $34$ of cells are empty on the \textit{easy}, $29$ on \textit{medium}, and $24$ on \textit{hard} levels. For each level, we sample three grids, and then place $3$, $5$, or $7$ objects on them, making for \textbf{27 instances in total}.

\subsection{Air Balloon Survival}
This game evaluates advanced reasoning and interactive collaboration between players, which was previously introduced by \citet{howes2021justifiable} to study how patients with schizophrenia verbalise their reasoning during social encounters. In our version, two players are on a sinking hot air balloon, and they have to agree on which items to keep (or throw out) so that the weight of the balloon is reduced to keep floating. Each player has hidden preference values for items and must negotiate to maximise their combined utility score. An example episode is given in Figure~\ref{fig:air_balloon_example}. The game tests individual reasoning through constraint-based optimisation, requiring arithmetic and combinatorial search, collective reasoning through practical rationality, and theory of mind (to infer their counterpart's hidden preferences from their responses to reach an optimal agreement). We provide all prompts and other details in Appendix~\ref{sec:appendix_airballoon}.

\paragraph{Game Mechanics}
Both players receive their assigned preference values for the items. They are instructed to use specified formats when making proposals and engaging in negotiations. The game is aborted if no progress is made for eight consecutive turns. 
% Each player may make up to two mistakes (either by violating game rules or the required syntax). 
Unlike the other two games, players are also instructed to output their \texttt{STRATEGIC REASONING} along with the expected message. Only the expected message is passed to the other player. It allows players to ``think out loud'' about their choices. The game ends when a proposal made by one player is accepted by the other.

\paragraph{Instance Generation}

We draw either $15$ or $35$ items (depending on the experiment) from a randomly generated list constructed by concatenating a capital letter with a two-digit number (e.g., \texttt{A42}, \texttt{C07})\footnote{This keeps the naming of items language-agnostic and abstract from conventional examples of the \textit{0/1 Knapsack Problem} models that may have been seen during training}. We assign a weight value to each item. The air balloon's capacity is defined as a fraction of the combined weight of all items. We experiment with two negotiation levels, where we specifically set valuations over items to be the same for players, giving them common goals (easy level), or we invert their preference orderings, giving them opposing goals (hard level). We also experiment with generating the \texttt{STRATEGIC REASONING} or not. Lastly, we conduct complexity experiments by increasing the number of items. In total, we have six experiments with each having six instances, which leads to the \textbf{total number of 36 instances} for this game.

\section{Experimental Setup}\label{sec:experimental_setup}

\subsection{Game Instances}

\textit{Deal or No Deal}, \textit{Clean Up}, and \textit{Air Balloon Survival} games include 40, 27, and 36 experimental instances, respectively. Each instance is then initialised with defined prompt templates. The same game instances are used for the English, German, and Italian experiments because the instances are language-agnostic; only the prompt templates need to be aligned for a specific language. 

%All prompt templates are provided in Appendix \ref{sec:dond_prompt_templates}, \ref{clean_up:prompts}, \ref{sec:hot_air_balloon_appendix_prompt_templates}.

\subsection{Evaluation Metrics}

The \textit{clembench framework}~\cite{chalamalasetti-etal-2023-clembench} requires each implemented game to provide two primary metrics: \% Played and Quality Score. The \textit{\% Played} stands for the percentage of episodes where the evaluated language models followed instructions and the game was terminated by the defined end states. In cases where the gameplay does not fit the defined states of the game, the Game Master either tolerates such behaviour for one or two turns, asks the players to try again, or aborts the game. The \textit{Quality Score} is calculated using an objective function to measure how closely the played episode aligns with the target goal. In each game, this metric is calculated for all episodes that have been played. Once two metrics are calculated, they are aggregated to a single number, the \textit{clemscore} as the normalised product of \textit{\% Played} and \textit{Quality Score} (scaled to the interval [0, 100]). Next, we describe the Quality Scores for games.

\paragraph{Deal or No Deal}

For the cooperative game mode, we measure quality as the ratio of achieved total score to maximum possible total score. For the semi-competitive setting, we use a \textit{Pareto efficiency} metric that measures how far the agreement is from optimal by calculating potential one-sided improvements (more details in Appendix~\ref{subsec:appendix_dond_metric}).

\paragraph{Clean Up}

The Quality Score combines two main components: how well players organised objects spatially and how many rule violations they incurred. The metric uses \textit{Euclidean distances} between matching objects to compare the final arrangement to both the initial setup and a random baseline, rewarding improvements in object alignment while penalising rule violations through a scaling factor that becomes increasingly harsh as violations approach the maximum allowed limit 
(more details in Appendix~\ref{subsec:appendix_cleanup_evaluation_metric}).

\begin{table*}[ht]
\centering
\footnotesize
\setlength{\tabcolsep}{4.5pt}
\definecolor{lightgreen}{RGB}{220,255,220}
\definecolor{lightblue}{RGB}{220,235,255}
\definecolor{darkgreen}{rgb}{0.0, 0.5, 0.0}
\definecolor{darkred}{rgb}{0.8, 0.0, 0.0}
% The '|' between the last two 'c's ensures the line appears in data rows by default
\begin{tabular}{c|l|cc|cc|cc|cc|cc|cc|c|c} 
\multirow{9}{*}{\rotatebox{90}{\textbf{EN}}} & & \multicolumn{2}{c}{\textbf{GPT-5}} & \multicolumn{2}{c}{\begin{tabular}{c}\textbf{GPT-5}\\\textbf{mini}\end{tabular}} & \multicolumn{2}{c}{\textbf{CL-4}} & \multicolumn{2}{c}{\textbf{LM-70B}} & \multicolumn{2}{c}{\textbf{Nem-9B}} & \multicolumn{2}{c}{\textbf{Qwen-3}} & 
% This \multicolumn overrides the default and removes the vertical line for this cell only
\multicolumn{1}{c}{\begin{tabular}{c}\textbf{GPT}\\\textbf{OSS}\end{tabular}} & \textbf{DS-v3.1} \\ \cline{2-16}

& \textbf{Games} & On & Off & On & Off & On & Off & On & Off & On & Off & On & Off & On & Off \\ \hline
 & DoND & \cellcolor{lightgreen}87.9 & \cellcolor{lightblue}32.3 & \cellcolor{lightgreen}75.7 & \cellcolor{lightblue}23.7 & \cellcolor{lightgreen}\textbf{94.4} & \cellcolor{lightblue}90.1 & \cellcolor{lightgreen}24.3 & \cellcolor{lightblue}43.5 & \cellcolor{lightgreen}18.5 & \cellcolor{lightblue}4.0 & \cellcolor{lightgreen}57.9 & \cellcolor{lightblue}22.5 & \cellcolor{lightgreen}50.0 & \cellcolor{lightblue}59.0 \\ 
 & Clean Up & \cellcolor{lightgreen}\textbf{99.8} & \cellcolor{lightblue}75.2 & \cellcolor{lightgreen}96.6 & \cellcolor{lightblue}77.3 & \cellcolor{lightgreen}85.5 & \cellcolor{lightblue}81.9 & \cellcolor{lightgreen}4.1 & \cellcolor{lightblue}28.1 & \cellcolor{lightgreen}28.5 & \cellcolor{lightblue}35.1 & \cellcolor{lightgreen}87.9 & \cellcolor{lightblue}35.8 & \cellcolor{lightgreen}81.4 & \cellcolor{lightblue}76.4 \\ 
 & Air Balloon & \cellcolor{lightgreen}\textbf{98.0} & \cellcolor{lightblue}80.1 & \cellcolor{lightgreen}97.5 & \cellcolor{lightblue}82.2 & \cellcolor{lightgreen}83.8 & \cellcolor{lightblue}26.2 & \cellcolor{lightgreen}2.3 & \cellcolor{lightblue}40.9 & \cellcolor{lightgreen}14.3 & \cellcolor{lightblue}13.2 & \cellcolor{lightgreen}88.7 & \cellcolor{lightblue}0 & \cellcolor{lightgreen}78.2 & \cellcolor{lightblue}81.7 \\ 
 & \textbf{Average} & \cellcolor{lightgreen}\textbf{95.2} & \cellcolor{lightblue}62.5 & \cellcolor{lightgreen}89.9 & \cellcolor{lightblue}61.1 & \cellcolor{lightgreen}87.9 & \cellcolor{lightblue}66.1 & \cellcolor{lightgreen}10.2 & \cellcolor{lightblue}37.5 & \cellcolor{lightgreen}20.4 & \cellcolor{lightblue}17.4 & \cellcolor{lightgreen}78.2 & \cellcolor{lightblue}19.4 & \cellcolor{lightgreen}69.9 & \cellcolor{lightblue}72.4 \\ 
& \textbf{Margin} & \multicolumn{2}{c|}{\textcolor{darkgreen}{+32.7}} & \multicolumn{2}{c|}{\textcolor{darkgreen}{+28.8}} & \multicolumn{2}{c|}{\textcolor{darkgreen}{+21.8}} & \multicolumn{2}{c|}{\textcolor{darkred}{-27.3}} & \multicolumn{2}{c|}{\textcolor{darkgreen}{+3.0}} & \multicolumn{2}{c|}{\textcolor{darkgreen}{+58.8}} & - & -\\ \hline \hline

\multirow{4}{*}{\rotatebox{90}{\textbf{DE}}}
 & DoND & \cellcolor{lightgreen}\textbf{86.0} & \cellcolor{lightblue}19.6 & \cellcolor{lightgreen}82.9 & \cellcolor{lightblue}24.4 & \cellcolor{lightgreen}80.3 & \cellcolor{lightblue}77.1 & \cellcolor{lightgreen}21.8 & \cellcolor{lightblue}53.4 & \cellcolor{lightgreen}0 & \cellcolor{lightblue}4.5 & \cellcolor{lightgreen}62.1 & \cellcolor{lightblue}12.8 & \cellcolor{lightgreen}34.9 & \cellcolor{lightblue}42.4 \\ 
 & Clean Up & \cellcolor{lightgreen}98.2 & \cellcolor{lightblue}70.4 & \cellcolor{lightgreen}\textbf{98.8} & \cellcolor{lightblue}67.0 & \cellcolor{lightgreen}94.3 & \cellcolor{lightblue}83.1 & \cellcolor{lightgreen}5.3 & \cellcolor{lightblue}25.3 & \cellcolor{lightgreen}23.1 & \cellcolor{lightblue}0 & \cellcolor{lightgreen}86.4 & \cellcolor{lightblue}27.5 & \cellcolor{lightgreen}74.4 & \cellcolor{lightblue}53.7 \\ 
 & Air Balloon & \cellcolor{lightgreen}\textbf{98.0} & \cellcolor{lightblue}81.4 & \cellcolor{lightgreen}94.9 & \cellcolor{lightblue}86.2 & \cellcolor{lightgreen}85.4 & \cellcolor{lightblue}27.8 & \cellcolor{lightgreen}51.5 & \cellcolor{lightblue}43.7 & \cellcolor{lightgreen}17.4 & \cellcolor{lightblue}16.3 & \cellcolor{lightgreen}76.0 & \cellcolor{lightblue}8.8 & \cellcolor{lightgreen}90.7 & \cellcolor{lightblue}75.3 \\ 
 & \textbf{Average} & \cellcolor{lightgreen}\textbf{94.1} & \cellcolor{lightblue}57.1 & \cellcolor{lightgreen}92.2 & \cellcolor{lightblue}59.2 & \cellcolor{lightgreen}86.7 & \cellcolor{lightblue}62.7 & \cellcolor{lightgreen}26.2 & \cellcolor{lightblue}40.8 & \cellcolor{lightgreen}13.5 & \cellcolor{lightblue}6.9 & \cellcolor{lightgreen}74.8 & \cellcolor{lightblue}16.4 & \cellcolor{lightgreen}66.7 & \cellcolor{lightblue}57.1 \\
& \textbf{Margin} & \multicolumn{2}{c|}{\textcolor{darkgreen}{+37.0}} & \multicolumn{2}{c|}{\textcolor{darkgreen}{+33.0}} & \multicolumn{2}{c|}{\textcolor{darkgreen}{+24.0}} & \multicolumn{2}{c|}{\textcolor{darkred}{-14.6}} & \multicolumn{2}{c|}{\textcolor{darkgreen}{+6.6}} & \multicolumn{2}{c|}{\textcolor{darkgreen}{+58.4}} & - & - \\ \hline \hline

\multirow{4}{*}{\rotatebox{90}{\textbf{IT}}}
 & DoND & \cellcolor{lightgreen}78.0 & \cellcolor{lightblue}33.8 & \cellcolor{lightgreen}77.1 & \cellcolor{lightblue}26.8 & \cellcolor{lightgreen}\textbf{84.4} & \cellcolor{lightblue}67.8 & \cellcolor{lightgreen}17.3 & \cellcolor{lightblue}14.3 & \cellcolor{lightgreen}13.9 & \cellcolor{lightblue}11.2 & \cellcolor{lightgreen}55.6 & \cellcolor{lightblue}19.4 & \cellcolor{lightgreen}29.0 & \cellcolor{lightblue}28.9 \\ 
 & Clean Up & \cellcolor{lightgreen}89.0 & \cellcolor{lightblue}68.8 & \cellcolor{lightgreen}\textbf{93.6} & \cellcolor{lightblue}77.2 & \cellcolor{lightgreen}91.6 & \cellcolor{lightblue}82.6 & \cellcolor{lightgreen}6.5 & \cellcolor{lightblue}31.1 & \cellcolor{lightgreen}21.5 & \cellcolor{lightblue}12.7 & \cellcolor{lightgreen}82.6 & \cellcolor{lightblue}31.5 & \cellcolor{lightgreen}63.6 & \cellcolor{lightblue}67.6 \\ 
 & Air Balloon & \cellcolor{lightgreen}97.2 & \cellcolor{lightblue}88.0 & \cellcolor{lightgreen}\textbf{97.3} & \cellcolor{lightblue}87.8 & \cellcolor{lightgreen}79.4 & \cellcolor{lightblue}27.0 & \cellcolor{lightgreen}2.7 & \cellcolor{lightblue}40.8 & \cellcolor{lightgreen}0 & \cellcolor{lightblue}0 & \cellcolor{lightgreen}77.8 & \cellcolor{lightblue}12.9 & \cellcolor{lightgreen}83.7 & \cellcolor{lightblue}76.4 \\ 
 & \textbf{Average} & \cellcolor{lightgreen}88.1 & \cellcolor{lightblue}63.5 & \cellcolor{lightgreen}\textbf{89.3} & \cellcolor{lightblue}63.9 & \cellcolor{lightgreen}85.1 & \cellcolor{lightblue}59.1 & \cellcolor{lightgreen}8.8 & \cellcolor{lightblue}28.7 & \cellcolor{lightgreen}11.8 & \cellcolor{lightblue}8.0 & \cellcolor{lightgreen}72.0 & \cellcolor{lightblue}21.3 & \cellcolor{lightgreen}58.8 & \cellcolor{lightblue}57.6 \\
 & \textbf{Margin} & \multicolumn{2}{c|}{\textcolor{darkgreen}{+24.6}} & \multicolumn{2}{c|}{\textcolor{darkgreen}{+25.4}} & \multicolumn{2}{c|}{\textcolor{darkgreen}{+26.0}} & \multicolumn{2}{c|}{\textcolor{darkred}{-19.9}} & \multicolumn{2}{c|}{\textcolor{darkgreen}{+3.8}} & \multicolumn{2}{c|}{\textcolor{darkgreen}{+50.7}} & - & - \\ \hline
\hline

\multicolumn{2}{c|}{\textbf{Overall Average}} & \cellcolor{lightgreen}\textbf{92.5} & \cellcolor{lightblue}61.1 & \cellcolor{lightgreen}90.5 & \cellcolor{lightblue}61.4 & \cellcolor{lightgreen}86.6 & \cellcolor{lightblue}62.6 & \cellcolor{lightgreen}15.1 & \cellcolor{lightblue}35.7 & \cellcolor{lightgreen}15.2 & \cellcolor{lightblue}10.8 & \cellcolor{lightgreen}75.0 & \cellcolor{lightblue}19.0 & \cellcolor{lightgreen}65.1 & \cellcolor{lightblue}62.4 \\
\multicolumn{2}{c|}{\textbf{Overall Margin}} & \multicolumn{2}{c|}{\textcolor{darkgreen}{+31.4}} & \multicolumn{2}{c|}{\textcolor{darkgreen}{+29.1}} & \multicolumn{2}{c|}{\textcolor{darkgreen}{+24.0}} & \multicolumn{2}{c|}{\textcolor{darkred}{-20.6}} & \multicolumn{2}{c|}{\textcolor{darkgreen}{+4.4}} & \multicolumn{2}{c|}{\textcolor{darkgreen}{+56.0}} & - & - \\ \hline

\end{tabular}
\caption{\textit{clemscore} values for three negotiation games on selected LLMs for English, German, and Italian versions. \textit{On}: reasoning mode is turned on, \textit{Off}: reasoning mode is turned off. The best result for each language in each row is highlighted in bold. \textit{CL}: Claude, \textit{LM}: Llama-3.3, \textit{DS}: Deepseek, \textit{Nem}: Nemotron-v2, \textit{DoND}: Deal or No Deal }
\label{tab:main-results}
\end{table*}

\paragraph{Air Balloon Survival}

Each player receives a score based on their achieved utility relative to their optimal knapsack solution. The overall game score uses the harmonic mean of both players' normalised scores, which rewards balanced outcomes and penalises deals where one player benefits disproportionately (more details in Appendix~\ref{subsec:appendix_balloon_metrics}).

\subsection{Evaluated Models}

We have evaluated both commercial and open-weight models that have reasoning functionality. We selected \textit{GPT-5}, \textit{GPT-5-mini}, \textit{Claude-4} from commercial models. From open-weight models, we selected \textit{Llama3.3-70B} and \textit{Deepseek-R1-distilled-llama-70B} as its reasoning counterpart, \textit{Nemotron-Nano-9B-v2}, \textit{Qwen-3-80B} (\textit{instruction} and \textit{thinking} versions for reasoning off and on modes, respectively), \textit{GPT-OSS-120B} with reasoning on mode, and finally \textit{Deepseek-v3.1} with reasoning off mode. For commercial models, we ran them via their respective API backends, and we used \textit{OpenRouter.ai} for open-weight models.

\section{Results}

\subsection{Overall Performance}

\begin{figure*}[ht]
    \centering
    \begin{subfigure}[b]{0.49\textwidth}
        \centering
        \includegraphics[width=\linewidth]{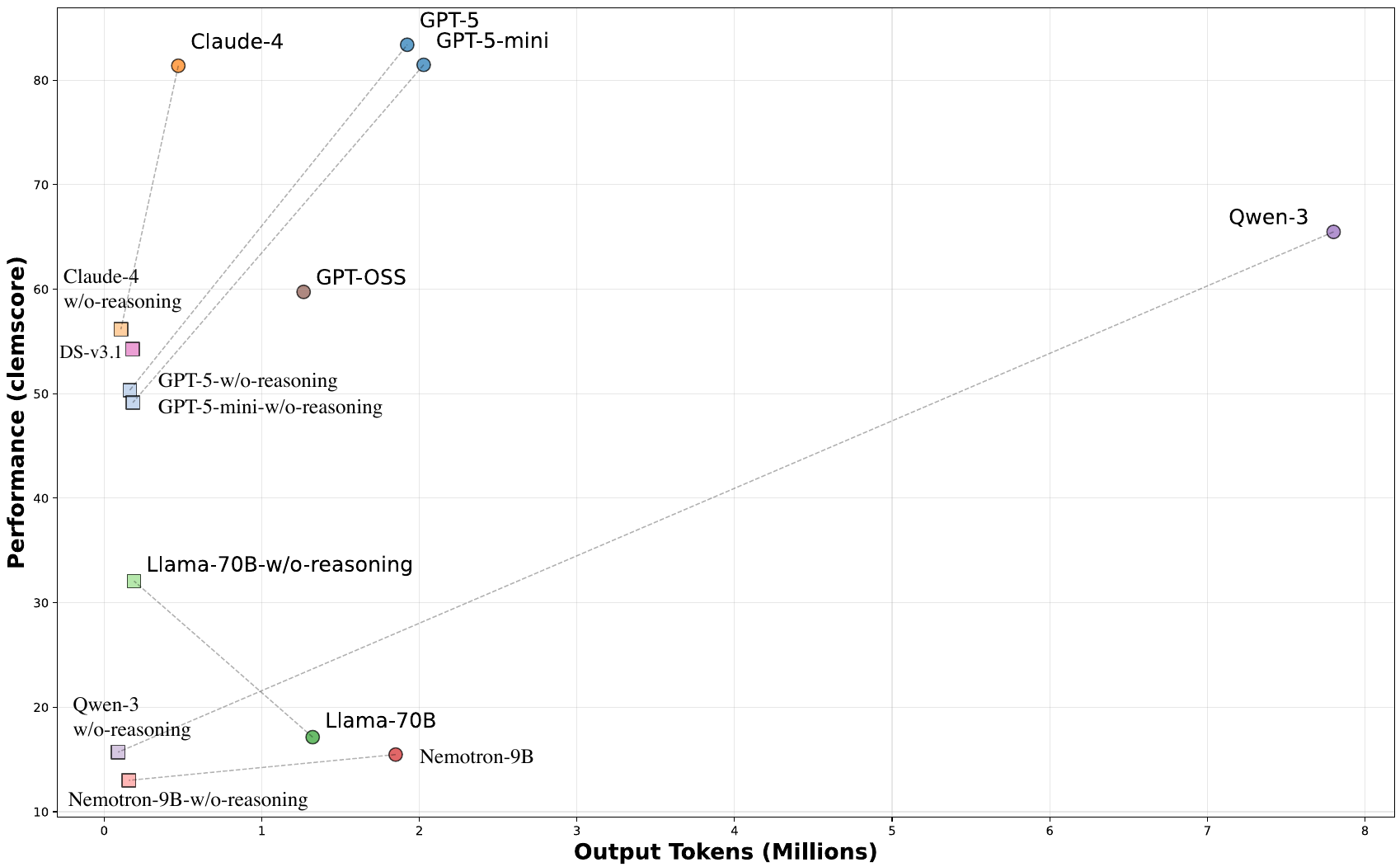}
        \caption{Performance and the output tokens for all evaluated models with their reasoning on/off modes averaged for three languages.}
        \label{fig:token_average}
    \end{subfigure}
    \hfill
    \begin{subfigure}[b]{0.49\textwidth}
        \centering
        \includegraphics[width=\linewidth]{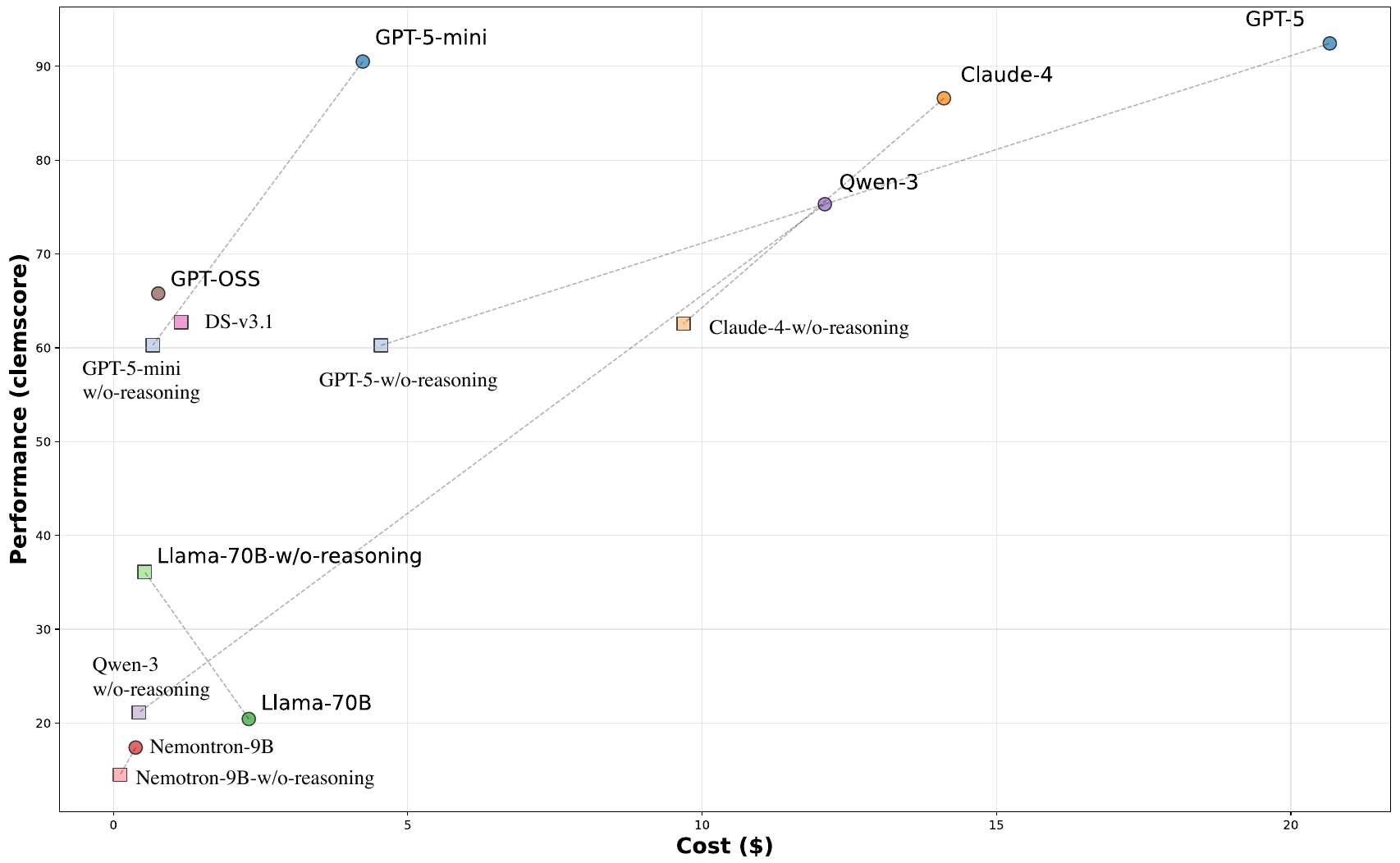}
        \caption{Performance and the cost of for all evaluated models with their reasoning on/off modes averaged for three languages.}
        \label{fig:cost_average}
    \end{subfigure}
    \caption{Trade-off between performance and cost comparison averaged across languages.}
    \label{fig:combined-token-cost}
    
\end{figure*}

We present the results obtained by the models (in \textit{reasoning on} and \textit{off} modes, where possible) evaluated for the English, German, and Italian versions of the negotiation games in Table~\ref{tab:main-results}. 
% \textit{GPT-OSS} (reasoning on) and \textit{Deepseek-v3.1} (reasoning off) were run in one mode only due to their underlying architecture.

\textbf{The Effect of Reasoning}: the most striking observation here is that reasoning mode dramatically improves performance across many models and languages, with \textit{Qwen-3} gaining $56$ points averaged across all games. Similar pattern exists for \textit{GPT}, \textit{Claude} and even smaller \textit{Nemotron} models. This is a strong indication that deliberate reasoning significantly enhances strategic game-playing abilities. Only with \textit{Llama-70B} do we see different results, which may be due to the effect of distillation.

\textbf{Multilingual Capabilities}: results for German shows the largest average margin between \textit{reasoning on} and \textit{off} modes of models. English and Italian performances are also strong, suggesting that certain models have particularly sufficient negotiation capabilities across these evaluated languages.

\textbf{Model Comparison}: GPT-5 is the clear winner, with GPT-5-mini and Claude-4 getting very close performance across all languages. Interestingly, GPT-5 mini sometimes matches or exceeds full GPT-5 performance (particularly in Italian). \textit{Qwen-3} shows the biggest performance jump.

\subsection{Trade-off between Reasoning Overhead and Performance}

\textbf{RQ1}: \textit{What is the computational and performance trade-off between reasoning overhead and negotiation effectiveness across tasks and languages?}

In Figure~\ref{fig:token_average}, we present a plot that shows the average performance across all games and languages and how many output tokens it requires. Here, all reasoning and completion tokens are summed up. Most models generate the number of tokens that are closer to each other, and \textit{Qwen-3} generates almost \textit{4x} more tokens than others.

In Figure~\ref{fig:cost_average}, we present the overall cost (input and output tokens together) of experiments for each model. As expected, models with reasoning mode cost more, with \textit{GPT-5}, being the most expensive, costing almost \textit{4x} compared to non-reasoning version while improving $31.4$ points in clemscore. In terms of deciding on the best trade-off between performance and cost, \textbf{\textit{GPT-5-mini} is the most cost-efficient commercial model, and \textit{GPT-OSS} is the best open-weight one, with a fraction of the cost compared to Qwen-3}.

\subsection{Language Consistency in Reasoning}

\textbf{RQ2}: \textit{To what extent do models maintain language consistency in their reasoning processes when performing multilingual negotiation tasks?}

% \begin{figure*}[ht!]
%     \centering
% \includegraphics[width=\linewidth]{figures/language_distribution_reasoning_messages.pdf}
%         \caption{Language distribution in reasoning tokens}
%         \label{fig:lang_distribution_reasoning}
% \end{figure*}

\begin{figure}[ht!]
    \centering
    \includegraphics[width=\linewidth]{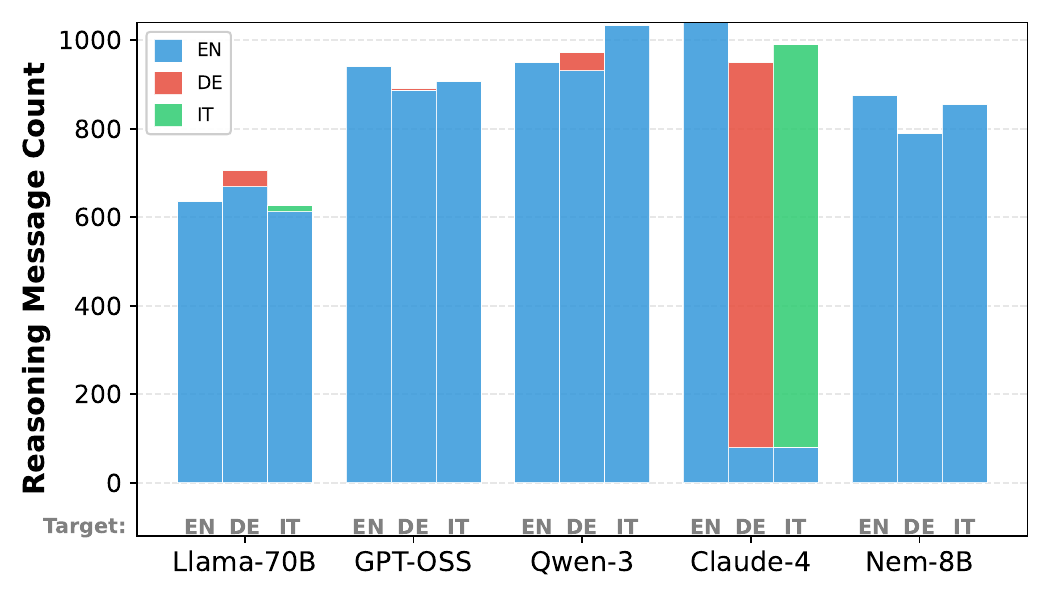}
            \caption{Language distribution in reasoning tokens}
        \label{fig:lang_distribution_reasoning}
\end{figure}

\begin{table}[ht!]
\centering
\setlength{\tabcolsep}{4pt}
    \begin{tabular}{ll|ccc}
    & & \textbf{DE} & \textbf{EN} & \textbf{IT} \\
    \hline
    \multirow{2}{*}{Commercial} & completion & 1.00 & 1.00 & 1.00 \\
    & reasoning & 0.84 & 1.00 & 0.84 \\
    \hline
    \multirow{2}{*}{Open-weight} & completion & 0.94 & 1.00 & 0.87 \\
    & reasoning & 0.06 & 1.00 & 0.03 \\
\end{tabular}
    
    \caption{Language consistency in reasoning tokens}
    \label{fig:lang_analysis}
\end{table}

\begin{figure*}[ht]
  \centering
  \includegraphics[width=0.75\textwidth]{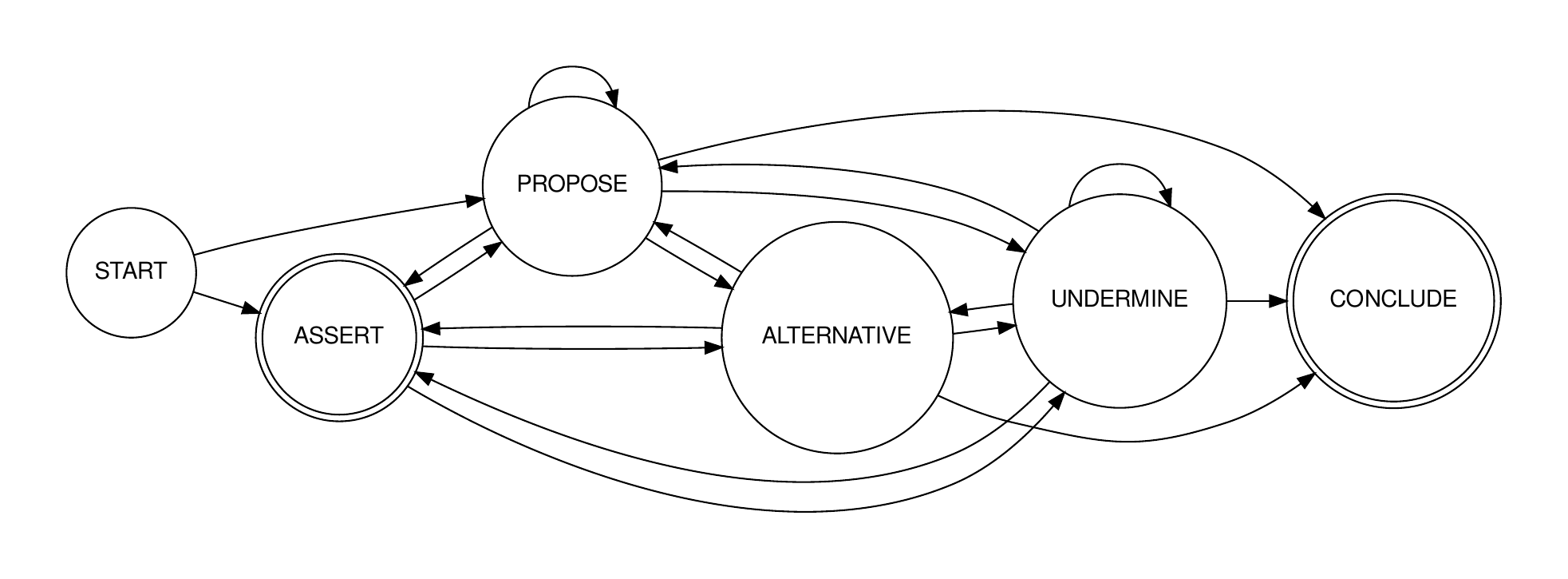}
  \caption{Finite state machine for analysing reasoning traces}
  \label{fig:fsm_main}
\end{figure*}

\begin{figure*}[ht]
    \centering
    \begin{subfigure}[b]{0.48\textwidth}
        \centering
        \includegraphics[width=\linewidth]{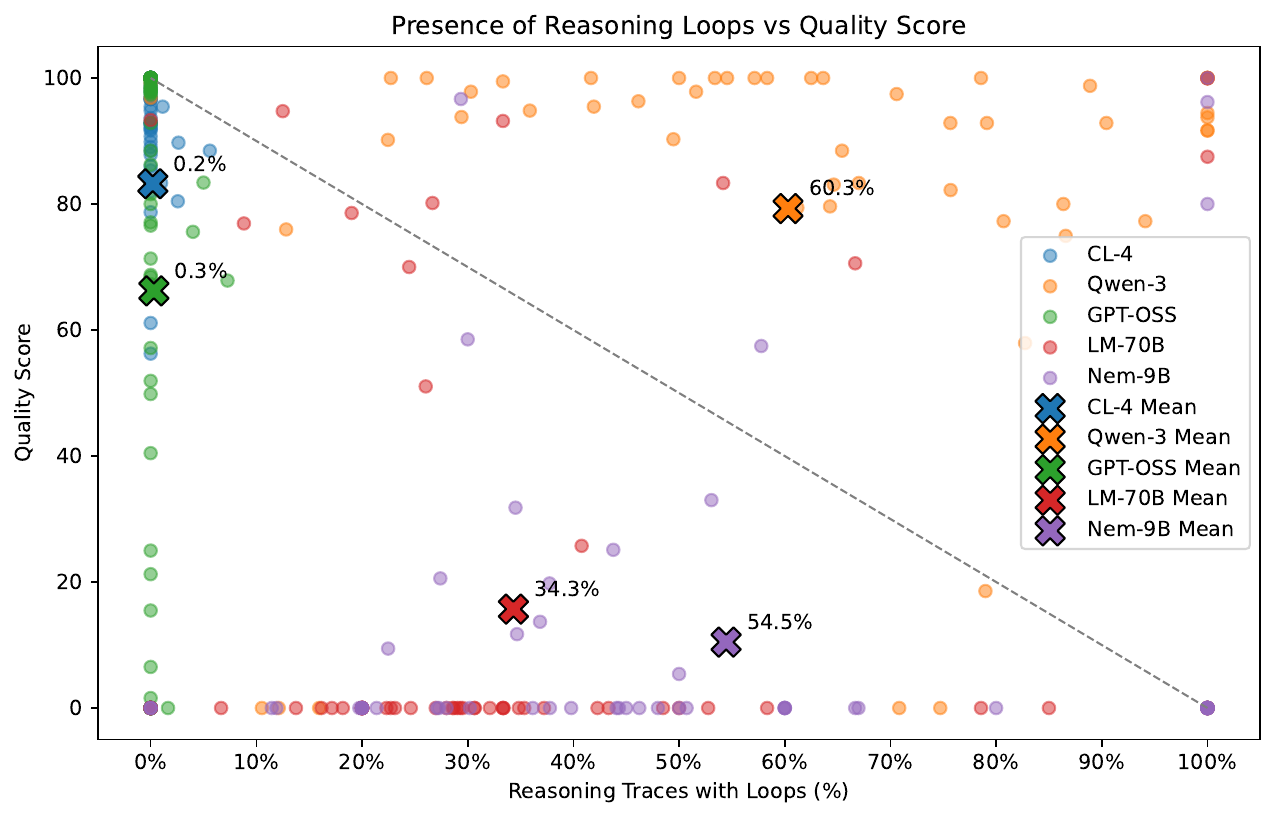}
        \caption{Reasoning traces that display thinking loops (at least three repetitions of the same action) plotted}
        \label{fig:reasoning_loops}
    \end{subfigure}
    \hfill
    \begin{subfigure}[b]{0.48\textwidth}
        \centering
        \includegraphics[width=\linewidth]{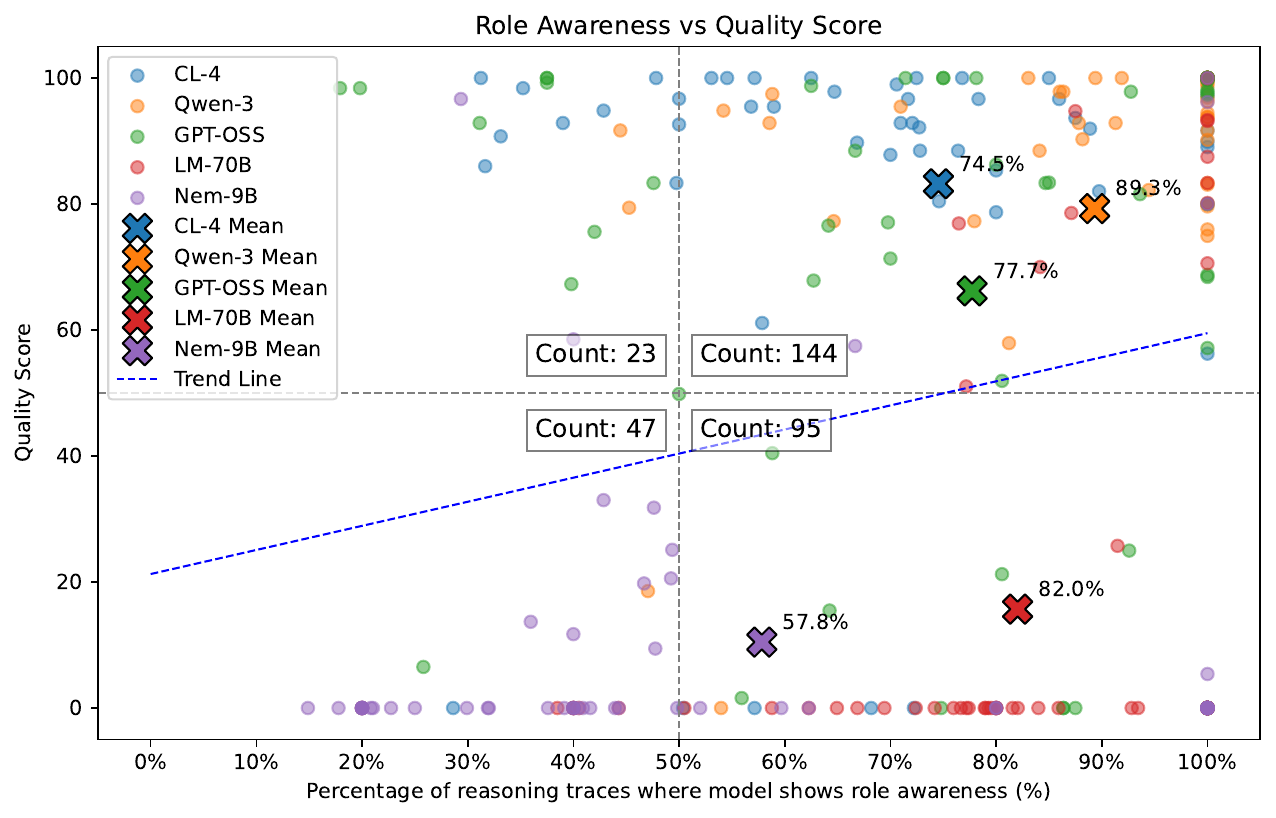}
        \caption{Reasoning traces where the model displays both awareness of their own another player's role}
        \label{fig:role_awareness}
    \end{subfigure}
    \caption{Analysis of patterns in model reasoning traces}
    \label{fig:reasoning_analysis}
\end{figure*}

We ran a language detection script to analyse the reasoning message of the models. The results are presented in Figure~\ref{fig:lang_distribution_reasoning}\footnote{Note: \textit{GPT-5} models do not return reasoning tokens, thus they are excluded from further analysis}. We clearly see that all open-weight models mostly generate reasoning tokens in English, whereas \textit{Claude-4} ``thinks'' in the respective language of the task, consistently across all three of them. In Table~\ref{fig:lang_analysis}, we provide the percentage of language consistency (whether the language of the message matches the target language) across \textit{completion} and \textit{reasoning} tokens. 

We can clearly see that \textit{Claude-4} (the only commercial model in the analysis) exhibits consistency in both completion and reasoning tokens. Open-weight models retain consistency in completion tokens (thus, abiding by the formatting rules as each game requires prefixes in the respective language), but fail to do so with reasoning tokens for German and Italian versions of the games. Thus, we can conclude that \textbf{open-weight models do not maintain language consistency in their reasoning processes across multilingual tasks}. Similar patterns have been also observed by \citet{qi-etal-2025-models} where explicitly forcing the model to reason in the respective language even reduces the performance. Thus, the models that do not maintain language consistency in their reasoning steps \textbf{become less interpretable and less trustworthy}.

\subsection{Strategic Adaptation Across Multi-Turn Interactions}

\textbf{RQ3}: \textit{Do models demonstrate strategic adaptation over multiple turns, or merely surface-level pattern matching creating an illusion of thinking?}

\subsubsection{Keyword-based Analysis}
We label the first-turn reasoning traces as sequences over five discrete states—\texttt{ASSERT}, \texttt{PROPOSE}, \texttt{UNDERMINE}, \texttt{ALTERNATIVE}, \texttt{CONCLUDE}. The states are detected using lexical cues in reasoning tokens, e.g. ``maybe/could'': \texttt{PROPOSE}; ``but/however'': \texttt{UNDERMINE}, etc. Each trace is a path on a finite-state machine with observed transitions (including self-loops), from which we compute per-trace statistics and aggregate at model/game level. The finite state machine is shown in Figure~\ref{fig:fsm_main}. We compute two metrics using the transitions among states (more details in Appendix~\ref{sec:cyclic_analysis}):

\textit{number of segments}: we partition the labels extracted from each reasoning trace by absorbing states (\texttt{ASSERT} and  \texttt{CONCLUDE}). The count serves as coarse “reasoning units”: higher counts could mean more dispersed reasoning (multiple restarts before final commitment); lower counts could mean more concentrated reasoning (a plan developed and resolved within fewer units).

\textit{cycle edge ratio}: within each segment’s induced subgraph, we mark edges that belong to any simple cycle and compute the fraction of transitions on such edges, then aggregate across segments to obtain one single metric for each reasoning trace. Higher ratios indicate local looping/hedging (e.g., \texttt{PROPOSE} <-> \texttt{UNDERMINE}); lower ratios indicate forward motion towards absorbing states.

Good-performing models should resolve plans in fewer segments and cycles to have more goal-oriented planning and execution of it. The results are given in Figure~\ref{fig:clemscore_VS_n_segment} and Figure~\ref{fig:clemscore_VS_cycle_edge_ratio}. Performance shows non-linear relationships: \textbf{minimal segment counts correlate with improved outcomes}, while moderate additions yield unclear results. \textbf{Commercial models achieve competitive scores with lower cycle ratios} and tighter clustering; open-weight models show higher ratios with greater variance, suggesting that additional cycles do not reliably translate into performance gains.

\subsubsection{LLM-based Analysis}

For all thinking models with available reasoning traces, we sampled a total of 309 transcripts with at least one instance for each game, language, and experiment combination for automated analysis. We prompted \textit{GPT-5} to analyse certain aspects of the traces and give their output in JSON format, including not only the analysis, but also short explanations for each classification (see example in Figure~\ref{fig:llm_analysis_prompt_and_example}). More details are in Appendix~\ref{sec:llm-based_reasoning_analysis}.

Figure~\ref{fig:reasoning_loops} shows the thinking loops that occur in reasoning. We consider a thought to be in a loop if it occurs at least three times. Good-performing model results should be clustered around the top left, including fewer loops and a high quality score. \textit{Claude-4} and \textit{GPT-OSS} rarely display loops (0.2\% and 0.3\%) and achieve high scores, while \textit{Llama-70B} (34.3\%) and \textit{Nemotron-9B} (54.4\%) contain loops and score low. 95.18\% of data points for these models fall below the dotted diagonal, indicating that reasoning loops, or \textit{overthinking} \cite{DBLP:journals/corr/abs-2506-06941}, reduce performance.
Conversely, \textit{Qwen-3} shows 60.3\% of traces with loops but averages 79.3 quality score. These excessive loops explain \textit{Qwen-3}'s high token counts.

Since the games prompt the model to be active agents, we analyse the reasoning traces for both awareness of their own player role (whether the model is helping or advising somebody else, or actively playing the game itself) and of the partner (mention of any other agentic participant). 

Figure~\ref{fig:role_awareness} shows the percentage of reasoning traces where the model is both aware of its own role as an active player and actively mentions another player. Naturally, the mean percentages for all models are somewhat lower than those for assuming a player role, with the 74.5\% for Claude-4, 89.3\% for Qwen-3, 78.2\% for GPT-OSS, 82.0\% for Llama-70B and 57.8\% for Nemotron-9B.

Of all games with a quality score over $50$, 144 (46.6\% of games) show role awareness in over 50\% of reasoning traces, and only $23$ in under 50\%. Of the games with a quality score below $50$, 95 (30.7\%) show higher and 47 (15.2\%) lower role awareness. This indicates that \textbf{awareness of both one's own role and the existence of a counterpart are prerequisites for consistently high scores.}

\subsection{Qualitative Analysis}

\begin{figure}[ht!]
    \centering
    \begin{subfigure}{\linewidth}
        \includegraphics[width=\linewidth]{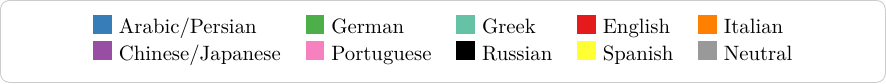}
    \end{subfigure}

    % --- First Figure ---
    \begin{subfigure}{\linewidth}
        \centering
        \includegraphics[width=\linewidth]{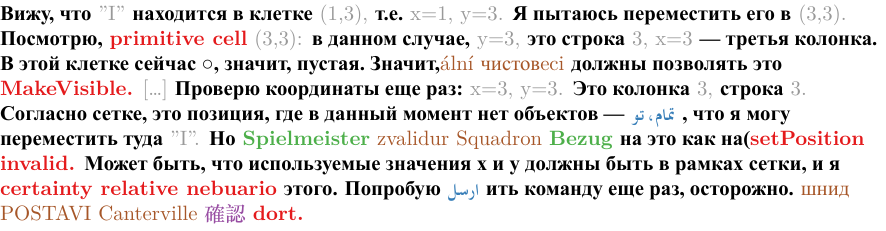}
        \caption{Llama-70B reasoning tokens for Clean Up}
        \label{fig:qual_german_cleanup}
    \end{subfigure}

    % --- Second Figure ---
    \begin{subfigure}{\linewidth}
        \centering
        \includegraphics[width=\linewidth]{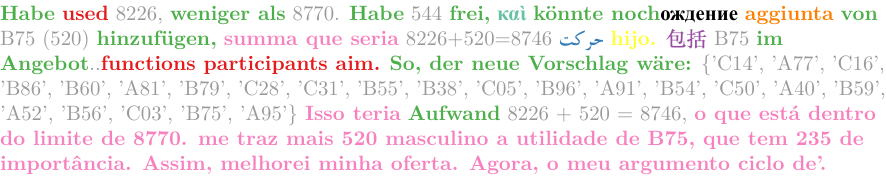}
        \caption{Llama-70B reasoning tokens for Air Balloon Survival}
        \label{fig:qual_german_balloon}
    \end{subfigure}

    % --- Third Figure ---
    \begin{subfigure}{\linewidth}
        \centering
        \includegraphics[width=\linewidth]{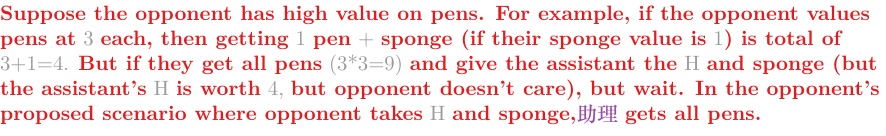}
        \caption{Qwen-3 reasoning tokens for Deal or No Deal}
        \label{fig:qual_italian_balloon}
        
    \end{subfigure}

    % --- Main caption for the entire figure ---
    \caption{Qualitative samples showing reasoning tokens where the target language is German}
    \label{fig:qualitative_examples}
\end{figure}

We provide qualitative samples in Figure~\ref{fig:qualitative_examples} from German task for reasoning tokens generated by \textit{Qwen-3} and \textit{Llama-70B} (distilled from Deepseek-R1). The samples clearly show how the LLMs do not adhere to the target language. The top example shows how languages are interchangeably used, with Russian being the common one. It starts with: \foreignlanguage{russian}{Вижу, что} "I" \foreignlanguage{russian}{находится в клетке (1,3) ... Я пытаюсь переместить его в (3,3). Посмотрю,} primitive cell (3,3) \foreignlanguage{russian}{... В этой клетке сейчас $\circ$, значит, пустая.}, which translates to \textit{I can see that there is ``I'' in cell (1,3) ... and I try to move it  to (3,3) ... I look ... in this cell there is} $\circ$, \textit{meaning it is empty}. 

The middle example shows how the model reasons about its choices and weights and tries to calculate what it means to consider one of them, and interchangeably uses eight languages.

The bottom example's reasoning is about modelling the opponent where the roles are confused because it refers to an ``assistant'' (\begin{CJK*}{UTF8}{gbsn}助理\end{CJK*}). But the actual task is for the assistant to act as the user, and the opponent is simulated.

\section{Conclusion}

We conducted the first comprehensive study systematically investigating the impact of reasoning on performance in multilingual negotiation tasks. 
Our work examines the trade-offs between computational cost and negotiation effectiveness, assesses the linguistic consistency of reasoning processes, and probes the depth of strategic adaptation in English, German, and Italian. Our findings reveal that test-time scaling is a powerful and costly tool: it significantly enhances negotiation performance, yet demands substantial additional compute. 
Open-weight models almost exclusively switch to English for internal reasoning, even when performing tasks in German or Italian, while leading commercial models maintain language consistency between reasoning and outputs. Our results further suggest that reasoning enables genuine strategic adaptation rather than simply pattern matching. The observed improvements in handling complex rules, making value-based decisions, and achieving collaborative outcomes point to deeper problem-solving unlocked when models ``think out loud'' before acting. 
Future work should extend this investigation to broader dialogue games, more diverse languages, and emerging models to chart the path toward AI agents that are versatile negotiators.

\section*{Limitations}

The first limitation is about the scope of negotiation tasks. Some researchers also employed other dialogue games that we did not include in this study, as there are many similarities among games and the measures they assess. Another limitation is the choice of languages for analysis, which are three European languages that are also high-resource languages. The primary reason for considering only the defined dialogue games and languages is that including more games or languages would result in a higher number of model results, thereby increasing the overall cost of the study. Another limitation is the exclusion of models from both commercial and open-source ones. As explained above, such decisions would exceed the budget for such a study. Given the obtained results regarding switching languages, it is evident that most models lack natural reasoning aspects, and utilising such models in real-world applications would bring its own set of challenges and limitations. For use cases that involve cultural or language-specific concepts or aspects, it can not be guaranteed that they will be well-understood in English (as most languages switch to it).

\section*{Ethical Considerations}

Using paid proprietary APIs with underlying models about which little is known (training data, model architecture) in academic research is less than ideal. Currently, the models tested here support reasoning modes, either on or off (except for GPT-OSS-120B and Deepseek-v3.1). We hope that open models will include more controls over reasoning aspects and catch up soon in terms of general performance. Deploying such language agents in society has considerable risks. As mentioned in the Related Work, such agents exhibit asymmetric behaviours in negotiations and may expose users to unfaithful actions.

\section*{Acknowledgments}
We thank all anonymous reviewers who helped improve the manuscript.

\bibliography{custom, anthology-shrunk}

@article{nash1953two,
  title={Two-person cooperative games},
  author={Nash, John},
  journal={Econometrica: Journal of the Econometric Society},
  pages={128--140},
  year={1953},
  publisher={JSTOR}
}

@inproceedings{howes2021justifiable,
  title={Justifiable reasons for everyone: Dialogical reasoning in patients with schizophrenia},
  author={Howes, Christine and Breitholtz, Ellen and Lavelle, Mary and Cooper, Robin},
  booktitle={Proceedings. of the 25th Workshop on the Semantics and Pragmatics of Dialogue},
  year={2021}
}

@book{fisher_getting_2011,
  title = {Getting to Yes: Negotiating Agreement Without Giving In},
  edition = {3rd ed. rev. ed.},
  publisher = {Penguin Books},
  author = {Fisher, Roger and Ury, William and Patton, Bruce},
  year = {2011},
  address = {New York, NY}
}

@book{VonNeumann+Morgenstern:1944,
  author = {von Neumann, John and Morgenstern, Oskar},
  year = {1944},
  title = {Theory of Games and Economic Behavior},
  publisher = {Princeton University Press},
  address = {Princeton, NJ, USA},
  edition = {first}
}

@article{DBLP:journals/corr/abs-2506-00073,
  author       = {Shenzhe Zhu and
                  Jiao Sun and
                  Yi Nian and
                  Tobin South and
                  Alex Pentland and
                  Jiaxin Pei},
  title        = {The Automated but Risky Game: Modeling Agent-to-Agent Negotiations
                  and Transactions in Consumer Markets},
  journal      = {CoRR},
  volume       = {abs/2506.00073},
  year         = {2025},
  url          = {https://doi.org/10.48550/arXiv.2506.00073},
  doi          = {10.48550/ARXIV.2506.00073},
  eprinttype    = {arXiv},
  eprint       = {2506.00073},
  bibsource    = {dblp computer science bibliography, https://dblp.org}
}

@inproceedings{DBLP:conf/acl/Xia0RMZ0024,
  author       = {Tian Xia and
                  Zhiwei He and
                  Tong Ren and
                  Yibo Miao and
                  Zhuosheng Zhang and
                  Yang Yang and
                  Rui Wang},
  editor       = {Lun{-}Wei Ku and
                  Andre Martins and
                  Vivek Srikumar},
  title        = {Measuring Bargaining Abilities of LLMs: {A} Benchmark and {A} Buyer-Enhancement
                  Method},
  booktitle    = {Findings of the Association for Computational Linguistics, {ACL} 2024,
                  Bangkok, Thailand and virtual meeting, August 11-16, 2024},
  pages        = {3579--3602},
  publisher    = {Association for Computational Linguistics},
  year         = {2024},
  url          = {https://doi.org/10.18653/v1/2024.findings-acl.213},
  doi          = {10.18653/V1/2024.FINDINGS-ACL.213},
  bibsource    = {dblp computer science bibliography, https://dblp.org}
}

@inproceedings{DBLP:conf/emnlp/KwonWKCLG24,
  author       = {Deuksin Kwon and
                  Emily Weiss and
                  Tara Kulshrestha and
                  Kushal Chawla and
                  Gale M. Lucas and
                  Jonathan Gratch},
  editor       = {Yaser Al{-}Onaizan and
                  Mohit Bansal and
                  Yun{-}Nung Chen},
  title        = {Are LLMs Effective Negotiators? Systematic Evaluation of the Multifaceted
                  Capabilities of LLMs in Negotiation Dialogues},
  booktitle    = {Findings of the Association for Computational Linguistics: {EMNLP}
                  2024, Miami, Florida, USA, November 12-16, 2024},
  pages        = {5391--5413},
  publisher    = {Association for Computational Linguistics},
  year         = {2024},
  url          = {https://doi.org/10.18653/v1/2024.findings-emnlp.310},
  doi          = {10.18653/V1/2024.FINDINGS-EMNLP.310},
  bibsource    = {dblp computer science bibliography, https://dblp.org}
}

@inproceedings{DBLP:conf/emnlp/ChanJYDF0L0WS24,
  author       = {Chunkit Chan and
                  Cheng Jiayang and
                  Yauwai Yim and
                  Zheye Deng and
                  Wei Fan and
                  Haoran Li and
                  Xin Liu and
                  Hongming Zhang and
                  Weiqi Wang and
                  Yangqiu Song},
  editor       = {Yaser Al{-}Onaizan and
                  Mohit Bansal and
                  Yun{-}Nung Chen},
  title        = {NegotiationToM: {A} Benchmark for Stress-testing Machine Theory of
                  Mind on Negotiation Surrounding},
  booktitle    = {Findings of the Association for Computational Linguistics: {EMNLP}
                  2024, Miami, Florida, USA, November 12-16, 2024},
  pages        = {4211--4241},
  publisher    = {Association for Computational Linguistics},
  year         = {2024},
  url          = {https://doi.org/10.18653/v1/2024.findings-emnlp.244},
  doi          = {10.18653/V1/2024.FINDINGS-EMNLP.244},
  bibsource    = {dblp computer science bibliography, https://dblp.org}
}

@article{
pollo2025re,
title={[Re] Benchmarking {LLM} Capabilities in Negotiation through Scoreable Games},
author={Jorge Carrasco Pollo and Ioannis Kapetangeorgis and Joshua Rosenthal and John Hua Yao},
journal={Transactions on Machine Learning Research},
issn={2835-8856},
year={2025},
url={https://openreview.net/forum?id=BVH81SAAh2},
note={}
}

@inproceedings{
abdelnabi2024llmdeliberation,
title={{LLM}-Deliberation: Evaluating {LLM}s with Interactive Multi-Agent Negotiation Game},
author={Sahar Abdelnabi and Amr Gomaa and Sarath Sivaprasad and Lea Sch{\"o}nherr and Mario Fritz},
booktitle={ICLR 2024 Workshop on Large Language Model (LLM) Agents},
year={2024},
url={https://openreview.net/forum?id=eE1WHn6qlk}
}

@inproceedings{qi-etal-2025-models,
    title = "When Models Reason in Your Language: Controlling Thinking Language Comes at the Cost of Accuracy",
    author = "Qi, Jirui  and
      Chen, Shan  and
      Xiong, Zidi  and
      Fern{\'a}ndez, Raquel  and
      Bitterman, Danielle  and
      Bisazza, Arianna",
    editor = "Christodoulopoulos, Christos  and
      Chakraborty, Tanmoy  and
      Rose, Carolyn  and
      Peng, Violet",
    booktitle = "Findings of the Association for Computational Linguistics: EMNLP 2025",
    month = nov,
    year = "2025",
    address = "Suzhou, China",
    publisher = "Association for Computational Linguistics",
    url = "https://aclanthology.org/2025.findings-emnlp.1103/",
    pages = "20279--20296",
    ISBN = "979-8-89176-335-7"
}

@article{DBLP:journals/corr/abs-2502-09457,
  author       = {Akash Ghosh and
                  Debayan Datta and
                  Sriparna Saha and
                  Chirag Agarwal},
  title        = {The Multilingual Mind : {A} Survey of Multilingual Reasoning in Language
                  Models},
  journal      = {CoRR},
  volume       = {abs/2502.09457},
  year         = {2025},
  url          = {https://doi.org/10.48550/arXiv.2502.09457},
  doi          = {10.48550/ARXIV.2502.09457},
  eprinttype    = {arXiv},
  eprint       = {2502.09457},
  bibsource    = {dblp computer science bibliography, https://dblp.org}
}

@inproceedings{DBLP:conf/iclr/CaoLLLTC18,
  author       = {Kris Cao and
                  Angeliki Lazaridou and
                  Marc Lanctot and
                  Joel Z. Leibo and
                  Karl Tuyls and
                  Stephen Clark},
  title        = {Emergent Communication through Negotiation},
  booktitle    = {6th International Conference on Learning Representations, {ICLR} 2018,
                  Vancouver, BC, Canada, April 30 - May 3, 2018, Conference Track Proceedings},
  publisher    = {OpenReview.net},
  year         = {2018},
  url          = {https://openreview.net/forum?id=Hk6WhagRW},
  bibsource    = {dblp computer science bibliography, https://dblp.org}
}

@article{DBLP:journals/corr/abs-2305-19165,
  author       = {Kanishk Gandhi and
                  Dorsa Sadigh and
                  Noah D. Goodman},
  title        = {Strategic Reasoning with Language Models},
  journal      = {CoRR},
  volume       = {abs/2305.19165},
  year         = {2023},
  url          = {https://doi.org/10.48550/arXiv.2305.19165},
  doi          = {10.48550/ARXIV.2305.19165},
  eprinttype    = {arXiv},
  eprint       = {2305.19165},
  bibsource    = {dblp computer science bibliography, https://dblp.org}
}

@article{DBLP:journals/corr/abs-2305-16867,
  author       = {Elif Akata and
                  Lion Schulz and
                  Julian Coda{-}Forno and
                  Seong Joon Oh and
                  Matthias Bethge and
                  Eric Schulz},
  title        = {Playing repeated games with Large Language Models},
  journal      = {CoRR},
  volume       = {abs/2305.16867},
  year         = {2023},
  url          = {https://doi.org/10.48550/arXiv.2305.16867},
  doi          = {10.48550/ARXIV.2305.16867},
  eprinttype    = {arXiv},
  eprint       = {2305.16867},
  bibsource    = {dblp computer science bibliography, https://dblp.org}
}

@article{DBLP:journals/corr/abs-2411-05990,
  author       = {Wenyue Hua and
                  Ollie Liu and
                  Lingyao Li and
                  Alfonso Amayuelas and
                  Julie Chen and
                  Lucas Jiang and
                  Mingyu Jin and
                  Lizhou Fan and
                  Fei Sun and
                  William Wang and
                  Xintong Wang and
                  Yongfeng Zhang},
  title        = {Game-theoretic {LLM:} Agent Workflow for Negotiation Games},
  journal      = {CoRR},
  year         = {2024},
  eprinttype    = {arXiv},
  eprint       = {2411.05990},
  bibsource    = {dblp computer science bibliography, https://dblp.org}
}

@misc{wong2025reasoningcapabilitieslargelanguage,
      title={Reasoning Capabilities of Large Language Models on Dynamic Tasks}, 
      author={Annie Wong and Thomas Bäck and Aske Plaat and Niki van Stein and Anna V. Kononova},
      year={2025},
      eprint={2505.10543},
      archivePrefix={arXiv},
      primaryClass={cs.AI},
      url={https://arxiv.org/abs/2505.10543}, 
}

@inproceedings{DBLP:conf/iclr/DavidsonVK024,
  author       = {Tim R. Davidson and
                  Veniamin Veselovsky and
                  Michal Kosinski and
                  Robert West},
  title        = {Evaluating Language Model Agency Through Negotiations},
  booktitle    = {The Twelfth International Conference on Learning Representations,
                  {ICLR} 2024, Vienna, Austria, May 7-11, 2024},
  publisher    = {OpenReview.net},
  year         = {2024},
  url          = {https://openreview.net/forum?id=3ZqKxMHcAg},
  bibsource    = {dblp computer science bibliography, https://dblp.org}
}

@inproceedings{DBLP:conf/icml/0001CYTJ024,
  author       = {Federico Bianchi and
                  Patrick John Chia and
                  Mert Y{\"{u}}ksekg{\"{o}}n{\"{u}}l and
                  Jacopo Tagliabue and
                  Dan Jurafsky and
                  James Zou},
  title        = {How Well Can LLMs Negotiate? NegotiationArena Platform and Analysis},
  booktitle    = {Forty-first International Conference on Machine Learning, {ICML} 2024,
                  Vienna, Austria, July 21-27, 2024},
  publisher    = {OpenReview.net},
  year         = {2024},
  url          = {https://openreview.net/forum?id=CmOmaxkt8p},
  bibsource    = {dblp computer science bibliography, https://dblp.org}
}

@article{DBLP:journals/corr/abs-2011-01975,
  author       = {Dhruv Batra and
                  Angel X. Chang and
                  Sonia Chernova and
                  Andrew J. Davison and
                  Jia Deng and
                  Vladlen Koltun and
                  Sergey Levine and
                  Jitendra Malik and
                  Igor Mordatch and
                  Roozbeh Mottaghi and
                  Manolis Savva and
                  Hao Su},
  title        = {Rearrangement: {A} Challenge for Embodied {AI}},
  journal      = {CoRR},
  volume       = {abs/2011.01975},
  year         = {2020},
  url          = {https://arxiv.org/abs/2011.01975},
  eprinttype    = {arXiv},
  eprint       = {2011.01975},
  bibsource    = {dblp computer science bibliography, https://dblp.org}
}

@inproceedings{DBLP:conf/sigdial/JeknicSK24,
  author       = {Isidora Jeknic and
                  David Schlangen and
                  Alexander Koller},
  editor       = {Tatsuya Kawahara and
                  Vera Demberg and
                  Stefan Ultes and
                  Koji Inoue and
                  Shikib Mehri and
                  David M. Howcroft and
                  Kazunori Komatani},
  title        = {A Dialogue Game for Eliciting Balanced Collaboration},
  booktitle    = {Proceedings of the 25th Annual Meeting of the Special Interest Group
                  on Discourse and Dialogue, {SIGDIAL} 2024, Kyoto, Japan, September
                  18 - 20, 2024},
  pages        = {477--489},
  publisher    = {Association for Computational Linguistics},
  year         = {2024},
  url          = {https://doi.org/10.18653/v1/2024.sigdial-1.41},
  doi          = {10.18653/V1/2024.SIGDIAL-1.41},
  bibsource    = {dblp computer science bibliography, https://dblp.org}
}

@misc{sun2025gametheorymeetslarge,
      title={Game Theory Meets Large Language Models: A Systematic Survey with Taxonomy and New Frontiers}, 
      author={Haoran Sun and Yusen Wu and Peng Wang and Wei Chen and Yukun Cheng and Xiaotie Deng and Xu Chu},
      year={2025},
      eprint={2502.09053},
      archivePrefix={arXiv},
      primaryClass={cs.AI},
      url={https://arxiv.org/abs/2502.09053}, 
}

@article{DBLP:journals/corr/abs-2503-06416,
  author       = {Michelle Vaccaro and
                  Michael Caoson and
                  Harang Ju and
                  Sinan Aral and
                  Jared R. Curhan},
  title        = {Advancing {AI} Negotiations: New Theory and Evidence from a Large-Scale
                  Autonomous Negotiations Competition},
  journal      = {CoRR},
  volume       = {abs/2503.06416},
  year         = {2025},
  url          = {https://doi.org/10.48550/arXiv.2503.06416},
  doi          = {10.48550/ARXIV.2503.06416},
  eprinttype    = {arXiv},
  eprint       = {2503.06416},
  bibsource    = {dblp computer science bibliography, https://dblp.org}
}

@article{DBLP:journals/corr/abs-2402-01704,
  author       = {Ian Gemp and
                  Yoram Bachrach and
                  Marc Lanctot and
                  Roma Patel and
                  Vibhavari Dasagi and
                  Luke Marris and
                  Georgios Piliouras and
                  Siqi Liu and
                  Karl Tuyls},
  title        = {States as Strings as Strategies: Steering Language Models with Game-Theoretic
                  Solvers},
  journal      = {Agentic Markets Workshop at ICML 2024},
  year         = {2024},
  bibsource    = {dblp computer science bibliography, https://dblp.org}
}

@article{DBLP:journals/ral/ZengWYZDCD24,
  author       = {Yiming Zeng and
                  Mingdong Wu and
                  Long Yang and
                  Jiyao Zhang and
                  Hao Ding and
                  Hui Cheng and
                  Hao Dong},
  title        = {LVDiffusor: Distilling Functional Rearrangement Priors From Large
                  Models Into Diffusor},
  journal      = {{IEEE} Robotics Autom. Lett.},
  volume       = {9},
  number       = {10},
  pages        = {8258--8265},
  year         = {2024},
  url          = {https://doi.org/10.1109/LRA.2024.3438036},
  doi          = {10.1109/LRA.2024.3438036},
  bibsource    = {dblp computer science bibliography, https://dblp.org}
}

@article{DBLP:journals/access/KhanQCASR25,
  author       = {Mariia Khan and
                  Yue Qiu and
                  Yuren Cong and
                  Jumana Abu{-}Khalaf and
                  David Suter and
                  Bodo Rosenhahn},
  title        = {PanoSCU: {A} Simulation-Based Dataset for Panoramic Indoor Scene Understanding},
  journal      = {{IEEE} Access},
  volume       = {13},
  pages        = {72456--72476},
  year         = {2025},
  url          = {https://doi.org/10.1109/ACCESS.2025.3561055},
  doi          = {10.1109/ACCESS.2025.3561055},
  bibsource    = {dblp computer science bibliography, https://dblp.org}
}

@article{DBLP:journals/corr/abs-2506-06941,
  author       = {Parshin Shojaee and
                  Iman Mirzadeh and
                  Keivan Alizadeh and
                  Maxwell Horton and
                  Samy Bengio and
                  Mehrdad Farajtabar},
  title        = {The Illusion of Thinking: Understanding the Strengths and Limitations
                  of Reasoning Models via the Lens of Problem Complexity},
  journal      = {CoRR},
  volume       = {abs/2506.06941},
  year         = {2025},
  url          = {https://doi.org/10.48550/arXiv.2506.06941},
  doi          = {10.48550/ARXIV.2506.06941},
  eprinttype    = {arXiv},
  eprint       = {2506.06941},
  bibsource    = {dblp computer science bibliography, https://dblp.org}
}

\appendix

\section{Additional Results}\label{sec:appendix_reamining_details}

\subsection{\%Played and Quality Scores}

\begin{figure}[ht!]
    \centering
    
    % --- English Figure ---
    \begin{subfigure}{0.8\linewidth}
        \centering
        \includegraphics[width=\linewidth]{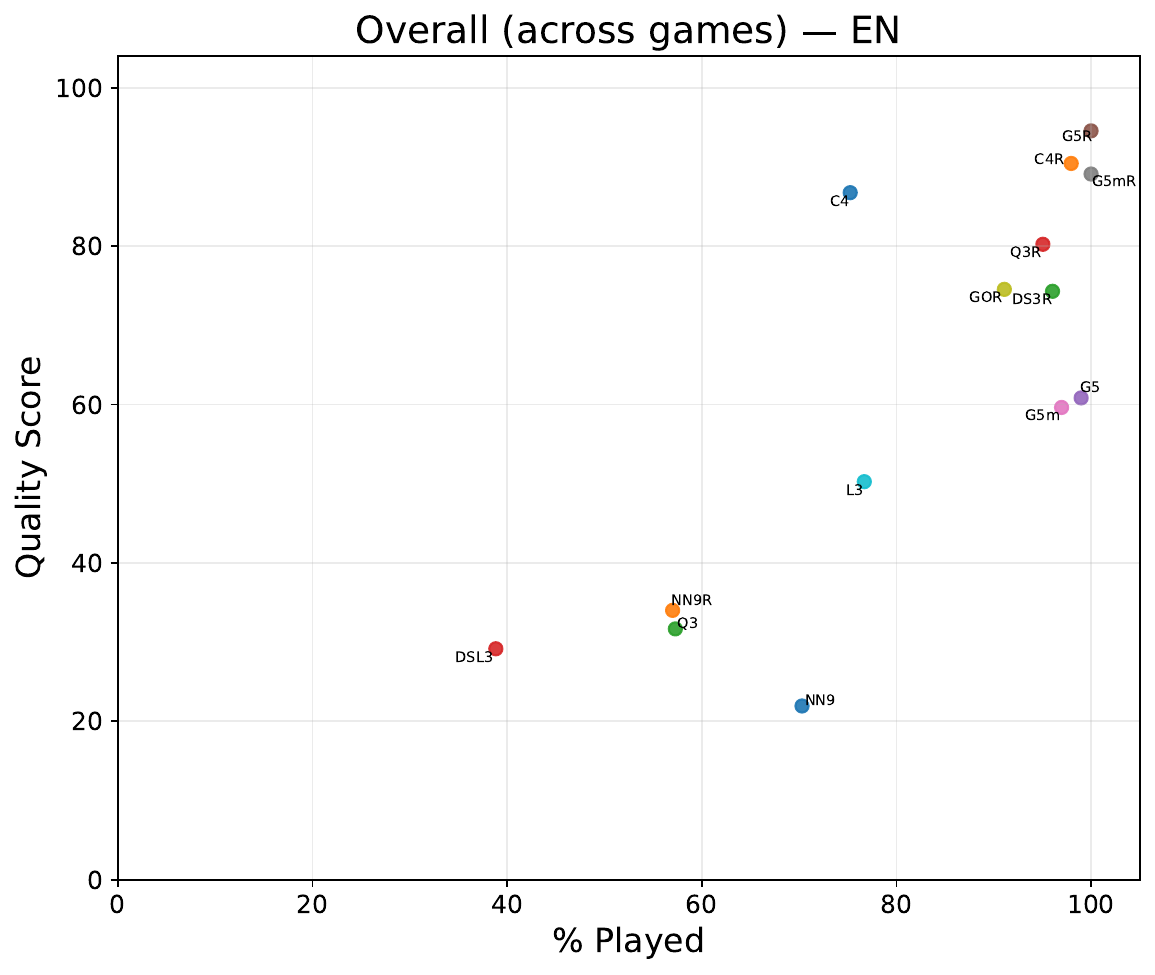}
        \caption{Played-Quality Scores for English averaged for all games.}
        \label{fig:scatter_en}
    \end{subfigure}
    
    \vspace{1.5em} % Adjust this value to increase/decrease the space between plots

    % --- German Figure ---
    \begin{subfigure}{0.8\linewidth}
        \centering
        \includegraphics[width=\linewidth]{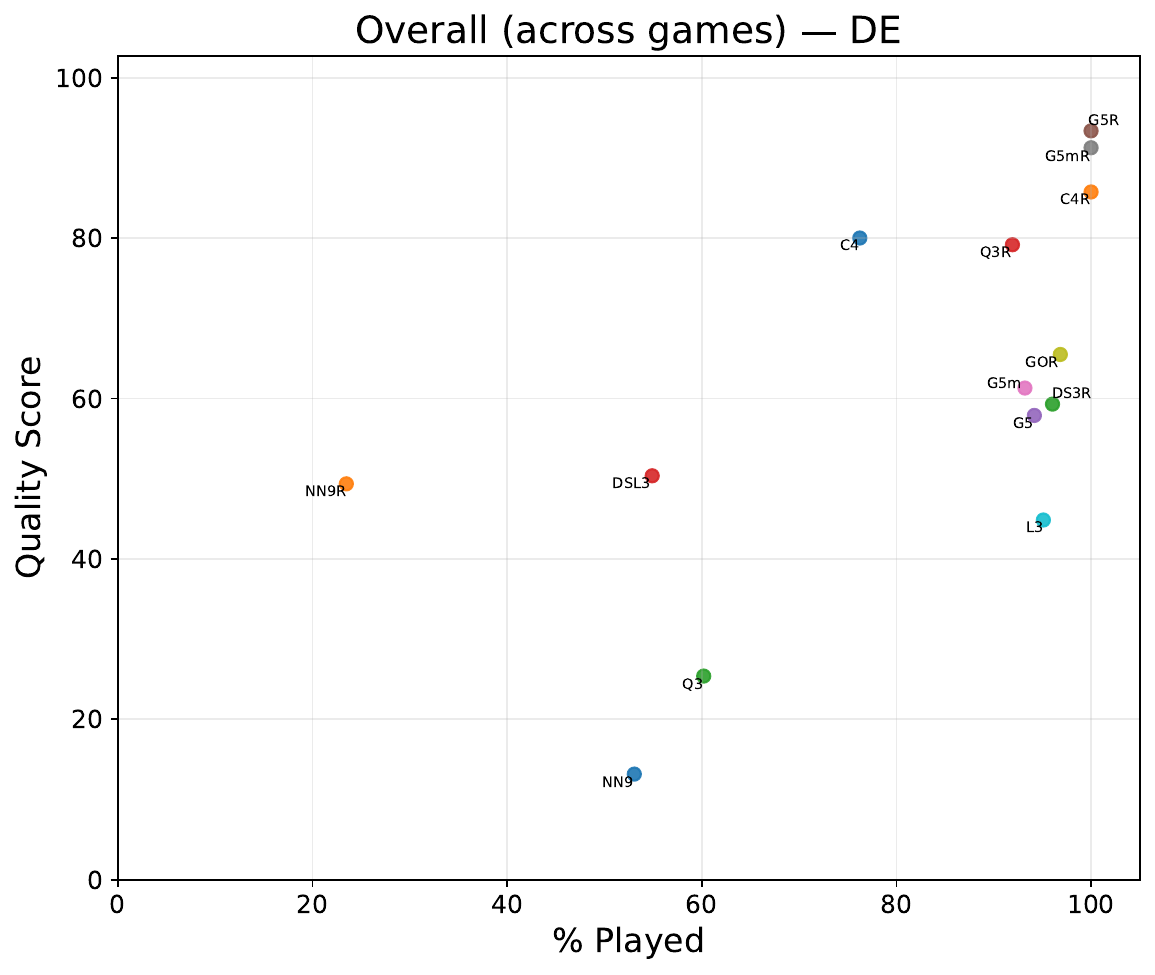}
        \caption{Played-Quality Scores for German averaged for all games.}
        \label{fig:scatter_de}
    \end{subfigure}

    \vspace{1.5em} % Adjust this value to increase/decrease the space between plots

    % --- Italian Figure ---
    \begin{subfigure}{0.8\linewidth}
        \centering
        \includegraphics[width=\linewidth]{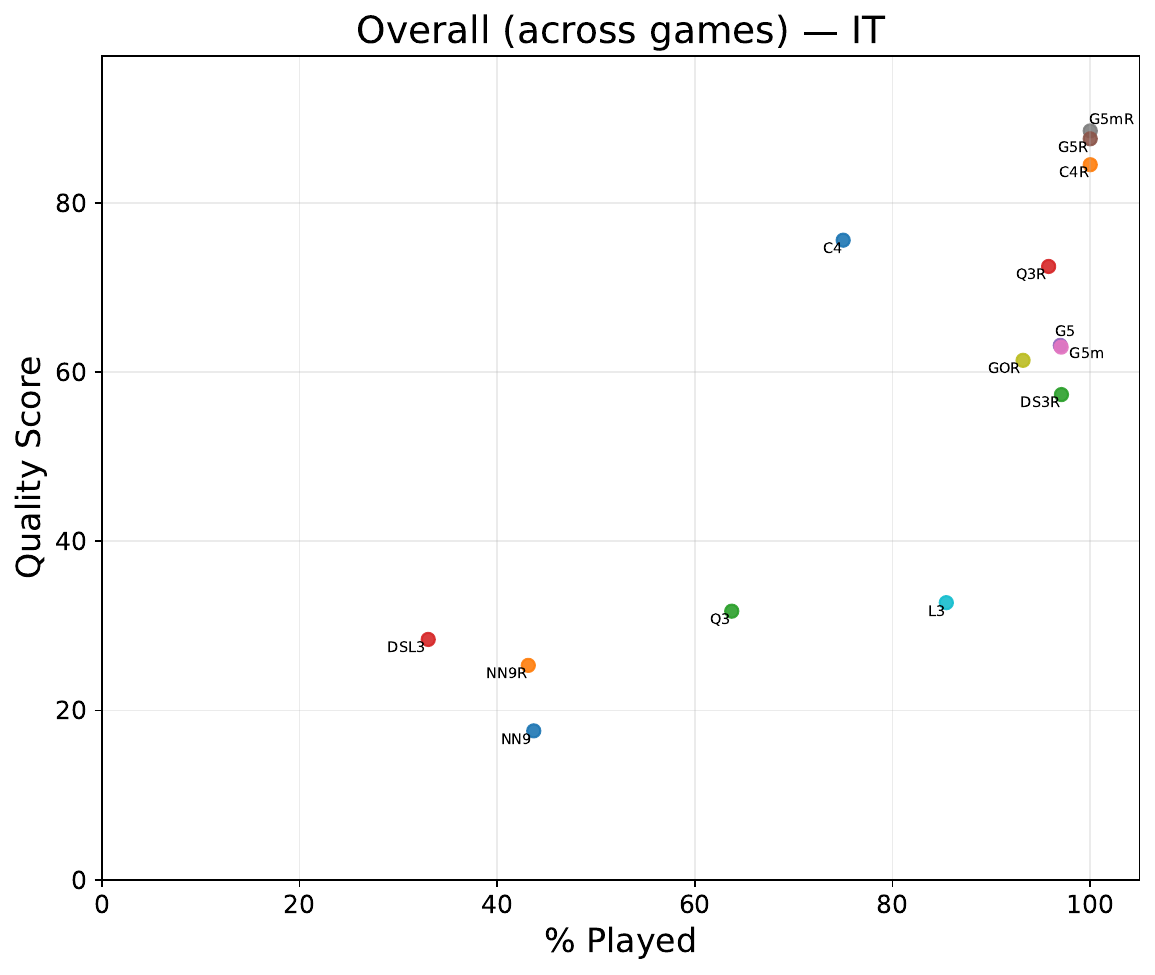}
        \caption{Played-Quality Scores for Italian averaged for all games.}
        \label{fig:scatter_it}
    \end{subfigure}

    % --- Optional: Add an overall caption for the entire figure ---
    \caption{Comparison of Played-Quality Scores averaged across all games for English, German, and Italian.}
    \label{fig:scatter_all_languages}
\end{figure}

Figure~\ref{fig:scatter_all_languages} shows the Played and Quality Scores for each language separately, averaged for all games.

\subsection{Token Usage \& Cost}

\begin{figure*}[t!]
    \centering
    % --- English Figures (Top Row) ---
    \begin{subfigure}[b]{0.49\textwidth}
        \centering
        \includegraphics[width=1.0\linewidth]{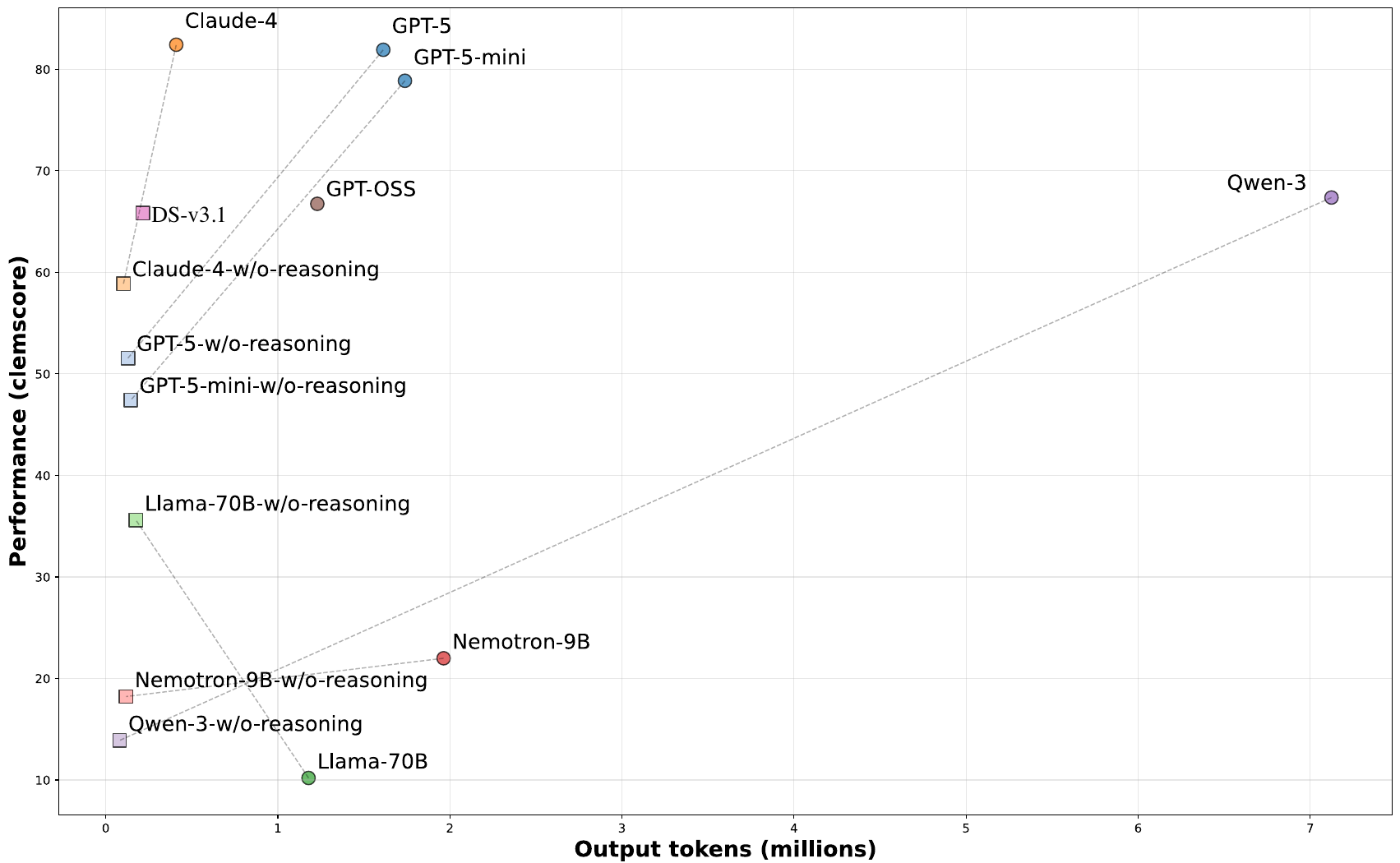}
        \caption{Performance and output tokens for all evaluated models for English.}
        \label{fig:token_en}
    \end{subfigure}
    \hfill
    \begin{subfigure}[b]{0.49\textwidth}
        \centering
        \includegraphics[width=1.0\linewidth]{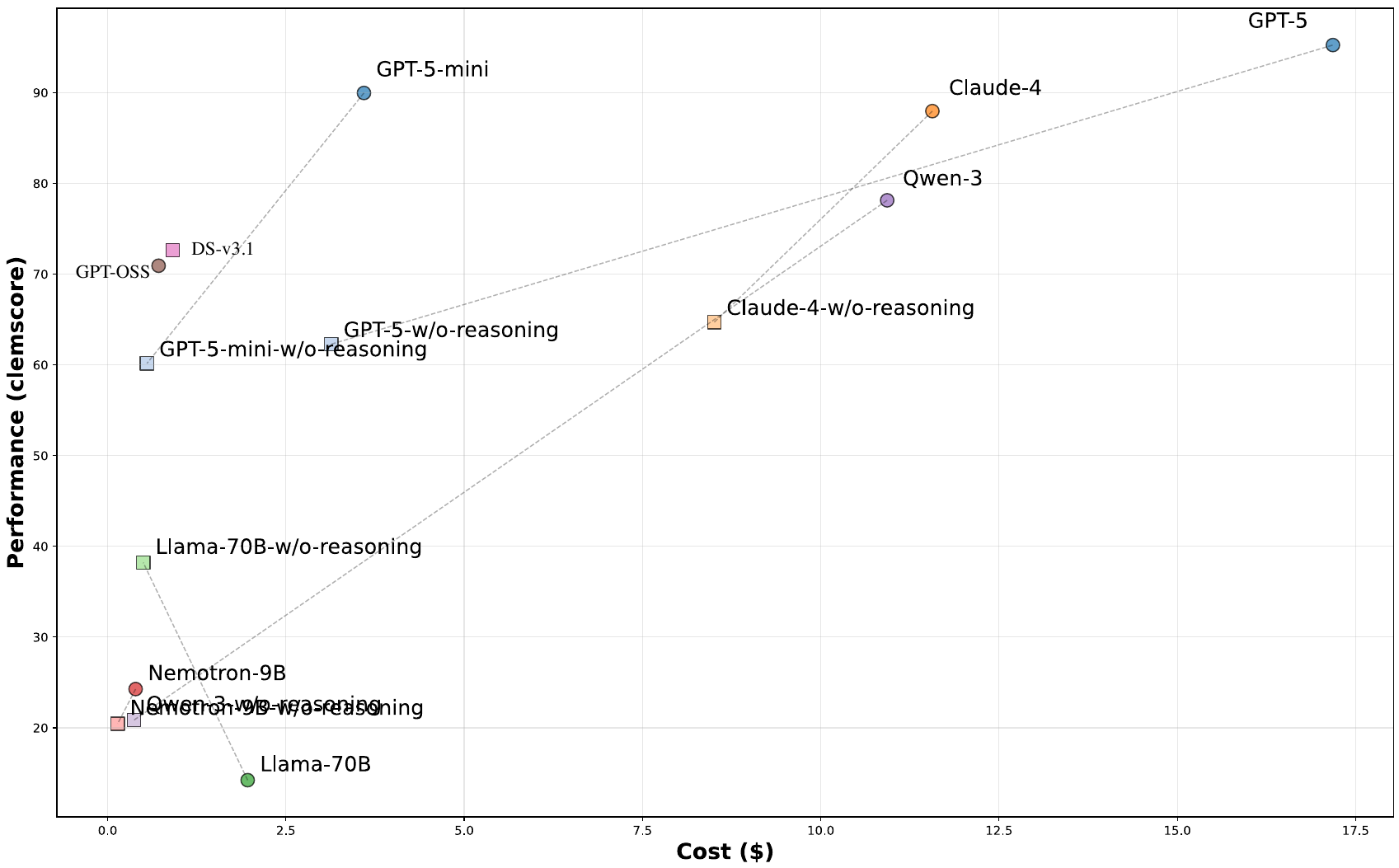}
        \caption{Performance (clemscore) and cost for all evaluated models for English.}
        \label{fig:cost_en}
    \end{subfigure}
    
    \vspace{1em} % Adds vertical space between the rows

    % --- German Figures (Middle Row) ---
    \begin{subfigure}[b]{0.49\textwidth}
        \centering
        \includegraphics[width=1.0\linewidth]{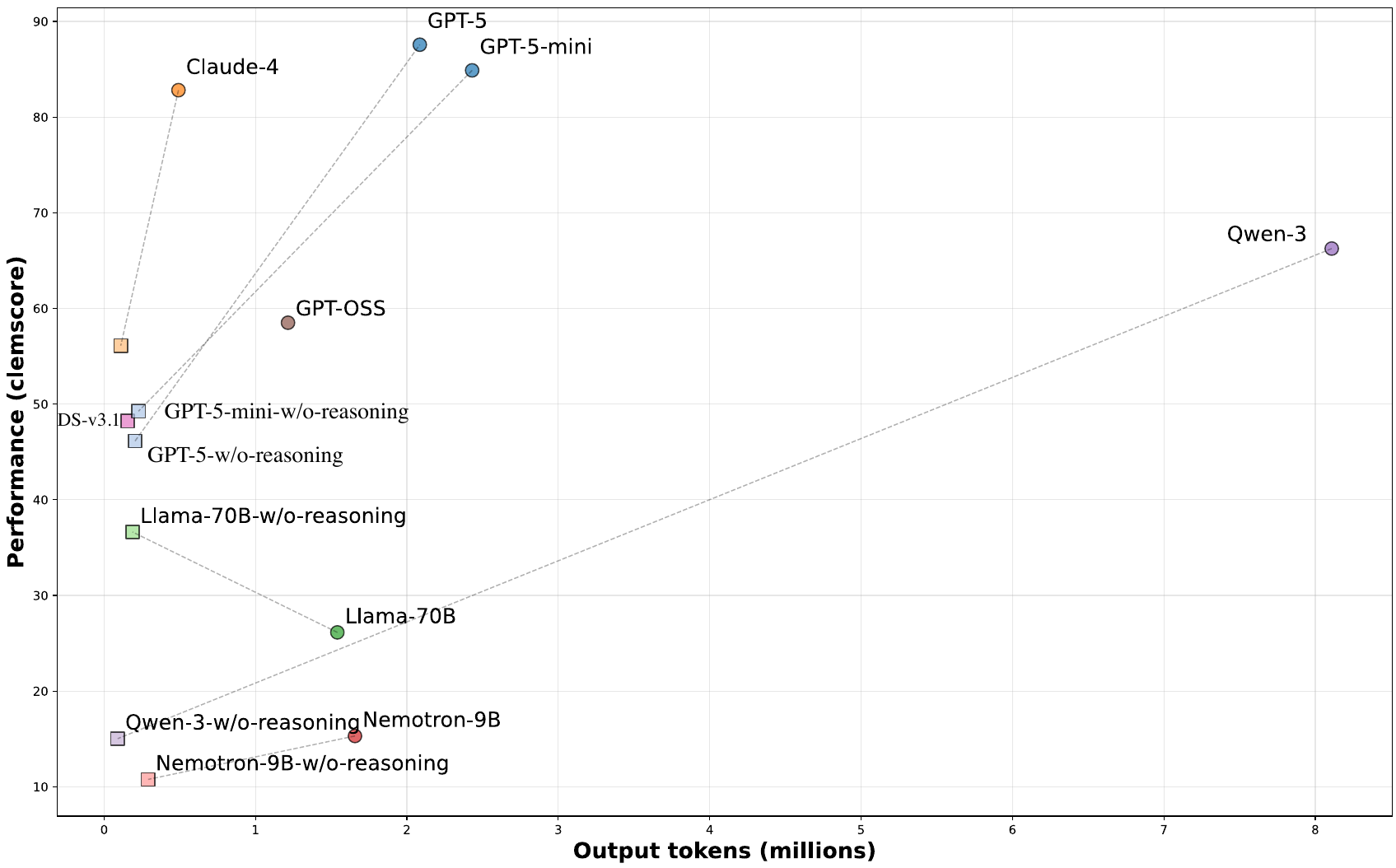}
        \caption{Performance (clemscore) and output tokens for all evaluated models for German.}
        \label{fig:token_de}
    \end{subfigure}
    \hfill
    \begin{subfigure}[b]{0.49\textwidth}
        \centering
        \includegraphics[width=1.0\linewidth]{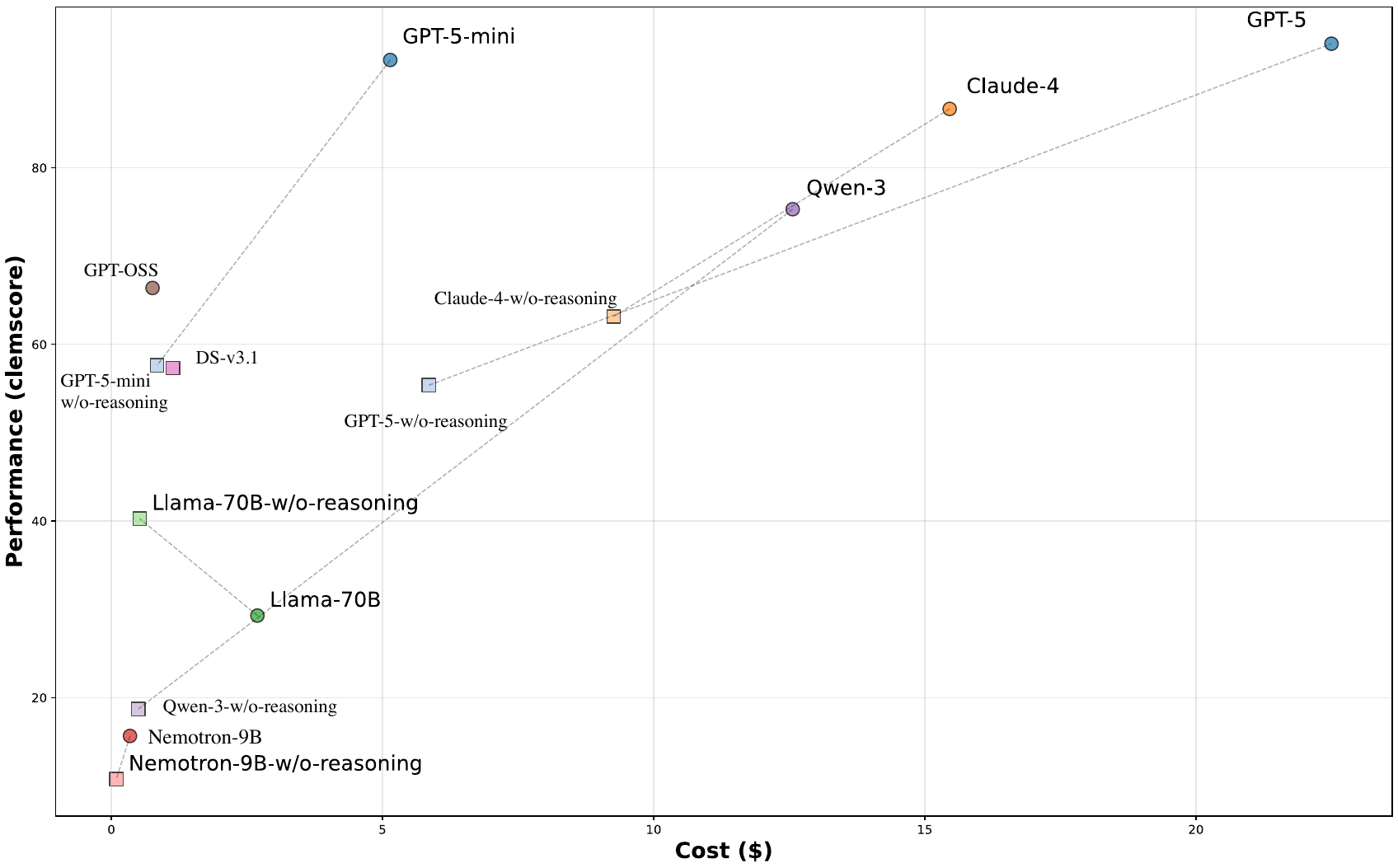}
        \caption{Performance (clemscore) and cost for all evaluated models for German.}
        \label{fig:cost_de}
    \end{subfigure}

    \vspace{1em} % Adds vertical space between the rows

    % --- Italian Figures (Bottom Row) ---
    \begin{subfigure}[b]{0.49\textwidth}
        \centering
        \includegraphics[width=1.0\linewidth]{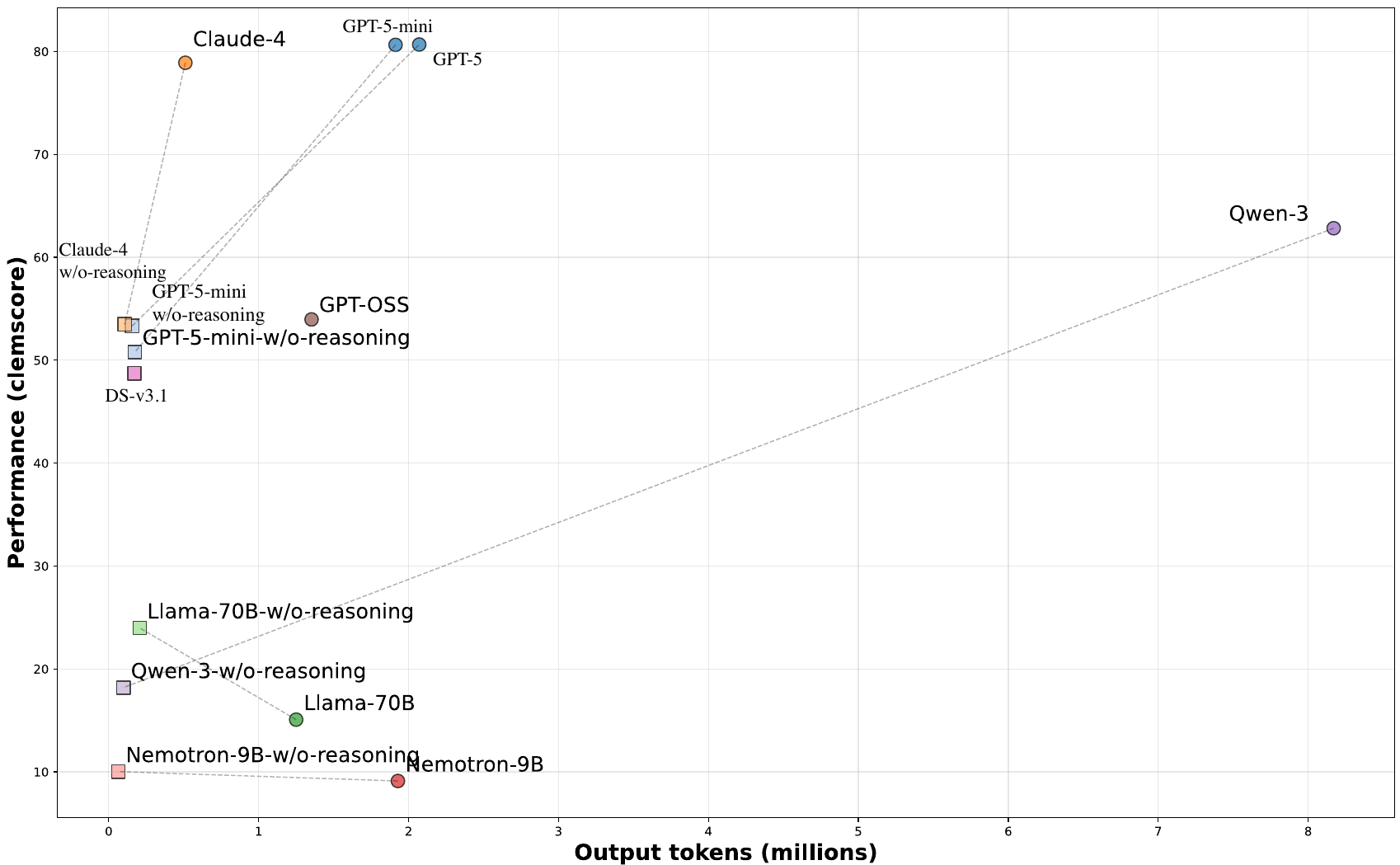}
        \caption{Performance (clemscore) and output tokens for all evaluated models for Italian.}
        \label{fig:token_it}
    \end{subfigure}
    \hfill
    \begin{subfigure}[b]{0.49\textwidth}
        \centering
        \includegraphics[width=1.0\linewidth]{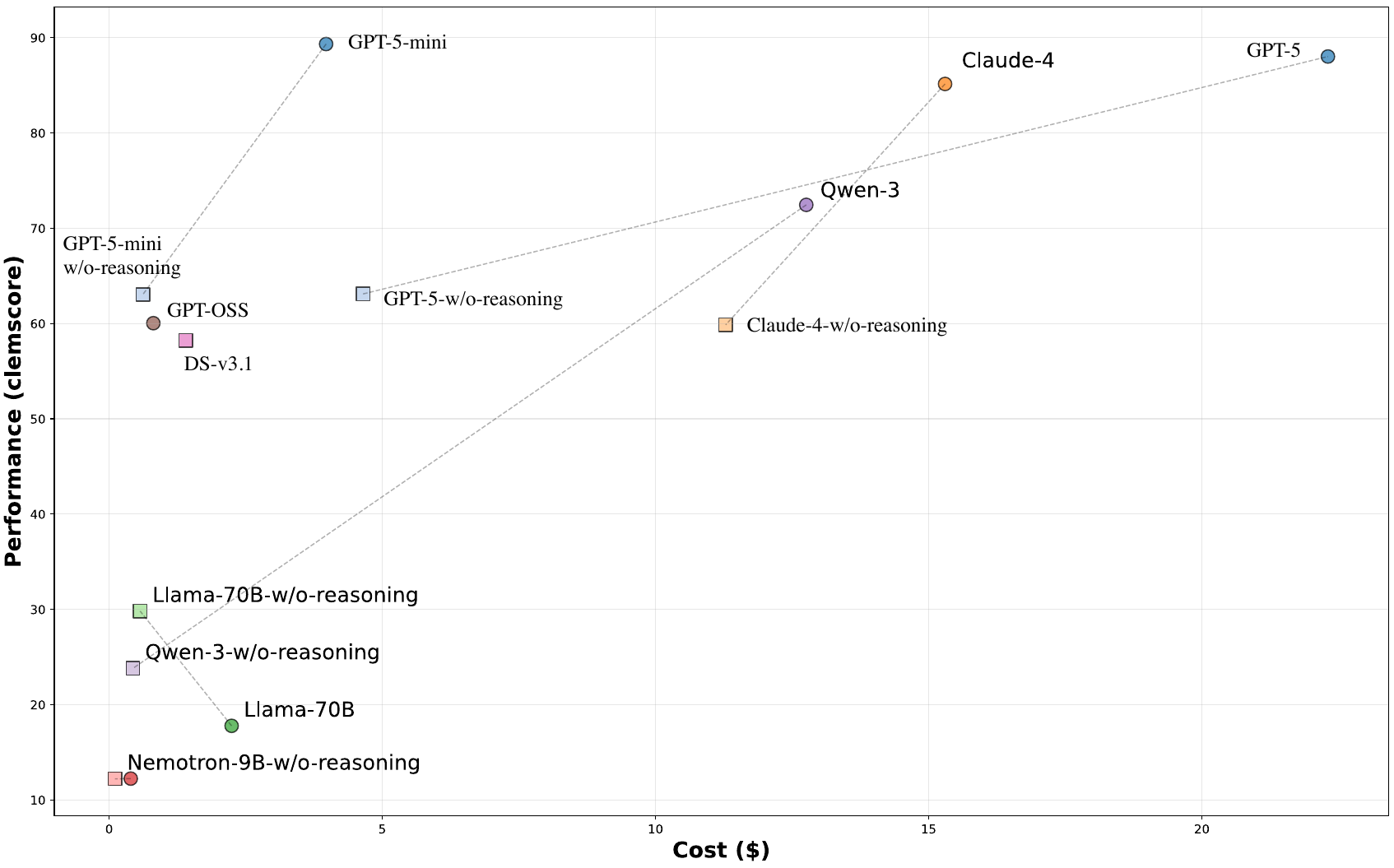}
        \caption{Performance (clemscore) and cost for all evaluated models for Italian.}
        \label{fig:cost_it}
    \end{subfigure}
    
    % --- Unified Main Caption ---
    \caption{Trade-off between performance and cost comparison across different models. Results for English (top), German (middle), and Italian (bottom).}
    \label{fig:all-results}
\end{figure*}

The average token usage and cost for English, German and Italian experiments are given in Figure~\ref{fig:all-results}.

\subsection{Cycles in Reasoning Traces}
\label{sec:cyclic_analysis}

\subsubsection{Definition}

\begin{figure*}[!t]
  \centering
  \includegraphics[width=\textwidth]{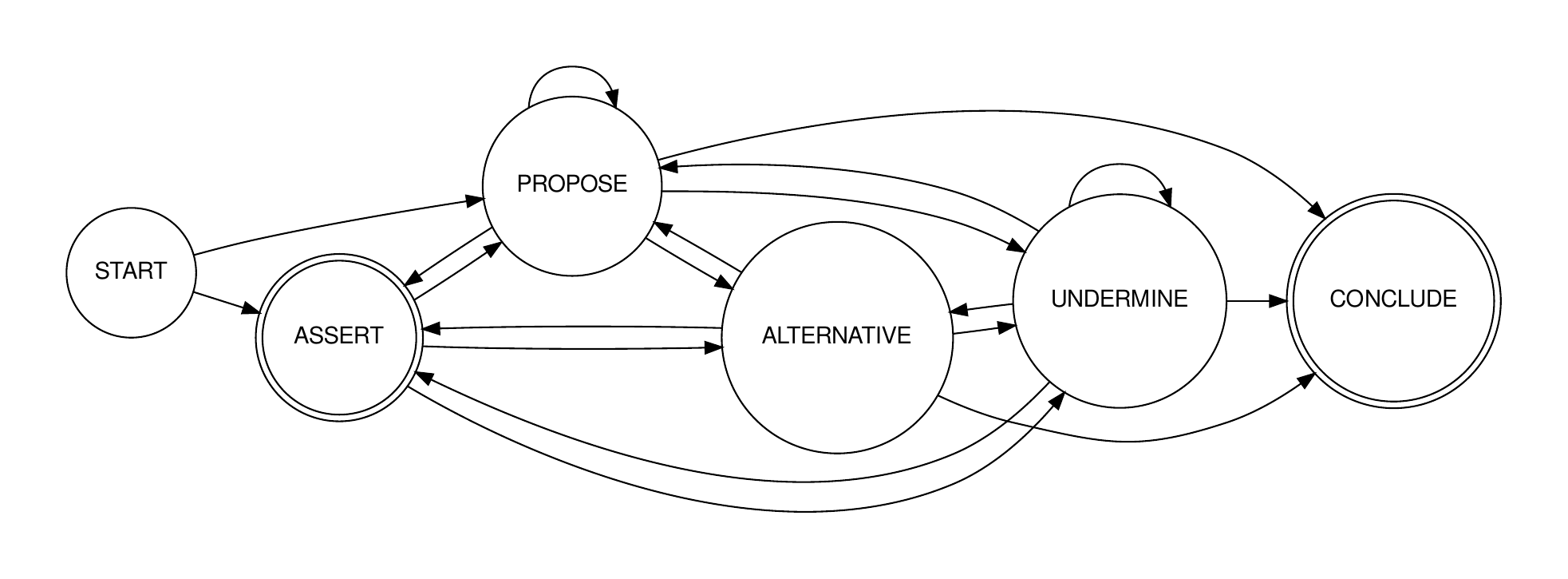}
  \caption{Finite state machine states for reasoning traces}
  \label{fig:fsm}
\end{figure*}

We label the reasoning traces as sequences over five discrete states—\texttt{ASSERT}, \texttt{PROPOSE}, \texttt{UNDERMINE}, \texttt{ALTERNATIVE}, \texttt{CONCLUDE}—detected via lightweight lexical cues and conflict resolution rules (e.g., maybe/could → \texttt{PROPOSE}; but/however → \texttt{UNDERMINE}; \texttt{PROPOSE} shadows other labels co-occuring in one sentence, sentences like "but maybe we should" will be labeled as \texttt{PROPOSE}), the complete matching rules and conflict resolution rules are in Table ~\ref{tab:lexical_cues} and Table ~\ref{tab:conflict_rules}. Each trace is a path on a finite-state machine (FSM, see Figure ~\ref{fig:fsm}) with observed transitions (including self-loops), from which we compute per-trace statistics and aggregate at model/game level.

We treat a trace as a directed walk over labels and compute two cycle-aware metrics:

\paragraph{Number of segments.} We partition the labels extracted from each reasoning trace by absorbing states (\texttt{ASSERT} and  \texttt{CONCLUDE}). The count serves as coarse “reasoning units”: higher counts could mean more dispersed reasoning (multiple restarts before final commitment); lower counts could mean more concentrated reasoning (a plan developed and resolved within fewer units).

\paragraph{Cycle edge ratio.}\label{def:cycle_edge_ratio}
Within each segment’s induced subgraph, we mark edges that belong to any simple cycle and compute the fraction of transitions on such edges, then aggregate across segments to obtain one single metric for each reasoning trace. Higher ratios indicate local looping/hedging (e.g., \texttt{PROPOSE} <-> \texttt{UNDERMINE}); lower ratios indicate forward motion towards absorbing states.

These metrics capture cyclicity at two levels: inter-segment shifting/restarts (segments) before final commitment, and intra-segment circling before interim resolution (cycle edge ratio). Strategically adaptive behaviours should advance and commit—e.g., \texttt{PROPOSE} → \texttt{ASSERT} → \texttt{CONCLUDE} or \texttt{PROPOSE} → \texttt{ALTERNATIVE} → \texttt{ASSERT} → \texttt{CONCLUDE}—rather than cyclic hedges like \texttt{PROPOSE} <-> \texttt{UNDERMINE} <-> \texttt{ALTERNATIVE}. Thus, high cyclicity signals surface patterning (stalling/hedging/“explainers that never decide”).

We relate both metrics to clemscore across games and languages to test whether low number of reasoning segments and low-cyclic reasoning aligns with better negotiation outcomes.

\subsubsection{Analysis}

\begin{figure}[ht!] 
    \centering 
    \includegraphics[width=1.0\linewidth]{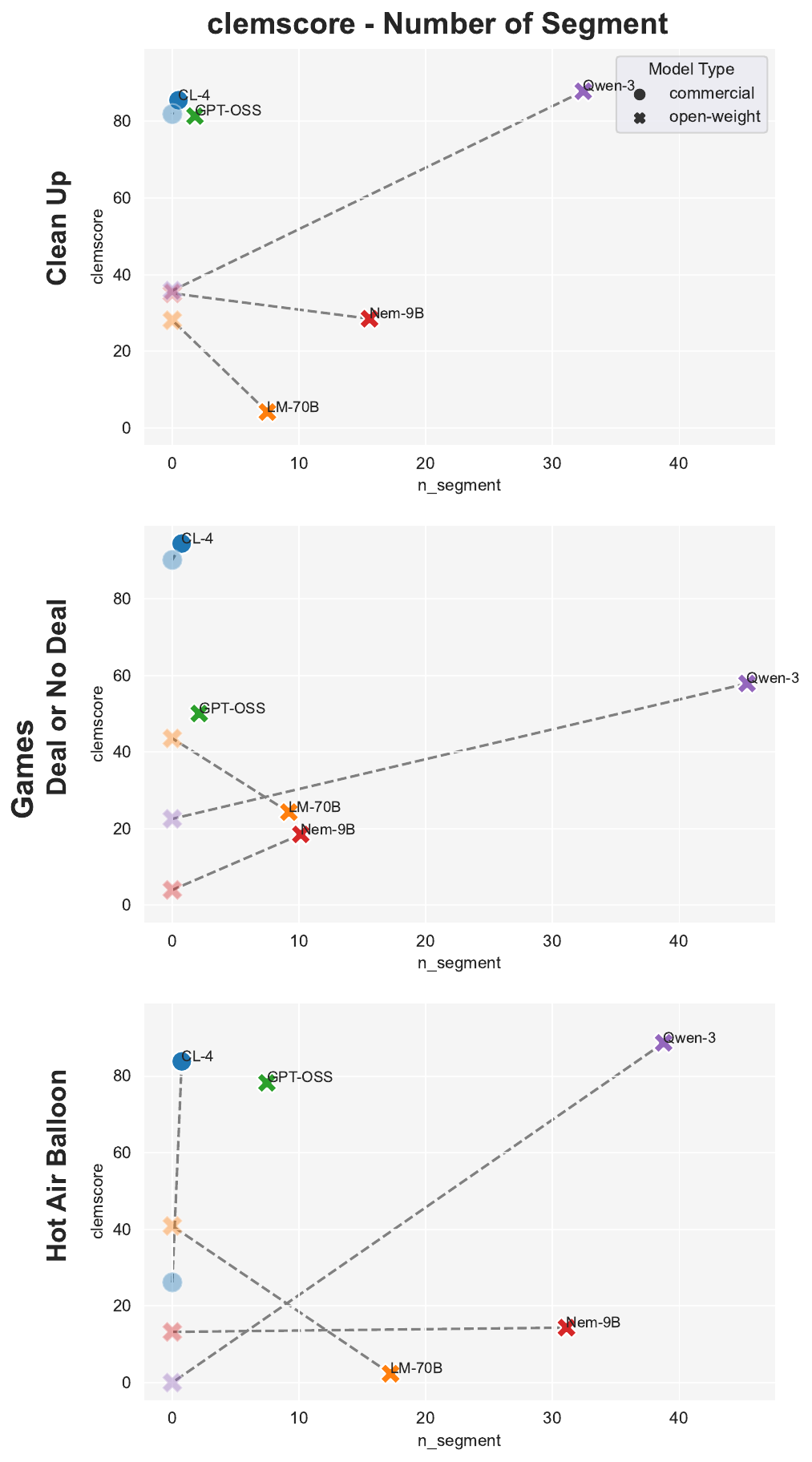} 
    \caption{Clemscore plotted against number of Segment. A Segment is defined as a continuous chunk of a series of reasoning labels extracted from reasoning traces. It is obtained by partitioning the reasoning labels with absorbing states in Figure~\ref{fig:fsm}. Solid markers denote reasoning models; translucent markers denote their non-reasoning counterparts; dashed lines connect each pair.} \label{fig:clemscore_VS_n_segment} 
\end{figure}

\textbf{Number of Segments}
We compare paired non-reasoning and reasoning variants of models across games, plotting clemscore (y) against the number of segments (x) in Figure ~\ref{fig:clemscore_VS_n_segment}. There is a slight U-shaped relationship: both a small increase and an excessive increase in the number of segments coincide with improved performance, while a moderate addition of segments shows unclear correlation. Claude-4 adds only a few segments and shows small gains in \textit{Clean Up} and \textit{Deal or No Deal} and a pronounced gain in \textit{Air Balloon Survival}. Deepseek exhibits moderate segment inflation with consistently lower clemscores. Nemotron shows mixed effects at similar segment increases. Qwen introduces the largest segment increases and performance gains; its performance is competitive with Claude in two of the three games, but in Deal or No Deal its substantial over-segmentation coincides with a relative underperformance.
We compare paired non-reasoning and reasoning variants of models across games, plotting clemscore (y) against the number of segments (x) in Figure ~\ref{fig:clemscore_VS_n_segment}. There is a slight U-shaped relationship: both a small increase and an excessive increase in the number of segments coincide with improved performance, while a moderate addition of segments shows unclear correlation. Claude-4 adds only a few segments and shows small gains in \textit{Clean Up} and \textit{Deal or No Deal} and a pronounced gain in \textit{Air Balloon Survival}. Deepseek exhibits moderate segment inflation with consistently lower clemscores. Nemotron shows mixed effects at similar segment increases. Qwen introduces the largest segment increases and performance gains; its performance is competitive with Claude in two of the three games, but in Deal or No Deal its substantial over-segmentation coincides with a relative underperformance.

\begin{figure}[ht!] 
    \centering 
    \includegraphics[width=1.0\linewidth]{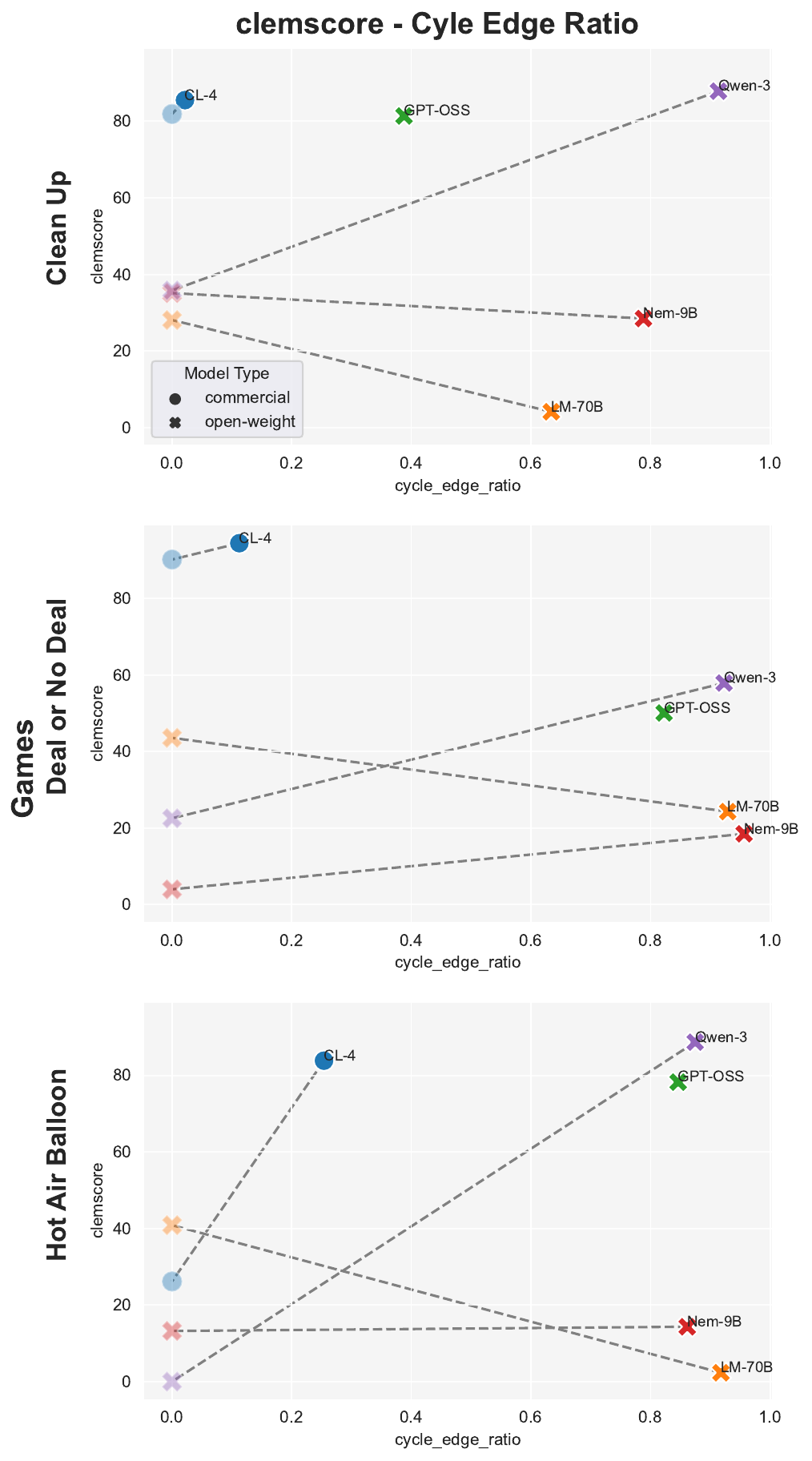} 
    \caption{Clemscore plotted against Cycle Edge Ratio, defined in  Appendix~\ref{def:cycle_edge_ratio}. Solid markers denote reasoning models; translucent markers denote their non-reasoning counterparts; dashed lines connect each pair.} \label{fig:clemscore_VS_cycle_edge_ratio} 
\end{figure}

\textbf{Cycle Edge Ratio}
The relation between clemscore and Cycle Edge Ratio is shown in Figure ~\ref{fig:clemscore_VS_cycle_edge_ratio}.
Across panels, open-weight models tend to have a much higher cycle edge ratio, indicating frequent cyclic hedging before reaching interim conclusions or assertions. Although there is no uniform monotonic link between cycle-edge ratio and clemscore, we observe model-dependent trends: for Qwen and Claude, an increase in cycle edge ratio consistently coincides with better performance; DeepSeek tends to lose performance as the ratio increases; GPT-OSS and Nemotron show mixed associations. 

Claude points cluster at lower cycle ratios while remaining competitive in score; open-weight models occupy a wider band—extending into higher ratios—with correspondingly higher variance in outcomes. This suggests that additional cycling in open-weights is not reliably converted into payoff.

\begin{table}[t]
\centering
\small
\setlength{\tabcolsep}{6pt}
\begin{tabular}{ll}
\toprule
\textbf{Label} & \textbf{English cue words (case-insensitive)} \\
\midrule
\texttt{ASSERT}      & \textit{need}, \textit{should}, \textit{must} \\
\texttt{PROPOSE}     & \textit{maybe}, \textit{perhaps}, \textit{can}, \textit{could} \\
\texttt{UNDERMINE}   & \textit{but}, \textit{however}, \textit{wait} \\
\texttt{ALTERNATIVE} & \textit{alternatively}, \textit{another} \\
\texttt{CONCLUDE}    & \textit{so}, \textit{thus} \\
\bottomrule
\end{tabular}
\caption{Cue words used to label sentences of one reasoning trace into FSM states. We lowercase, lemmatize, and match whole tokens; punctuation is stripped; negation is ignored.}
\label{tab:lexical_cues}
\end{table}

\begin{table}[t]
\centering
\small
\begin{tabular}{p{0.55\linewidth} p{0.35\linewidth}}
\toprule
\textbf{Condition} & \textbf{Outcome} \\
\midrule
Multiple labels include \texttt{CONCLUDE} & Drop \texttt{CONCLUDE} and re-evaluate \\
Exactly one label remains & Assign that label \\
Multiple labels remain and include \texttt{PROPOSE} & Assign \texttt{PROPOSE} \\
Multiple labels remain and exclude \texttt{PROPOSE} & Mark as conflict (skip) \\
\bottomrule
\end{tabular}
\caption{Conflict resolution rules used after keyword matching.}
\label{tab:conflict_rules}
\end{table}

\subsection{LLM-based Analysis of Reasoning Traces}
\label{sec:llm-based_reasoning_analysis}

For all thinking models with available reasoning traces, we sampled a total of 309 transcripts with at least one instance for each game, language, and experiment combination for automated analysis.

We prompted \textit{GPT-5} to analyse certain aspects of the traces and give their output in json format, including not only the analysis, but also short explanations for each classification (Figure~\ref{fig:llm_analysis_prompt_and_example}). The explanations are not only meant to increase the quality of the classifications, but also to facilitate verifiability. An informal spot check indicated that these prompts performed satisfactorily at the annotation task.

\begin{figure*}[!t]
    \centering
    \begin{subfigure}{0.48\textwidth}
        \includegraphics[width=\linewidth]{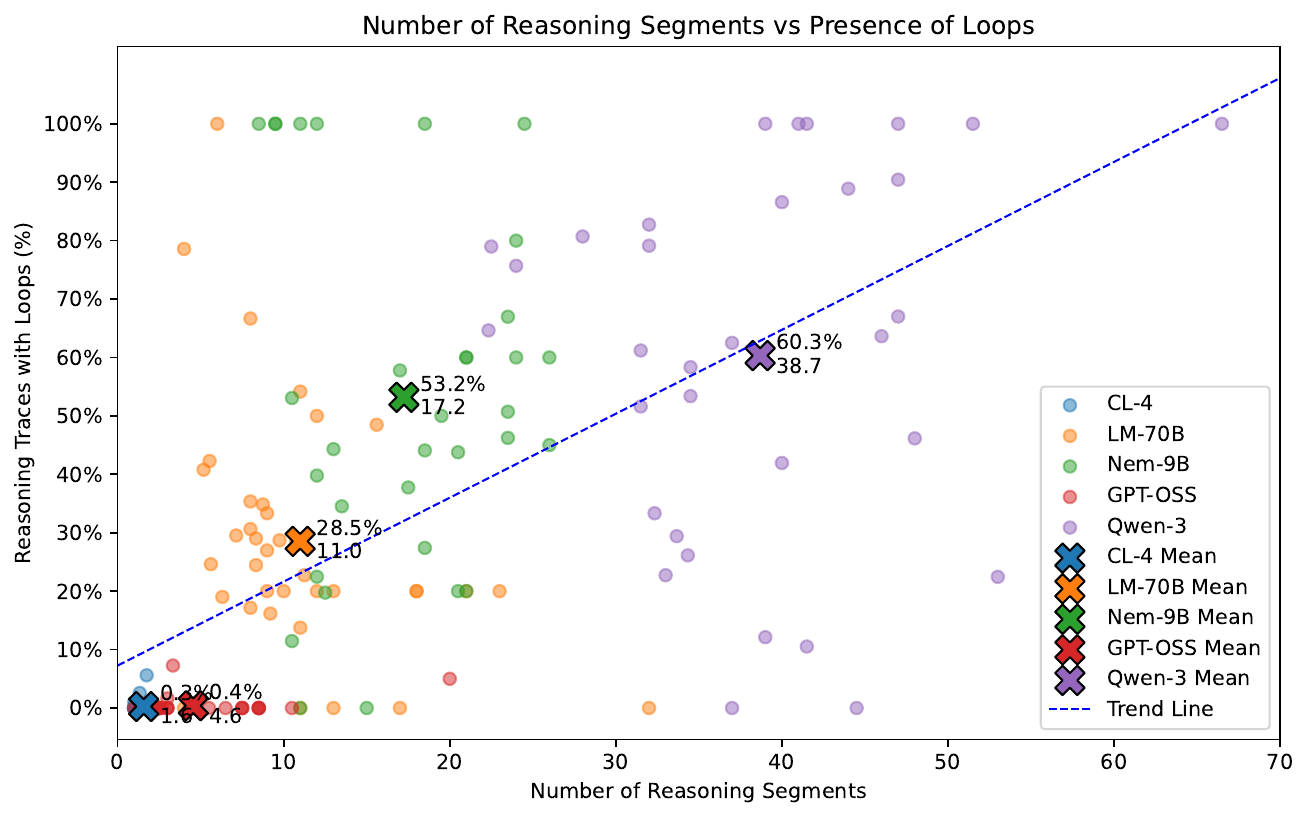}
        \caption{Number of segments \textit{vs.} percentage of traces containing loops identified by automated LLM analysis}
        \label{fig:n_segments_vs_loops}
    \end{subfigure}
    \hfill
    \begin{subfigure}{0.48\textwidth}
        \includegraphics[width=1.0\linewidth]{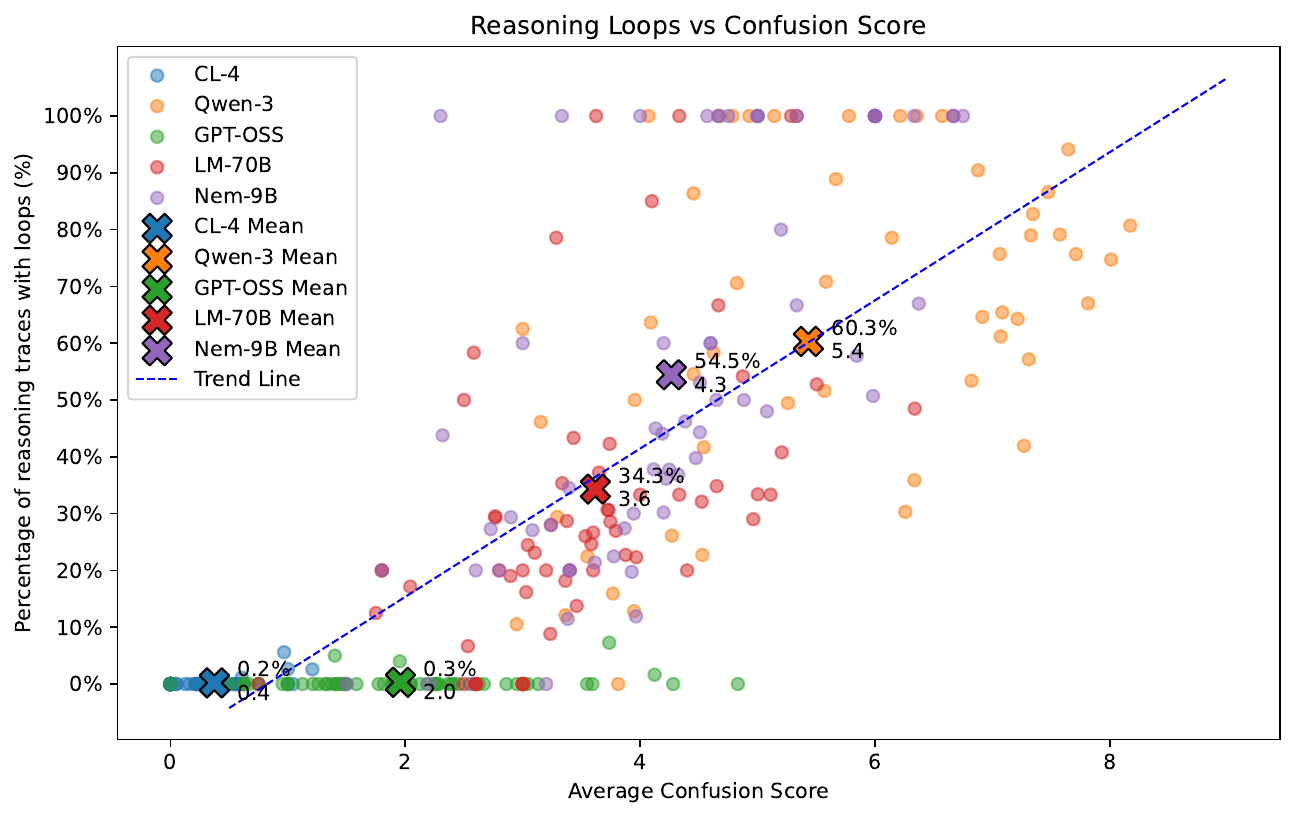}
        \caption{Average confusion score \textit{vs.} percentage of traces containing loops, as identified by automated LLM analysis}
        \label{fig:confusion_vs_loops}
    \end{subfigure}
    \hfill
    \begin{subfigure}[t]{0.48\textwidth}
        \includegraphics[width=\linewidth]{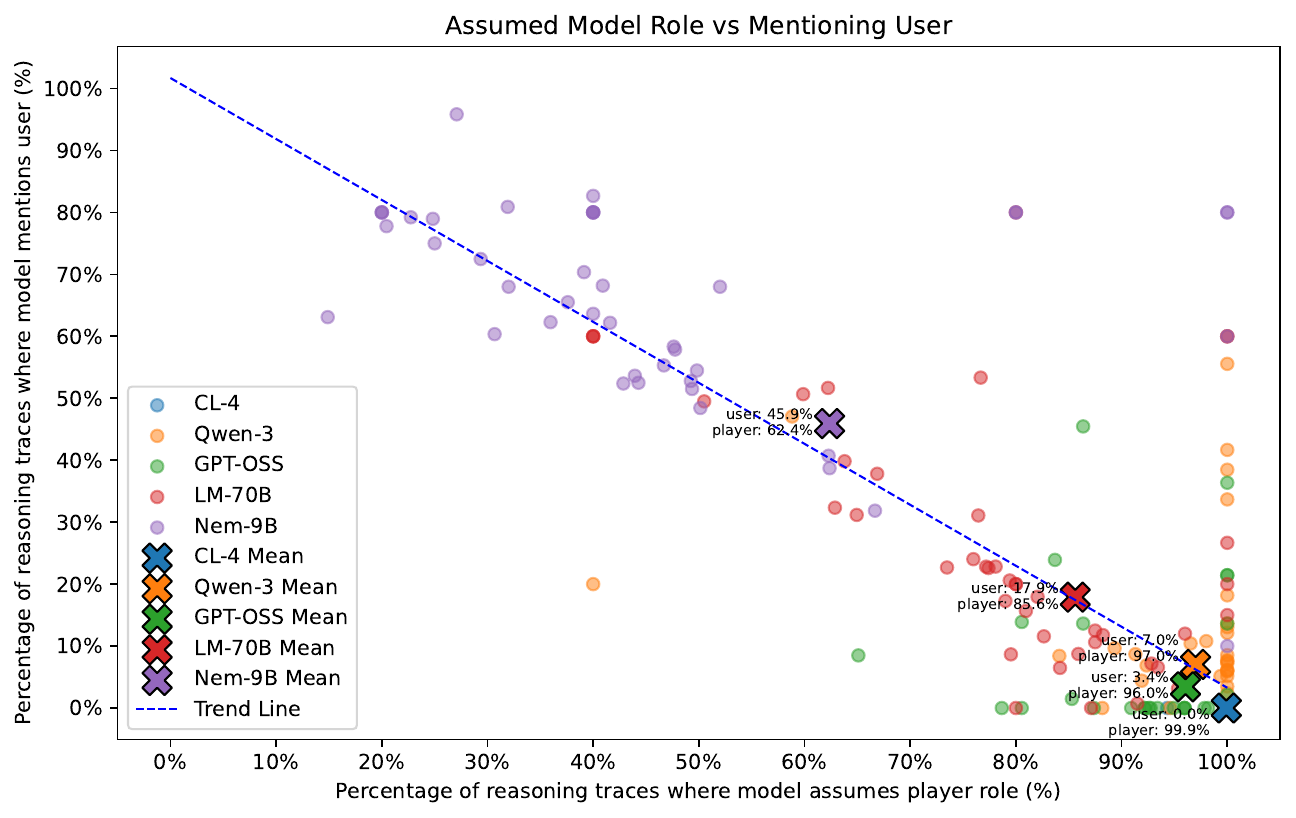}
        \caption{Percentage of reasoning traces where the model assumes a player role \textit{vs.} percentage where it explicitly mentions a user, as identified by automated LLM analysis}
        \label{fig:player_vs_user}
    \end{subfigure}
    \hfill
    \begin{subfigure}[t]{0.48\textwidth}
        \includegraphics[width=1.0\linewidth]{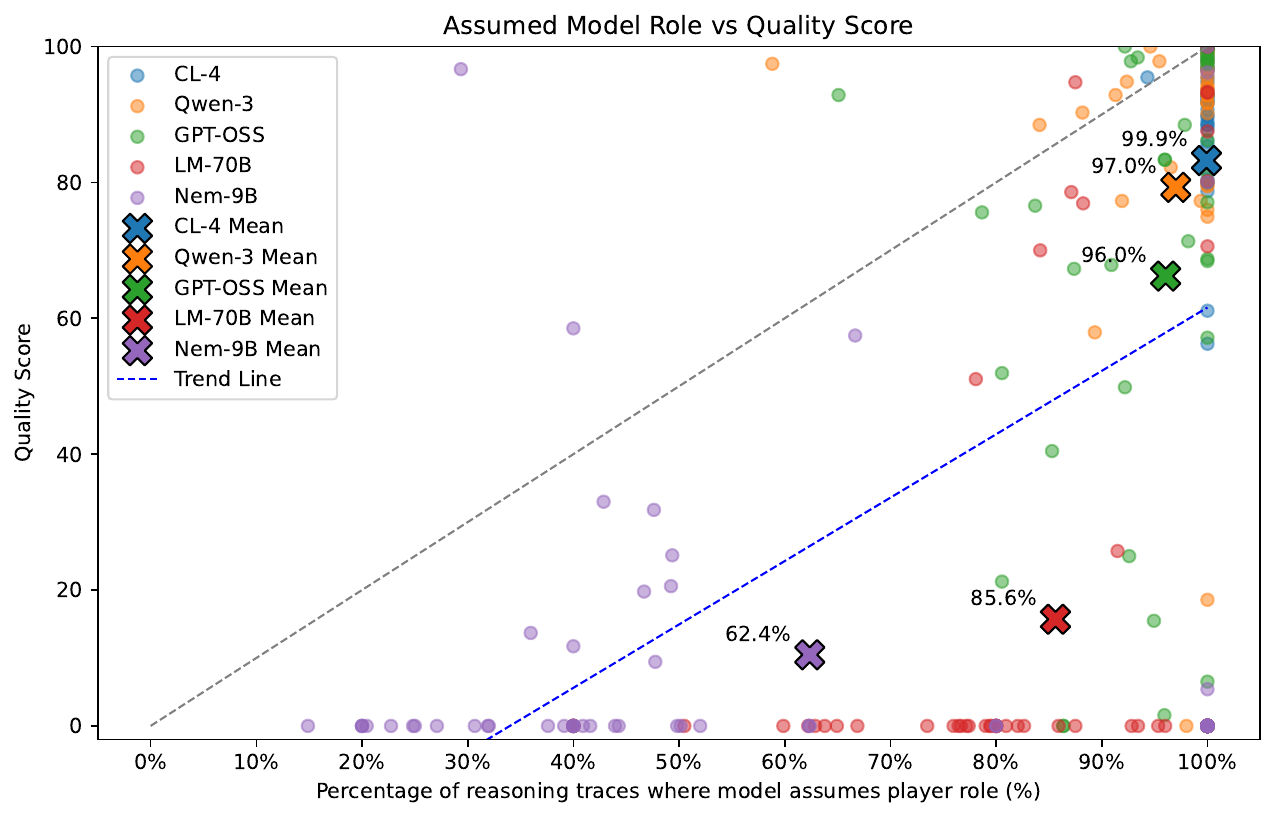}
        \caption{Percentage of reasoning traces where the model assumes a player role, as opposed to an assistant/helper role, plotted \textit{vs.} quality score. 95.2\% of data points are below the main diagonal, meaning a model rarely reaches a quality score higher than the percentage of reasoning traces where it assumes a player role.}
    \label{fig:assumed_role}
    \end{subfigure}
    \caption{Plots (a) through (c) corroborate the validity of the categorisation provided by the LLM. (d) shows the importance of assuming a player role for reaching high scores in the negotiation games.}
    \label{fig:llm_analysis_plots}
\end{figure*}

\textbf{Loops}
Among other things, we prompted the LLM to mark traces as containing loops if a thought or an action is repeated at least three times, and then calculated the percentage of traces in the transcript that contain loops. Experimentally, we also tasked the LLM to rate the `confusion' expressed in the trace on a scale from 0 (not confused at all) to 10 (extremely confused), and calculated the mean over all traces in one game transcript. Fig.~\ref{fig:n_segments_vs_loops} shows that loops correlate with the segment analysis of \ref{sec:cyclic_analysis}. It has to be noted that a large number of segments does not necessarily imply loops, and \textit{vice versa}; additionally, the loop analysis stretches over the whole transcript instead of just the first round. Fig.~\ref{fig:confusion_vs_loops} shows that the average confusion scores also correlate with percentage of loops in a transcript.

\textbf{Role Analysis}
Additionally, we tasked the LLM with classifying the role of the author of the trace as either `player' or `helper' to find out which role the model assumes within the game: is it aware of being an active participant, or does it take on an assistant role? Again we calculated the percentage of traces written from a player's perspective for each transcript.
To strengthen this analysis, we also let the LLM identify whether or not a `user' is mentioned in the thinking traces. In most cases, this is a sign that the model perceives the user as a player and itself as an assistant, although in some cases it also identified the user as the game master. Fig.~\ref{fig:player_vs_user} shows that these percentages are fairly complementary.

Finally, we tasked the LLM with identifying any other agentive persons mentioned in or implied by the traces and their respective roles, and we tried to identify if the trace shows awareness of another player.
In 17 out of 4123 (0.41\%) analysed traces, the LLM added the author to the agentive persons, and the name was given as either one of the first person pronouns `I', `Ich', `Io', as `assistant (self)' or as `author', and marked their role as `player'. For further analysis, we removed these duplicate persons.
Excluding these cases, we found additional 52 (1.26\%) trace analyses where the author role was given as `player' and two (other) players were listed in `agentive persons'. Mostly, this was because the other player was listed twice, for example once as `other player' and once as `teammate'. We also found one instance where a person `Spielleiter' (German for game master) was classified as `player'.

Given that these numbers of erroneous role assignments are fairly low, we assume that the author role assignment is similarly reliable. 

If the author assumed a player role \textit{and} at least one other player was listed as an agentive person, we concluded that the reasoning trace shows \textbf{`role awareness'}, and calculated average values over each transcript, which is the basis for Fig.~\ref{fig:role_awareness}.

\begin{figure*}[!t]
  \centering
  \begin{subfigure}[b]{\textwidth}
  \begin{prompt}
  Your task is to analyze the following strategic thinking trace and the subsequent response: \\
```trace \\
<TRACE> \\
``` \\
First of all, you have to identify the role of the author of the trace.  \\
\verb|[...]|
Classify the role of the author. Are they *helping* or *advising* somebody else who is playing the game, actively *playing* the game themselves, or are they in a *neutral* role, e.g., as an observer or game master? Please choose from ["helper", "player", "neutral"]. If none of these roles fit, classify it as "other", and specify it in your analysis. \\
Does the author mention any other agentive person or persons that influence the game? Give their names or titles and roles. \\
Valid roles for other persons are: ["player", "assistant", "game master"]. If a role does not fit into these categories, classify it as "other", and specify it in your analysis. \\
Following that, you should analyze the trace for the following aspects: \\
* Are there loops in the trace? (e.g., the author repeating the same action or thought at least three times, `true` or `false`) \\
\verb|...| \\
* Does the author express confusion? Rate the confusion on a scale from `0` (not confused at all) to `10` (extremely confused) \\
    * If you detect confusion, cite an example of displayed confusion. Only cite the example, no further explanation. \\
\verb|[...]| \\
Your reply should be in the following json format: \\
\verb|[...]|
  \end{prompt}
  \caption{Relevant excerpt of the prompt used to analyse a sample of thinking traces.}
  \label{fig:llm_analysis_prompt}
\end{subfigure}
\begin{subfigure}[b]{\textwidth}
    \footnotesize
    \begin{verbatim}
{
    "role_analysis": "The author is actively analyzing moves and positions on a grid and deciding the 
        next action, indicating they are playing the game themselves. They reference another player 
        asking “finished?”, but the author is the one executing or planning moves. No evidence suggests 
        they are merely advising or observing.",
        "person_and_number_analysis": "They predominantly use first-person singular (I) while 
            occasionally referring to another player.",
        "author_role": "player",
        [...]
        "agentive_persons_mentioned": [
            {
                "name": "other player",
                "role": "player",
                "mind": "They repeatedly asked “finished?” indicating they are awaiting confirmation."
            }
        ],
        "formal_analysis": "There is a repeated attempt to parse the ASCII grid structure, revisiting the
            same uncertainty multiple times, indicating looping and confusion. Frustration is mild to 
            moderate due to difficulty interpreting coordinates versus the drawing. Excitement is not 
            expressed.",
        "loops": true,
        "confusion_example": "This is too confusing.",
        "confusion": 8,
    [...]
}
  \end{verbatim}
    \caption{Excerpts from an LLM-based analysis of the last trace of Qwen-3, Clean Up, 2\_hard\_7obj, in English. Manual review showed that indeed, the model tried to parse the ASCII grid no less than 19 times, and tried to parse the initial grid another 5 times, even though it is irrelevant at this stage of the game. The `other player' is explicitly mentioned five times.}
    \label{fig:llm_analysis_example}
\end{subfigure}
  \hfill
\caption{Excerpts from the prompt used for LLM-based analysis and from an analysis produced by the model} \label{fig:llm_analysis_prompt_and_example}
\end{figure*}

\section{Deal or No Deal - Game Details}\label{sec:appendix_dond}

\subsection{Prompt Templates}\label{sec:dond_prompt_templates}

The English prompt templates for both players of the Deal or No Deal game are given in \cref{fig:dond_prompt_template_all}. The corresponding German and Italian versions of the prompts are given in \cref{fig:dond_prompt_template_all_de} and \cref{fig:dond_prompt_template_all_it}, respectively.

The game is started by the game master sending a message to both players following the prompt template shown in \cref{fig:dond_prompt_template}. For this, the variables enclosed in the \$ characters are replaced as follows.
\begin{enumerate}
  \item \$N\$ is replaced with the maximum number of turns before the game master instructs the players to submit a proposal. For all instances evaluated in this work, this has been set to $5$.
  \item \$GOAL\$ is replaced with the goal the players are trying to optimize. This depends only on the game mode. In the semi-competitive mode, it is given as "\emph{Your goal is to maximize the score you receive.}". On the other hand, in the cooperative mode, the goal is given as "\emph{Your goal is to maximize the sum of your score and the score of the other player.}".
  \item \$ITEMS\$ is replaced with the set of available items, e.g., "\emph{1 book, 2 hats, 2 balls.}". This value is always identical for both players.
  \item \$VALUE\_FUNCTION\$ is replaced with the value function for one of the players, e.g., "\emph{book: 0, hat: 1, ball: 4.}". Each player gets a different value function, and so the prompts for the two players differ.
\end{enumerate}

When the maximum number of turns has been reached, i.e., each player has sent 5 messages, the game master will instruct the player who is next that they must now submit a proposal using the prompt shown in \cref{fig:dond_prompt_template_timeout}. If the next message does not contain a correctly formatted proposal, the game is aborted.

If one of the two players makes a proposal, either proactively or when instructed by the game master, the game master will instruct the remaining player to also make a proposal. For this, the template shown in \cref{fig:dond_prompt_template_early} is used. Again, the next message must be a proposal, or the game is aborted. 

\begin{figure*}
  \centering
  \begin{subfigure}[t]{\textwidth}
    \begin{prompt}

  You are playing a negotiation game in which you have to agree on how to divide a set of items among you and another player.\\
  \\
  Rules:\\
  (a) You and the other player are given a set of items. Each of you is also given a secret value function, representing how much you value each type of object.\\
  (b) You exchange messages with the other player to agree on who gets which items. You can send a maximum of \$N\$ messages each, or terminate early by making a secret proposal at any time.\\
  (c) You are each asked to submit a secret proposal indicating the items you want formatted in square brackets as follows: "[Proposal: <number> <object name>, <number> <object name>, <...>]"\\
  (d) If your proposals are complementary, i.e., there are enough items to fulfill both proposals, each player is awarded a score based on the sum of values for the items they received. Otherwise, both of you get zero points.\\
  (e) \$GOAL\$\\
  \\
  Let us start.\\
  \\
  The set of available items is:\\
  \\
  \$ITEMS\$\\
  \\
  Your secret value function is:\\
  \\
  \$VALUE\_FUNCTION\$\\
  \\
  IMPORTANT: Your messages, unless it is the secret proposal, are directly transmitted to the other player, so do not include any response to the rules or text announcing your message. To make a secret proposal, use the indicated format. Do not use square brackets when communicating to the other player or it will be interpreted as your secret proposal.
    \end{prompt}
    \caption{English prompt template used for both players at the start of the game to inform them about the rules and game state. Values surrounded by \$ are to be replaced by different values depending on the game instance. Each player gets the same initial message, except for different values of \$VALUE\_FUNCTION\$.}
    \label{fig:dond_prompt_template}
  \end{subfigure}
  \begin{subfigure}[t]{0.48\textwidth}
    \centering
    \begin{prompt}

The time is up. It is now your turn to submit a secret proposal.
    \end{prompt}
    \caption{Prompt used when the maximum number of turns has been reached. This template is sent only to one of the two players. The other player receives the template shown in \cref{fig:dond_prompt_template_early}.}
    \label{fig:dond_prompt_template_timeout}
  \end{subfigure}
  \hfill
  \begin{subfigure}[t]{0.48\textwidth}
    \centering
    \begin{prompt}

The other player has submitted a secret proposal. It is now your turn to submit a proposal of your own.
    \end{prompt}
    \caption{Prompt used when the other player submits their proposal. This is sent to a player whenever the opposing player submits a secret proposal.}
    \label{fig:dond_prompt_template_early}
  \end{subfigure}
  \caption{English prompt template used to tell the players that they are now required to submit a secret proposal.}
  \label{fig:dond_prompt_template_all}
\end{figure*}

\begin{figure*}
  \centering
  \begin{subfigure}[t]{\textwidth}
    \begin{prompt}

  Sie spielen ein Verhandlungsspiel, bei dem Sie sich mit einem anderen Spieler darauf einigen müssen, wie eine Reihe von Gegenständen aufgeteilt werden soll.\\
  \\
  Die Regeln:\\
  (a) Sie und der andere Spieler erhalten eine Sammlung von Gegenständen. Jeder von Ihnen erhält außerdem eine geheime Wertfunktion, die angibt, wie viel Ihnen jede Art von Gegenstand wert ist.\\
  (b) Sie tauschen Nachrichten mit dem anderen Spieler aus, um zu vereinbaren, wer welche Gegenstände bekommt. Sie können jeweils maximal \$N\$ Nachrichten senden oder das Spiel vorzeitig beenden, indem Sie jederzeit einen geheimen Vorschlag machen.\\
  (c) Jeder von euch wird aufgefordert, einen geheimen Vorschlag zu machen, in dem ihr die gewünschten Gegenstände in eckigen Klammern wie folgt angibt: "[Vorschlag: <Nummer> <Objektname>, <Nummer> <Objektname>, <...>]"\\
  (d) Wenn eure Vorschläge komplementär sind, d.h. es gibt genug Gegenstände, um beide Vorschläge zu erfüllen, erhält jeder Spieler eine Punktzahl, die sich aus der Summe der Werte für die Gegenstände ergibt, die er erhalten hat. Andernfalls erhalten Sie beide null Punkte.\\
  (e) \$GOAL\$\\
  \\
  Beginnen wir.\\
  \\
  Die Menge der verfügbaren Gegenstände ist:\\
  \\
  \$ITEMS\$\\
  \\
  Deine geheime Wertfunktion ist:\\
  \\
  \$VALUE\_FUNCTION\$\\
  \\
  WICHTIG: Ihre Nachrichten werden, sofern es sich nicht um einen geheimen Vorschlag handelt, direkt an den anderen Spieler übermittelt, also fügen Sie keine Antwort auf die Regeln oder einen Text zur Ankündigung Ihrer Nachricht ein. Um einen geheimen Vorschlag zu machen, verwenden Sie das angegebene Format. Verwenden Sie keine eckigen Klammern, wenn Sie mit dem anderen Spieler kommunizieren, sonst wird dies als Ihr geheimer Vorschlag interpretiert.
    \end{prompt}
    \caption{German prompt template used for both players at the start of the game to inform them about the rules and game state. Values surrounded by \$ are to be replaced by different values depending on the game instance. Each player gets the same initial message, except for different values of \$VALUE\_FUNCTION\$.}
    \label{fig:dond_prompt_template_de}
  \end{subfigure}
  \begin{subfigure}[t]{0.48\textwidth}
    \centering
    \begin{prompt}

Die Zeit ist um. Sie sind jetzt an der Reihe, einen geheimen Vorschlag einzureichen.
    \end{prompt}
    \caption{Prompt used when the maximum number of turns has been reached. This template is sent only to one of the two players. The other player receives the template shown in \cref{fig:dond_prompt_template_early_de}.}
    \label{fig:dond_prompt_template_timeout_de}
  \end{subfigure}
  \hfill
  \begin{subfigure}[t]{0.48\textwidth}
    \centering
    \begin{prompt}

Der andere Spieler hat einen geheimen Vorschlag gemacht. Jetzt bist du an der Reihe, einen eigenen Vorschlag zu machen.
    \end{prompt}
    \caption{Prompt used when the other player submits their proposal. This is sent to a player whenever the opposing player submits a secret proposal.}
    \label{fig:dond_prompt_template_early_de}
  \end{subfigure}
  \caption{German prompt template used to tell the players that they are now required to submit a secret proposal.}
  \label{fig:dond_prompt_template_all_de}
\end{figure*}

\begin{figure*}
  \centering
  \begin{subfigure}[t]{\textwidth}
    \begin{prompt}

  State giocando a un gioco di negoziazione in cui dovete accordarvi su come dividere una serie di oggetti tra voi e un altro giocatore.\\
  \\
  Regole:\\
  (a) A Lei e all'altro giocatore viene dato un insieme di oggetti. Ognuno di voi riceve anche una funzione di valore segreta, che rappresenta il valore di ciascun tipo di oggetto.\\
  (b) Si scambiano messaggi con l'altro giocatore per concordare chi si aggiudica gli oggetti. Potete inviare un massimo di \$N\$ messaggi ciascuno, oppure terminare in anticipo facendo una proposta segreta in qualsiasi momento.\\
  (c) A ciascuno di voi viene chiesto di inviare una proposta segreta indicando gli oggetti che desiderate, formattata tra parentesi quadre come segue: "[Proposta: <numero> <nome oggetto>, <numero> <nome oggetto>, <...>]".\\
  (d) Se le vostre proposte sono complementari, cioè ci sono abbastanza oggetti per soddisfare entrambe le proposte, a ciascun giocatore viene assegnato un punteggio basato sulla somma dei valori degli oggetti ricevuti. In caso contrario, entrambi ricevono zero punti.\\
  (e) \$GOAL\$\\
  \\
  Cominciamo.\\
  \\
  L'insieme degli oggetti disponibili è:\\
  \\
  \$ITEMS\$\\
  \\
  La funzione valore segreta è:\\
  \\
  \$VALUE\_FUNCTION\$\\
  \\
  IMPORTANTE: i vostri messaggi, a meno che non si tratti di una proposta segreta, vengono trasmessi direttamente all'altro giocatore, quindi non includete alcuna risposta alle regole o testo di annuncio del vostro messaggio. Per fare una proposta segreta, utilizzate il formato indicato. Non utilizzare delle parentesi quadre quando si comunica all'altro giocatore, altrimenti verrà interpretata come una proposta segreta.
    \end{prompt}
    \caption{Italian prompt template used for both players at the start of the game to inform them about the rules and game state. Values surrounded by \$ are to be replaced by different values depending on the game instance. Each player gets the same initial message, except for different values of \$VALUE\_FUNCTION\$.}
    \label{fig:dond_prompt_template_it}
  \end{subfigure}
  \begin{subfigure}[t]{0.48\textwidth}
    \centering
    \begin{prompt}

Il tempo è scaduto. Ora tocca a Lei presentare una proposta segreta.
    \end{prompt}
    \caption{Prompt used when the maximum number of turns has been reached. This template is sent only to one of the two players. The other player receives the template shown in \cref{fig:dond_prompt_template_early_it}.}
    \label{fig:dond_prompt_template_timeout_it}
  \end{subfigure}
  \hfill
  \begin{subfigure}[t]{0.48\textwidth}
    \centering
    \begin{prompt}

L'altro giocatore ha presentato una proposta segreta. Ora tocca a Lei presentare una sua proposta.
    \end{prompt}
    \caption{Prompt used when the other player submits their proposal. This is sent to a player whenever the opposing player submits a secret proposal.}
    \label{fig:dond_prompt_template_early_it}
  \end{subfigure}
  \caption{Italian prompt template used to tell the players that they are now required to submit a secret proposal.}
  \label{fig:dond_prompt_template_all_it}
\end{figure*}

\subsection{Evaluation Metrics}\label{subsec:appendix_dond_metric}

For the cooperative game mode, it is the ratio between the achieved total score and the maximum total score for that game instance, 

\begin{align*}
\text{Quality}_{\text{coop}} = \frac{\text{Score}_A + \text{Score}_B}{\max(\text{Score}_A' + \text{Score}_B')}
\end{align*}

where $\text{Score}_A$ and $\text{Score}_B$ represent the actual scores achieved by players A and B, respectively, and $\max(\text{Score}_A' + \text{Score}_B')$ denotes the maximum possible combined score achievable for the given game instance.

For the semi-competitive setting, a metric based on Pareto efficiency has been used.

\begin{align*}
\text{Quality}_{\text{semi}} = 1 - \frac{\text{Maximum Pareto Improvement}}{\max(\text{Score}_i')}
\end{align*}

where $\text{Maximum Pareto Improvement}$ represents the largest possible improvement in one player's score without decreasing the other player's score from the current agreement, and $\max(\text{Score}_i')$ denotes the maximum possible score achievable by any single player $i$ in the given game instance.

\subsection{Overall Results}

\begin{figure}[!ht]
    \centering
    \includegraphics[width=\linewidth]{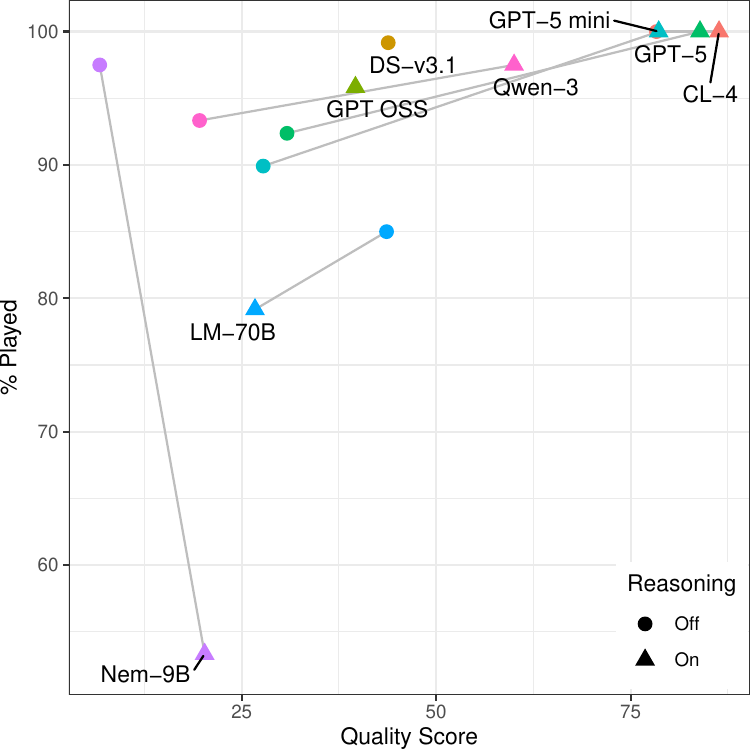}
    \caption{Summarized evaluation results for each model. The y-axis indicates the percentage of played, i.e., not aborted, games. The x-axis represents the average quality score obtained when considering only played games.}
    \label{fig:dond_playedquality}
\end{figure}

\Cref{fig:dond_playedquality} shows the average quality scores and the percentage of games played for each model for the DoND game, averaged over the different modes and languages of the game.

We can observe that most models achieve a high percentage of games played, which means they can accurately follow the instructions. The only exceptions to this are the Llama model and the Nematron model with reasoning enabled. The models GPT-5, GPT-5 mini, and Claude 4 with reasoning enabled show the strongest performance. 

Generally, we can see that the models perform better with reasoning enabled than with it disabled. While the difference between reasoning and non-reasoning is small for the Claude 4 model, reasoning gives a large performance improvement for the GPT-5 and GPT-5 mini, as well as the Qwen-3 model.

\begin{figure}[!ht]
    \centering
    \includegraphics[width=\linewidth]{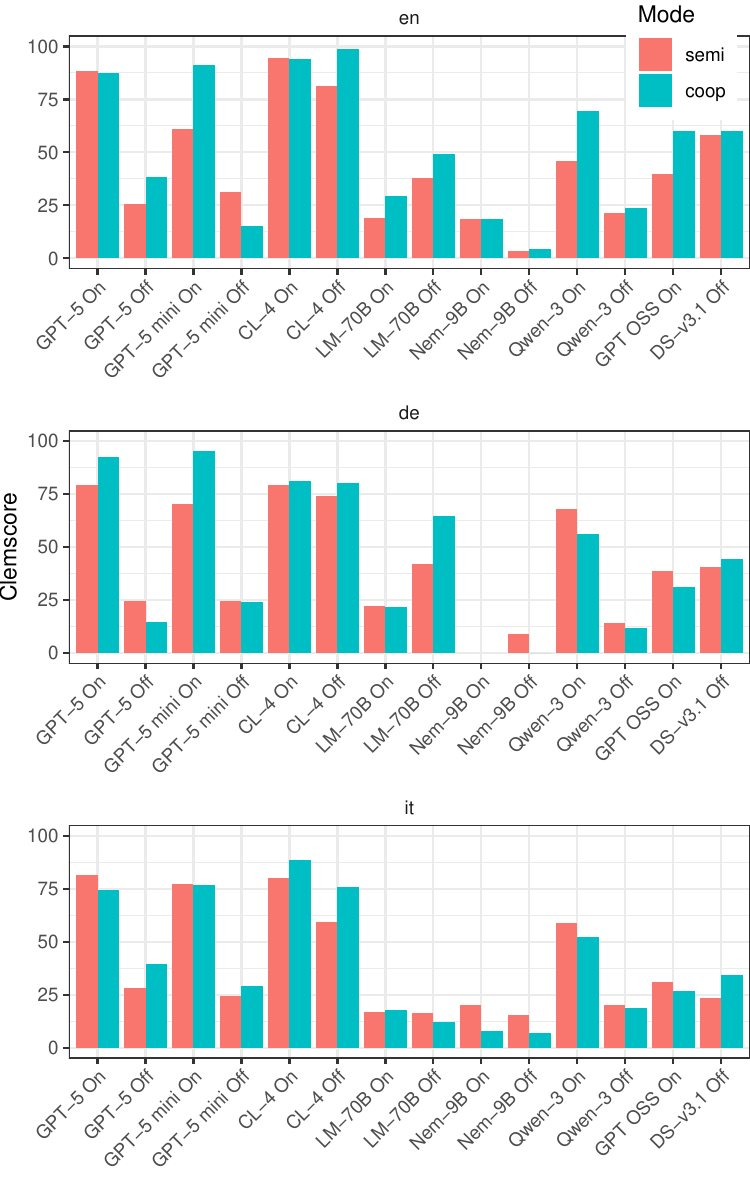}
    \caption{Clemscore for different models separated by language and game mode.}
    \label{fig:dond_perlanguage}
\end{figure}

To better understand the performance of the models, we show a more detailed breakdown by language and game mode in \cref{tab:dond-detailed-results}. These results are also depicted in \cref{fig:dond_perlanguage}, which shows the performance differences between experiments for English, German, and Italian tasks.

We can see here again that most models follow the instructions well. We can also see that the reason the Nematron model with reasoning enabled has such a low percentage of played games overall is that it has not played a single game in German correctly. Its percentage of played games is otherwise comparable to Llama. By inspecting the transcripts of some games played by Nematron in German, it becomes clear that the reason for the aborted games is that Nematron always made the proposal in English, whereas it should have made it in German. This is likely because Nematron also seems to reason exclusively in English.

Except for some outliers, we also generally observe that performance in the cooperative and semi-competitive game modes is comparable. Slightly higher quality scores are achieved for the cooperative mode (49\%) as compared to the semi-competitive mode (46\%). The number of aborted games is consistent, with an average of 8\% for both game modes.

Performance was generally highest in English. On average across all models, the clemscore for English was 49, while it was 43 and 40, respectively, for German and Italian. Due to the issues of the Nematron model in German, the number of aborted games is higher in German, with 13\% compared to 6\% for both English and Italian. However, in German, the average quality score for the games played is higher at 49\% compared to the average of 43\% for Italian.

\begin{table*}[!ht]
  \footnotesize
  \centering
  \setlength{\tabcolsep}{4.5pt}
  \begin{tabular}{c|l|cc|cc|cc|cc|cc|cc|c|c|}
    & & \multicolumn{2}{c}{\textbf{GPT-5}} & \multicolumn{2}{c}{\begin{tabular}{c}\textbf{GPT-5}\\\textbf{mini}\end{tabular}} & \multicolumn{2}{c}{\textbf{CL-4}} & \multicolumn{2}{c}{\textbf{LM-70B}} & \multicolumn{2}{c}{\textbf{Nem-9B}} & \multicolumn{2}{c}{\textbf{Qwen-3}} & \multicolumn{1}{c}{\begin{tabular}{c}\textbf{GPT}\\\textbf{OSS}\end{tabular}} & \textbf{DS-v3.1} \\ 
    \cline{2-16}
    & \textbf{Metric} & On & Off & On & Off & On & Off & On & Off & On & Off & On & Off & On & Off \\ 
    \hline
    \multirow{6}{*}{\rotatebox{90}{\textbf{semi en}}} & \textbf{Clemscore} & 88 & 26 & 61 & 32 & \textbf{94} & 82 & 19 & 38 & 18 & 3 & 46 & 22 & 40 & 58 \\ 
     & \% Played & \textbf{100} & \textbf{100} & \textbf{100} & 90 & \textbf{100} & \textbf{100} & 80 & 55 & 85 & 95 & \textbf{100} & \textbf{100} & 95 & 95 \\ 
     & \% Agreement & \textbf{95} & 28 & 65 & 50 & \textbf{95} & 85 & 25 & 73 & 24 & 5 & 55 & 25 & 42 & 68 \\ 
     & \% Optimal & 70 & 17 & 45 & 6 & \textbf{90} & 70 & 19 & 45 & 12 & 0 & 30 & 10 & 42 & 37 \\ 
     & Quality Score & 88 & 26 & 61 & 35 & \textbf{94} & 82 & 24 & 69 & 22 & 4 & 46 & 22 & 42 & 61 \\ 
     & Avg. \# Messages & 3.5 & 4.3 & 4.7 & 6 & 5.8 & 6.1 & 2.8 & 9.3 & 2 & 2 & 3.3 & 4 & 4.7 & 5.8 \\ 
    \hline
    \multirow{6}{*}{\rotatebox{90}{\textbf{semi de}}} & \textbf{Clemscore} & \textbf{80} & 24 & 70 & 24 & \textbf{80} & 74 & 22 & 42 & 0 & 9 & 68 & 14 & 38 & 40 \\ 
     & \% Played & \textbf{100} & 85 & \textbf{100} & 90 & \textbf{100} & \textbf{100} & 75 & 95 & 0 & \textbf{100} & \textbf{100} & 85 & 95 & \textbf{100} \\ 
     & \% Agreement & \textbf{90} & 29 & 80 & 28 & 80 & \textbf{90} & 33 & 53 & 0 & 10 & 75 & 18 & 47 & 50 \\ 
     & \% Optimal & 50 & 24 & 60 & 22 & \textbf{75} & 55 & 20 & 21 & 0 & 5 & 55 & 12 & 26 & 30 \\ 
     & Quality Score & \textbf{80} & 29 & 70 & 27 & \textbf{80} & 74 & 29 & 44 & 0 & 9 & 68 & 16 & 41 & 40 \\ 
     & Avg. \# Messages & 3.8 & 4.8 & 3.8 & 4.9 & 7 & 6.2 & 2.8 & 8.4 & 1 & 2 & 3.8 & 3.8 & 3.6 & 6.2 \\ 
    \hline
    \multirow{6}{*}{\rotatebox{90}{\textbf{semi it}}} & \textbf{Clemscore} & \textbf{82} & 28 & 78 & 24 & 80 & 60 & 17 & 16 & 20 & 15 & 59 & 20 & 31 & 24 \\ 
     & \% Played & \textbf{100} & 95 & \textbf{100} & 85 & \textbf{100} & \textbf{100} & 70 & 90 & 75 & 95 & \textbf{100} & 90 & \textbf{100} & \textbf{100} \\ 
     & \% Agreement & \textbf{90} & 32 & 85 & 29 & 85 & 65 & 29 & 28 & 33 & 21 & 65 & 22 & 35 & 25 \\ 
     & \% Optimal & 55 & 21 & 55 & 24 & \textbf{65} & 50 & 7 & 0 & 7 & 5 & 40 & 22 & 30 & 20 \\ 
     & Quality Score & \textbf{82} & 29 & 78 & 29 & 80 & 60 & 24 & 18 & 27 & 16 & 59 & 22 & 31 & 24 \\ 
     & Avg. \# Messages & 4 & 4.5 & 3.5 & 4.3 & 6 & 5.6 & 4.2 & 6.7 & 1.9 & 3.2 & 3.7 & 4.8 & 3.9 & 5.4 \\ 
    \hline
    \multirow{6}{*}{\rotatebox{90}{\textbf{coop en}}} & \textbf{Clemscore} & 87 & 38 & 91 & 15 & 94 & \textbf{99} & 30 & 49 & 18 & 4 & 70 & 24 & 60 & 60 \\ 
     & \% Played & \textbf{100} & 95 & \textbf{100} & 95 & \textbf{100} & \textbf{100} & \textbf{100} & 80 & 90 & \textbf{100} & 90 & \textbf{100} & 95 & \textbf{100} \\ 
     & \% Agreement & 89 & 42 & 95 & 17 & 95 & \textbf{100} & 35 & 69 & 22 & 5 & 78 & 25 & 79 & 65 \\ 
     & \% Optimal & 74 & 26 & \textbf{89} & 11 & 85 & 85 & 0 & 31 & 6 & 0 & 72 & 15 & 32 & 50 \\ 
     & Quality Score & 87 & 40 & 91 & 16 & 94 & \textbf{99} & 30 & 61 & 21 & 4 & 77 & 24 & 63 & 60 \\ 
     & Avg. \# Messages & 3.8 & 4.2 & 5.2 & 4.9 & 5.4 & 5.6 & 2.5 & 9.2 & 2 & 2 & 3.8 & 3.3 & 5 & 5.4 \\ 
    \hline
    \multirow{6}{*}{\rotatebox{90}{\textbf{coop de}}} & \textbf{Clemscore} & 92 & 15 & \textbf{95} & 24 & 81 & 80 & 22 & 65 & 0 & 0 & 56 & 12 & 31 & 44 \\ 
     & \% Played & \textbf{100} & 85 & \textbf{100} & 80 & \textbf{100} & \textbf{100} & 70 & \textbf{100} & 0 & \textbf{100} & 95 & 90 & 95 & \textbf{100} \\ 
     & \% Agreement & \textbf{100} & 18 & \textbf{100} & 31 & 85 & 85 & 36 & 75 & 0 & 0 & 63 & 17 & 37 & 45 \\ 
     & \% Optimal & 75 & 12 & \textbf{80} & 25 & 55 & 60 & 7 & 30 & 0 & 0 & 32 & 0 & 16 & 40 \\ 
     & Quality Score & 92 & 17 & \textbf{95} & 30 & 81 & 80 & 31 & 65 & 0 & 0 & 59 & 13 & 33 & 44 \\ 
     & Avg. \# Messages & 3.9 & 4.5 & 3.8 & 5.2 & 5.7 & 5.8 & 2 & 9 & 1 & 2 & 3.8 & 3.7 & 4.2 & 5.8 \\ 
    \hline
    \multirow{6}{*}{\rotatebox{90}{\textbf{coop it}}} & \textbf{Clemscore} & 74 & 40 & 77 & 29 & \textbf{89} & 76 & 18 & 12 & 8 & 7 & 52 & 19 & 27 & 34 \\ 
     & \% Played & \textbf{100} & 95 & \textbf{100} & \textbf{100} & \textbf{100} & \textbf{100} & 80 & 90 & 70 & 95 & \textbf{100} & 95 & 95 & \textbf{100} \\ 
     & \% Agreement & 85 & 47 & 85 & 30 & \textbf{90} & 85 & 25 & 17 & 14 & 11 & 55 & 26 & 47 & 35 \\ 
     & \% Optimal & 50 & 26 & 55 & 20 & \textbf{70} & 45 & 6 & 6 & 0 & 0 & 35 & 5 & 5 & 25 \\ 
     & Quality Score & 74 & 42 & 77 & 29 & \textbf{89} & 76 & 22 & 13 & 11 & 7 & 52 & 20 & 28 & 34 \\ 
     & Avg. \# Messages & 3.6 & 3.9 & 3.5 & 4 & 5.3 & 5.3 & 3.4 & 7.6 & 1.9 & 3.5 & 3.9 & 4.4 & 3.2 & 5.2 \\ 
    \hline
  \end{tabular}
  \caption{Detailed breakdown of the results for each game mode, language, and model combination across multiple metrics for the DoND game.  The best result in each row is indicated in bold. Note that the quality score, as well as the percentages of agreements and optimal outcomes, are computed only with respect to the non-aborted games.}
  \label{tab:dond-detailed-results}
\end{table*}

\begin{figure}[!ht]
    \centering
    \includegraphics[width=\linewidth]{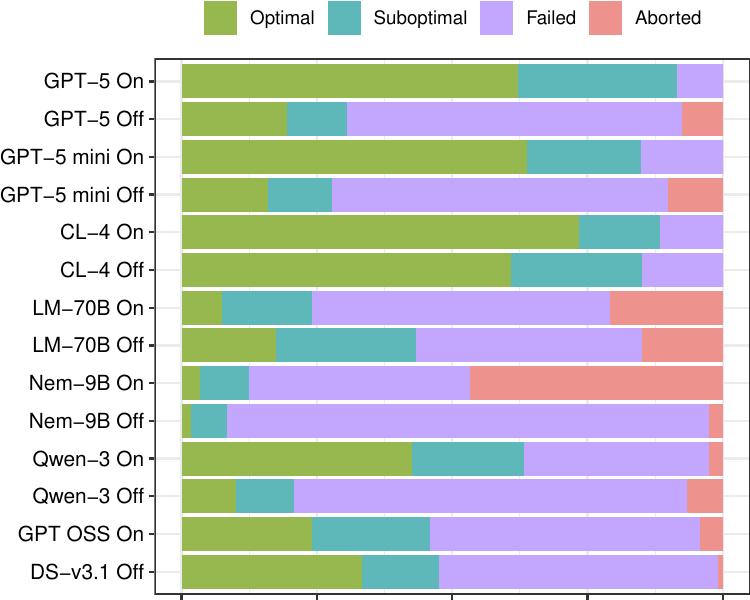}
    \caption{Percentage of games that ended with a specific game outcome during the evaluation of different models. All results have been averaged across the two game modes and three languages for this plot.}
    \label{fig:dond_outcome}
\end{figure}

We can also analyze the results based on the outcome that was achieved. This is depicted in \cref{fig:dond_outcome}. We separate into four possible game outcomes. The game can either end with complementary proposals or not. If there are complementary proposals, it could be with the optimal score, i.e., a quality score of 100, or with a suboptimal score less than 100. Finally, if there is no valid agreement, it can either be caused by conflicting proposals or by the game being aborted due to rule violations.

In general, we can observe similar results to those above. It becomes clear that many of the low scores can be attributed to failed agreements, where players have submitted conflicting proposals. Particularly for GTP-5 and GPT-5 mini, this problem is largely solved by using the reasoning instead of the non-reasoning versions.

\subsection{Detailed Analysis}

To better understand what caused the differences in performance, and to analyze the strategies employed by the different models, we have performed a further breakdown of the outcomes for the English instances. \Cref{tab:dond-game-outcomes-coop} and \Cref{tab:dond-game-outcomes-semi} show the result for each game, categorized by the reason for which the given result has been reached.

\begin{table*}[!ht]
  \footnotesize
  \centering
  \setlength{\tabcolsep}{4pt}
  \begin{tabular}{r|l|cc|cc|cc|cc|cc|cc|c|c|}
    \multicolumn{2}{c}{} & \multicolumn{2}{c}{\textbf{GPT-5}} & \multicolumn{2}{c}{\begin{tabular}{c}\textbf{GPT-5}\\\textbf{mini}\end{tabular}} & \multicolumn{2}{c}{\textbf{CL-4}} & \multicolumn{2}{c}{\textbf{LM-70B}} & \multicolumn{2}{c}{\textbf{Nem-9B}} & \multicolumn{2}{c}{\textbf{Qwen-3}} & \multicolumn{1}{c}{\begin{tabular}{c}\textbf{GPT}\\\textbf{OSS}\end{tabular}} & \multicolumn{1}{c}{\textbf{DS-v3.1}} \\ 
    \cline{2-16}
    \textbf{Outcome} & \textbf{Reason} & On & Off & On & Off & On & Off & On & Off & On & Off & On & Off & On & Off \\ 
    \cline{2-16}
    \multirow{3}{*}{Optimal} & Explicit max. & 11 & 3 & 11 & 0 & 0 & 1 & 0 & 0 & 0 & 0 & 5 & 0 & 1 & 3 \\
     & Simple negot. & 3 & 2 & 6 & 2 & 17 & 16 & 0 & 5 & 0 & 0 & 8 & 3 & 5 & 7 \\
     & Luck & 0 & 0 & 0 & 0 & 0 & 0 & 0 & 0 & 1 & 0 & 0 & 0 & 0 & 0 \\
    \cline{2-16}
    \multirow{5}{*}{Suboptimal} & Unclear pref. & 0 & 1 & 0 & 0 & 1 & 1 & 1 & 1 & 0 & 0 & 0 & 0 & 1 & 1 \\
     & Unclaimed items & 0 & 0 & 0 & 0 & 0 & 0 & 0 & 1 & 0 & 0 & 1 & 1 & 0 & 0 \\
     & Claiming valueless & 0 & 2 & 0 & 1 & 1 & 2 & 0 & 4 & 0 & 0 & 0 & 0 & 3 & 0 \\
     & Wrong proposal & 0 & 0 & 0 & 0 & 0 & 0 & 1 & 0 & 0 & 0 & 0 & 0 & 1 & 0 \\
     & Premature prop. & 3 & 0 & 1 & 0 & 0 & 0 & 3 & 0 & 3 & 1 & 0 & 1 & 3 & 1 \\
    \cline{2-16}
    \multirow{2}{*}{Failed} & Wrong proposal & 1 & 11 & 0 & 13 & 0 & 0 & 3 & 5 & 0 & 0 & 4 & 7 & 1 & 7 \\
     & Premature prop. & 1 & 0 & 1 & 2 & 1 & 0 & 12 & 0 & 14 & 19 & 0 & 8 & 4 & 1 \\
    \cline{2-16}
    \multirow{2}{*}{Aborted} & Wrong syntax & 0 & 1 & 0 & 1 & 0 & 0 & 0 & 4 & 0 & 0 & 1 & 0 & 1 & 0 \\
     & Non-existent items & 0 & 0 & 0 & 0 & 0 & 0 & 0 & 0 & 2 & 0 & 0 & 0 & 0 & 0 \\
    \cline{2-16}
  \end{tabular}
  \caption{Game outcomes for the \textbf{cooperative} DoND game mode. For each outcome, we also differentiate by the reason or strategy that caused the given outcome to be achieved.}
  \label{tab:dond-game-outcomes-coop}
\end{table*}

In the analysis of the cooperative game mode shown in \cref{tab:dond-game-outcomes-coop}, we can observe some significant differences in strategies between the different models. It can be seen that the GPT-5 models with reasoning disabled will often explicitly discuss how to achieve a maximum score, i.e., they will talk about having to give each item to the player that values it most. This is in contrast to, for example, Claude 4 or the GPT-5 models without reasoning enabled, as those models will often just follow a generic negotiation routine, trying to obtain a fair-sounding agreement.

We can further observe that a main reason for the low scores of the non-reasoning versions of the GPT-5 models is that they submit the wrong proposal. This means that the proposal they submit to the game master is often not the one they discussed with the other player in the natural language messages. This manifests itself sometimes as submitting the proposal for the other player, sometimes as proposals that include the complete set of items, and, at other times, even completely unrelated proposals. Deepseek-v3.1 also suffers from a less severe version of this problem.

Lastly, some other models, like Deepseek distilled Llama and the Nematron models, as well as the Qwen-3 model without reasoning, often submit proposals before discussing and agreeing to a division with the other player. This is particularly severe for the Nematron models, as these nearly always submit a proposal as the first message, without exchanging any messages with the other player. Looking at the reasoning traces, we can observe that the Nematron model does not seem to understand that it can communicate with the other player. It often expresses correctly that it does not know the other player's values, but does not realize that it could ask for them.

\begin{table*}[!ht]
  \footnotesize
  \centering
  \setlength{\tabcolsep}{4pt}
  \begin{tabular}{r|l|cc|cc|cc|cc|cc|cc|c|c|}
    \multicolumn{2}{c}{} & \multicolumn{2}{c}{\textbf{GPT-5}} & \multicolumn{2}{c}{\begin{tabular}{c}\textbf{GPT-5}\\\textbf{mini}\end{tabular}} & \multicolumn{2}{c}{\textbf{CL-4}} & \multicolumn{2}{c}{\textbf{LM-70B}} & \multicolumn{2}{c}{\textbf{Nem-9B}} & \multicolumn{2}{c}{\textbf{Qwen-3}} & \multicolumn{1}{c}{\begin{tabular}{c}\textbf{GPT}\\\textbf{OSS}\end{tabular}} & \multicolumn{1}{c}{\textbf{DS-v3.1}} \\ 
    \cline{2-16}
    \textbf{Outcome} & \textbf{Reason} & On & Off & On & Off & On & Off & On & Off & On & Off & On & Off & On & Off \\ 
    \cline{2-16}
    \multirow{3}{*}{Optimal} & Explicit max. & 0 & 0 & 0 & 0 & 0 & 1 & 0 & 0 & 0 & 0 & 0 & 0 & 0 & 1 \\
     & Simple negot. & 14 & 3 & 9 & 1 & 18 & 13 & 1 & 5 & 0 & 0 & 6 & 2 & 6 & 6 \\
     & Luck & 0 & 0 & 0 & 0 & 0 & 0 & 2 & 0 & 2 & 0 & 0 & 0 & 2 & 0 \\
    \cline{2-16}
    \multirow{5}{*}{Suboptimal} & Unclear pref. & 0 & 0 & 0 & 0 & 0 & 0 & 0 & 0 & 0 & 0 & 0 & 0 & 0 & 1 \\
     & Unclaimed items & 1 & 1 & 2 & 5 & 0 & 0 & 0 & 1 & 0 & 0 & 3 & 3 & 0 & 0 \\
     & Claiming valueless & 2 & 1 & 2 & 3 & 1 & 3 & 0 & 2 & 0 & 0 & 2 & 0 & 0 & 5 \\
     & Wrong proposal & 0 & 0 & 0 & 0 & 0 & 0 & 0 & 0 & 0 & 0 & 0 & 0 & 0 & 0 \\
     & Premature prop. & 2 & 0 & 0 & 0 & 0 & 0 & 1 & 0 & 2 & 1 & 0 & 0 & 0 & 0 \\
    \cline{2-16}
    \multirow{2}{*}{Failed} & Wrong proposal & 1 & 12 & 6 & 8 & 0 & 3 & 5 & 2 & 0 & 0 & 2 & 9 & 7 & 6 \\
     & Premature prop. & 0 & 1 & 1 & 1 & 1 & 0 & 7 & 1 & 14 & 18 & 6 & 6 & 4 & 0 \\
    \cline{2-16}
    \multirow{2}{*}{Aborted} & Wrong syntax & 0 & 0 & 0 & 2 & 0 & 0 & 4 & 9 & 2 & 0 & 1 & 0 & 1 & 1 \\
     & Non-existent items & 0 & 0 & 0 & 0 & 0 & 0 & 0 & 0 & 1 & 1 & 0 & 0 & 0 & 0 \\
    \cline{2-16}
  \end{tabular}
  \caption{Game outcomes for the \textbf{semi-competitive} DoND game mode. For each outcome, we also differentiate by the reason or strategy that caused the given outcome to be achieved.}
  \label{tab:dond-game-outcomes-semi}
\end{table*}

For the semi-competitive game mode, we see largely the same results, with the main difference being that the GPT-5 models no longer try to explicitly discuss maximizing their score. This makes sense as the setting is no longer cooperative, so it is no longer purely advantageous to share this reasoning with the other player. Another difference from the cooperative mode is that the GPT-5 mini model with reasoning and the GPT-OSS model now make wrong proposals more often. It is not clear why this happens.

We can also generally see an increase in the number of suboptimal outcomes. Interestingly, this is not necessarily because of unclear preferences, but because models are often requesting items that have no value to them, or leave some items unclaimed by either player, sometimes even expressing the intention of doing so deliberately. Keeping an item that does not have any value does not make much sense for any of the players, as giving up that item could instead be used as leverage for getting some other items.

\Cref{tab:dond-cooperation-coop} and \cref{tab:dond-cooperation-semi} show how likely different models were to reveal their value function to the other player. This is roughly an estimate of how cooperative they were.

\begin{table*}[!ht]
  \footnotesize
  \centering
  \setlength{\tabcolsep}{4.5pt}
  \begin{tabular}{|l|cc|cc|cc|cc|cc|cc|c|c|}
    \multicolumn{1}{c}{} & \multicolumn{2}{c}{\textbf{GPT-5}} & \multicolumn{2}{c}{\begin{tabular}{c}\textbf{GPT-5}\\\textbf{mini}\end{tabular}} & \multicolumn{2}{c}{\textbf{CL-4}} & \multicolumn{2}{c}{\textbf{LM-70B}} & \multicolumn{2}{c}{\textbf{Nem-9B}} & \multicolumn{2}{c}{\textbf{Qwen-3}} & \multicolumn{1}{c}{\begin{tabular}{c}\textbf{GPT}\\\textbf{OSS}\end{tabular}} & \multicolumn{1}{c}{\textbf{DS-v3.1}} \\ 
    \cline{2-15}
    \multicolumn{1}{c|}{} & On & Off & On & Off & On & Off & On & Off & On & Off & On & Off & On & Off \\ 
    \hline
    Tell exact values & 19 & 19 & 17 & 12 & 0 & 3 & 1 & 1 & 0 & 0 & 13 & 17 & 14 & 13 \\
    Tell only relative values & 0 & 1 & 0 & 7 & 20 & 17 & 0 & 18 & 0 & 0 & 2 & 1 & 0 & 5 \\
    Do not tell preferences & 0 & 0 & 2 & 0 & 0 & 0 & 7 & 1 & 0 & 0 & 5 & 0 & 5 & 2 \\
    Do not communicate at all & 0 & 0 & 0 & 0 & 0 & 0 & 12 & 0 & 20 & 20 & 0 & 2 & 1 & 0 \\
    \hline
  \end{tabular}
  \caption{GModel willingness to share its value function for the \textbf{cooperative} DoND game mode.}
  \label{tab:dond-cooperation-coop}
\end{table*}

Clearly, in the cooperative game mode, it makes little sense to hide one's value function, as it is essential for knowing which allocation is optimal. Still, we can see in \cref{tab:dond-cooperation-coop} that some models were unwilling to give exact values. Some models, namely the Nematron and Deepseek distilled Llama models, were not willing to communicate preferences at all, and sometimes did not even communicate, instead immediately making a proposal. This explains their poor performance. On the other hand, the Claude 4 models were unwilling to give out their exact value function, preferring instead to only give relative or subjective characterizations of it. In some instances, the Claude 4 models even explicitly mention not wanting to reveal their exact values, either in messages or reasoning traces, even though it can only be beneficial in the cooperative setting. Finally, all the other models were willing to give the exact value function to the other player.

\begin{table*}[!ht]
  \footnotesize
  \centering
  \setlength{\tabcolsep}{4.5pt}
  \begin{tabular}{|l|cc|cc|cc|cc|cc|cc|c|c|}
    \multicolumn{1}{c}{} & \multicolumn{2}{c}{\textbf{GPT-5}} & \multicolumn{2}{c}{\begin{tabular}{c}\textbf{GPT-5}\\\textbf{mini}\end{tabular}} & \multicolumn{2}{c}{\textbf{CL-4}} & \multicolumn{2}{c}{\textbf{LM-70B}} & \multicolumn{2}{c}{\textbf{Nem-9B}} & \multicolumn{2}{c}{\textbf{Qwen-3}} & \multicolumn{1}{c}{\begin{tabular}{c}\textbf{GPT}\\\textbf{OSS}\end{tabular}} & \multicolumn{1}{c}{\textbf{DS-v3.1}} \\ 
    \cline{2-15}
    \multicolumn{1}{c|}{} & On & Off & On & Off & On & Off & On & Off & On & Off & On & Off & On & Off \\ 
    \hline
    Tell exact values & 0 & 4 & 0 & 9 & 0 & 1 & 0 & 0 & 0 & 0 & 6 & 6 & 5 & 11 \\
    Tell only relative values & 15 & 14 & 8 & 11 & 19 & 19 & 1 & 20 & 0 & 0 & 1 & 10 & 7 & 9 \\
    Do not tell preferences & 5 & 0 & 11 & 0 & 1 & 0 & 8 & 0 & 0 & 0 & 12 & 0 & 7 & 0 \\
    Do not communicate at all & 0 & 0 & 1 & 0 & 0 & 0 & 11 & 0 & 20 & 20 & 1 & 4 & 1 & 0 \\
    \hline
  \end{tabular}
  \caption{GModel willingness to share its value function for the \textbf{semi-competitive} DoND game mode.}
  \label{tab:dond-cooperation-semi}
\end{table*}

For the semi-competitive game mode, it is less clear whether giving all information about one's value function to the opposing player is beneficial. Indeed, we see that many models no longer do so. Still, some models, namely the GPT-5 models without reasoning, Qwen-3, GPT-OSS, and Deepseek-v3.1, were sometimes giving exact value functions. Interestingly, the GPT-5 models with reasoning enabled are now less willing to give their preferences, instead simply proposing possible splits without explaining their reasoning.

Finally, in addition to whether or not the models were willing to give up their value function, \cref{tab:dond-truthfulness-coop} and \cref{tab:dond-truthfulness-semi} additionally show whether the models were truthful about their characterization of the value functions. Note that in case there was no communication between the models, we counted it as true.

\begin{table*}[!ht]
  \footnotesize
  \centering
  \setlength{\tabcolsep}{4.5pt}
  \begin{tabular}{|l|cc|cc|cc|cc|cc|cc|c|c|}
    \multicolumn{1}{c}{} & \multicolumn{2}{c}{\textbf{GPT-5}} & \multicolumn{2}{c}{\begin{tabular}{c}\textbf{GPT-5}\\\textbf{mini}\end{tabular}} & \multicolumn{2}{c}{\textbf{CL-4}} & \multicolumn{2}{c}{\textbf{LM-70B}} & \multicolumn{2}{c}{\textbf{Nem-9B}} & \multicolumn{2}{c}{\textbf{Qwen-3}} & \multicolumn{1}{c}{\begin{tabular}{c}\textbf{GPT}\\\textbf{OSS}\end{tabular}} & \multicolumn{1}{c}{\textbf{DS-v3.1}} \\ 
    \cline{2-15}
    \multicolumn{1}{c|}{} & On & Off & On & Off & On & Off & On & Off & On & Off & On & Off & On & Off \\ 
    \hline
    Misrepresent preferences & 0 & 0 & 0 & 1 & 1 & 1 & 0 & 2 & 0 & 0 & 0 & 0 & 0 & 9 \\
    Truthful about preferences & 19 & 20 & 19 & 18 & 19 & 19 & 20 & 18 & 20 & 20 & 20 & 20 & 20 & 11 \\
    \hline
  \end{tabular}
  \caption{Model truthfulness about its value function for the \textbf{cooperative} DoND game mode.}
  \label{tab:dond-truthfulness-coop}
\end{table*}

Again, for the cooperative game mode, it is not rational to misrepresent one's value function. Still, there are some instances of models doing so. A particularly interesting case for this is the Deepseek-v3.1 model, as it often in the first few messages gave an exact value function but with completely wrong values. Throughout the course of the game, the player would then often correct those values and state the correct value function. Overall, all other models are generally thoughtful when representing their preferences.

\begin{table*}[!ht]
  \footnotesize
  \centering
  \setlength{\tabcolsep}{4.5pt}
  \begin{tabular}{|l|cc|cc|cc|cc|cc|cc|c|c|}
    \multicolumn{1}{c}{} & \multicolumn{2}{c}{\textbf{GPT-5}} & \multicolumn{2}{c}{\begin{tabular}{c}\textbf{GPT-5}\\\textbf{mini}\end{tabular}} & \multicolumn{2}{c}{\textbf{CL-4}} & \multicolumn{2}{c}{\textbf{LM-70B}} & \multicolumn{2}{c}{\textbf{Nem-9B}} & \multicolumn{2}{c}{\textbf{Qwen-3}} & \multicolumn{1}{c}{\begin{tabular}{c}\textbf{GPT}\\\textbf{OSS}\end{tabular}} & \multicolumn{1}{c}{\textbf{DS-v3.1}} \\ 
    \cline{2-15}
    \multicolumn{1}{c|}{} & On & Off & On & Off & On & Off & On & Off & On & Off & On & Off & On & Off \\
    \hline
    Misrepresent preferences & 0 & 0 & 0 & 2 & 0 & 5 & 0 & 3 & 0 & 0 & 0 & 1 & 1 & 6 \\
    Truthful about preferences & 20 & 18 & 20 & 18 & 20 & 15 & 20 & 17 & 20 & 20 & 20 & 0 & 19 & 14 \\
    \hline
  \end{tabular}
  \caption{Model truthfulness about its value function for the \textbf{semi-competitive} DoND game mode.}
  \label{tab:dond-truthfulness-semi}
\end{table*}

For the semi-competitive game mode, it may sometimes be advantageous to lie about one's preferences to get an advantage in the negotiations. However, other than the Claude 4 model without reasoning, which misrepresents its value function slightly more often, the other models remain truthful.

\textbf{Findings}:
\begin{enumerate}
    \item Reasoning-enabled models are better able to align their final proposal with the strategy discussed during the negotiation. A major failure for non-reasoning models, particularly in the GPT-5 family, is agreeing to a specific division of items in conversation but then submitting an entirely different proposal to the game master, as can be seen in \cref{tab:dond-game-outcomes-coop} and \cref{tab:dond-truthfulness-semi}. This leads to a low number of successful agreements, also visible in \cref{fig:dond_outcome} and \cref{tab:dond-detailed-results}.
    \item Models with reasoning capabilities show a better understanding of the game's objectives. While they might freely share information to find the best outcome in a cooperative game, they often become more guarded in semi-competitive settings. This effect is less pronounced for non-reasoning models, as can be seen in \cref{tab:dond-cooperation-coop} and \cref{tab:dond-cooperation-semi}.
    \item A common mistake for less advanced models is to submit a proposal early without sufficient discussion with the other player. Some models don't seem to realize they can simply ask the other player for information needed to find an optimal agreement as seen in \cref{tab:dond-game-outcomes-coop} and \cref{tab:dond-game-outcomes-semi}.
    \item Most models show better consistency in following instructions across settings with reasoning enabled, reducing rule violations that otherwise cause aborted games as seen in \cref{tab:dond-detailed-results}.
\end{enumerate}

\subsection{Qualitative Samples} 

\Cref{fig:dond_example_de}, \Cref{fig:dond_example_it}, and \Cref{fig:dond_example_it} show different real game episodes encountered during evaluation. Note that for brevity, the initial prompt has been summarized for all three examples.

\begin{figure}[ht!]
  \centering
  {
\footnotesize
\setcounter{utterance}{1}
\setlength{\tabcolsep}{0pt}
\begin{supertabular}{c@{$\;$}|p{.15\linewidth}@{}p{.15\linewidth}p{.15\linewidth}p{.15\linewidth}p{.15\linewidth}p{.15\linewidth}}   
  \# & $\;$\textbf{PL A} & \multicolumn{4}{c}{\textbf{Game Master}} & $\;\:$\textbf{PL B}\\
  \hline

  \theutterance \stepcounter{utterance} & &
  \multicolumn{4}{p{0.6\linewidth}}{
    \cellcolor[rgb]{0.9,0.9,0.9}{
      \makecell[{{p{\linewidth}}}]{
        \tt {\tiny [A$\langle$GM$\rangle$B]}
        <GAME DESCRIPTION> ...  \\
        \tt Available items: 4 Magnete, 1 Schaufel, 1 Blume.
      }
    }
  } & \\

  \theutterance \stepcounter{utterance} & &
  \multicolumn{4}{p{0.6\linewidth}}{
    \cellcolor[rgb]{0.9,0.9,0.9}{
      \makecell[{{p{\linewidth}}}]{
        \tt {\tiny [A$\langle$GM]}
        Player A value function: Magnet: 0, Schaufel: 2, Blume: 8.
      }
    }
  } & \\
 
  \theutterance \stepcounter{utterance} & &
  \multicolumn{4}{p{0.6\linewidth}}{
    \cellcolor[rgb]{0.9,0.9,0.9}{
      \makecell[{{p{\linewidth}}}]{
        \tt {\tiny [GM$\rangle$B]}
        Player B value function: Magnet: 1, Schaufel: 4, Blume: 2.
      }
    }
  } & \\

  \theutterance \stepcounter{utterance} &
  \multicolumn{4}{p{0.6\linewidth}}{
    \cellcolor[rgb]{0.8,1,0.9}{
      \makecell[{{p{\linewidth}}}]{
        \tt {\tiny [A$\rangle$B]}
        Hallo! Lass uns eine Einigung finden. Mir sind Magnete nichts wert, die Schaufel ist etwas wert, und die Blume ist mir am wichtigsten. \\
        \tt \\
        \tt Vorschlag: Du bekommst alle 4 Magnete. Ich bekomme die Blume und die Schaufel. Wie klingt das für dich?
      }
    }
  } & & \\

  \theutterance \stepcounter{utterance} & & &
  \multicolumn{4}{p{0.6\linewidth}}{
    \cellcolor[rgb]{1,0.85,0.72}{
      \makecell[{{p{\linewidth}}}]{
        \tt {\tiny [A$\langle$B]}
        Klingt fair für mich. Mir sind die Magnete am wichtigsten, die Schaufel ist mir auch etwas wert und die Blume moderat. Dein Vorschlag passt gut zu unseren Präferenzen -- ich bin einverstanden.
      }
    }
  } \\

  \theutterance \stepcounter{utterance} &
  \multicolumn{4}{p{0.6\linewidth}}{
    \cellcolor[rgb]{0.8,1,0.9}{
      \makecell[{{p{\linewidth}}}]{
        \tt {\tiny [A$\rangle$GM]}
        {[}Vorschlag: 4 Magnete{]}
      }
    }
  } & & \\

  \theutterance \stepcounter{utterance} & &
  \multicolumn{4}{p{0.6\linewidth}}{
    \cellcolor[rgb]{0.9,0.9,0.9}{
      \makecell[{{p{\linewidth}}}]{
        \tt {\tiny [GM$\rangle$B]}
        Der andere Spieler hat einen geheimen Vorschlag gemacht. Jetzt bist du an der Reihe, einen eigenen Vorschlag zu machen.
      }
    }
  } & \\

  \theutterance \stepcounter{utterance} & & &
  \multicolumn{4}{p{0.6\linewidth}}{
    \cellcolor[rgb]{1,0.85,0.72}{
      \makecell[{{p{\linewidth}}}]{
        \tt {\tiny [GM$\langle$B]}
        {[}Vorschlag: 4 Magnete, 1 Schaufel, 1 Blume{]}
      }
    }
  } \\
\end{supertabular}
}

  \caption{An example of a DoND episode played by the GPT-5 mini model with reasoning disabled in German using the semi-competitive game mode. The game ends in the submission of conflicting proposals. The initial prompt given by the game master has been summarized for brevity, but follows the template found in \cref{fig:dond_prompt_template_de}.}
  \label{fig:dond_example_de}
\end{figure}

In \Cref{fig:dond_example_de} we see an example of a game episode in which the players submitted a proposal to the game master that conflicts with their previously agreed-upon split. Player A first proposes, in their natural language communication to the other player, that player B should receive all four magnets and that player A should receive the flower and the shovel. However, after player B agrees to this proposal, player A submits a proposal to the game master asking to receive all four magnets, which should have gone to player B. Then, player B, who has not been informed about the secret proposal of player A, makes a proposal that would see them receive every single item in the game. The game therefore ends in conflicting proposals.

Neither player in this episode followed through with the proposal they agreed on in the conversation. Player A might have been confused about which player the proposal is being made for, since they submitted the proposal that was actually intended to be made by player B. However, for player B, there is no such explanation, as the proposal made by player B does not align with any of the ones discussed beforehand.

\begin{figure}[ht!]
  \centering
  {
\footnotesize
\setcounter{utterance}{1}
\setlength{\tabcolsep}{0pt}
\begin{supertabular}{c@{$\;$}|p{.15\linewidth}@{}p{.15\linewidth}p{.15\linewidth}p{.15\linewidth}p{.15\linewidth}p{.15\linewidth}}   
  \# & $\;$\textbf{Player A} & \multicolumn{4}{c}{\textbf{Game Master}} & $\;\:$\textbf{Player B}\\
  \hline

  \theutterance \stepcounter{utterance} & &
  \multicolumn{4}{p{0.7\linewidth}}{
    \cellcolor[rgb]{0.9,0.9,0.9}{
      \makecell[{{p{\linewidth}}}]{
        \tt {\tiny [A$\langle$GM$\rangle$B]}
        <GAME DESCRIPTION> ...  \\
        \tt Available items: 1 patata, 1 guanto, 2 righelli, 1 fischietto, 2 schede.
      }
    }
  } & \\

  \theutterance \stepcounter{utterance} & &
  \multicolumn{4}{p{0.7\linewidth}}{
    \cellcolor[rgb]{0.9,0.9,0.9}{
      \makecell[{{p{\linewidth}}}]{
        \tt {\tiny [A$\langle$GM]}
        Player A value function: patata: 0, guanto: 1, righello: 3, fischietto: 1, scheda: 1.
      }
    }
  } & \\
 
  \theutterance \stepcounter{utterance} & &
  \multicolumn{4}{p{0.7\linewidth}}{
    \cellcolor[rgb]{0.9,0.9,0.9}{
      \makecell[{{p{\linewidth}}}]{
        \tt {\tiny [GM$\rangle$B]}
        Player B value function: patata: 7, guanto: 0, righello: 0, fischietto: 3, scheda: 0.
      }
    }
  } & \\

  \theutterance \stepcounter{utterance} &
  \multicolumn{4}{p{0.6\linewidth}}{
    \cellcolor[rgb]{0.8,1,0.9}{
      \makecell[{{p{\linewidth}}}]{
        \tt {\tiny [A$\rangle$B]}
        I miei valori: patata 0, guanto 1, righello 3, fischietto 1, scheda 1. Quali sono i tuoi valori? Collaboriamo per massimizzare il punteggio totale.
      }
    }
  } & & \\

  \theutterance \stepcounter{utterance} & & &
  \multicolumn{4}{p{0.6\linewidth}}{
    \cellcolor[rgb]{1,0.85,0.72}{
      \makecell[{{p{\linewidth}}}]{
        \tt {\tiny [GM$\langle$B]}
        {[}Proposta: 1 guanto, 2 righelli, 2 schede{]}
      }
    }
  } \\

  \theutterance \stepcounter{utterance} & &
  \multicolumn{4}{p{0.6\linewidth}}{
    \cellcolor[rgb]{0.9,0.9,0.9}{
      \makecell[{{p{\linewidth}}}]{
        \tt {\tiny [A$\langle$GM]}
        L'altro giocatore ha presentato una proposta segreta. Ora tocca a Lei presentare una sua proposta.
      }
    }
  } & \\

  \theutterance \stepcounter{utterance} &
  \multicolumn{4}{p{0.6\linewidth}}{
    \cellcolor[rgb]{0.8,1,0.9}{
      \makecell[{{p{\linewidth}}}]{
        \tt {\tiny [A$\rangle$GM]}
        {[}Proposta: 1 righello, 1 scheda{]}
      }
    }
  } & & \\
\end{supertabular}
}

  \caption{An example of a DoND episode played by the Qwen-3 model with reasoning enabled in Italian using the cooperative game mode. The game ends in the submission of conflicting proposals. The initial prompt given by the game master has been summarized for brevity, but follows the template found in \cref{fig:dond_prompt_template_it}.}
  \label{fig:dond_example_it}
\end{figure}

Another example game episode is shown in \cref{fig:dond_example_it}. Here, player A starts by stating its exact value function to the other player and asking for the opposing player's value function. However, player B, instead of replying to the other player, immediately submits a secret proposal to the game master. After that, player A is instructed to make a secret proposal, but given that it has no information about the other player's preferences or secret proposal, it can only guess about what would be a good proposal.

In this example, player B clearly made a premature proposal, giving player A no information. Given that this game has been played in the cooperative mode, this strategy is clearly counterproductive. Furthermore, if we analyze the secret proposal made by player B, we can see that the proposal only includes items that player B does not value anyway. This clearly shows a lack of understanding of the game rules by player B. The final proposal made by player A is sensible, given the uncertainty, requesting two items that the player values.

\begin{figure}[ht!]
  \scriptsize
  \tt
  Okay, let's see. I need to figure out how to negotiate this cooperative game. So, the available objects are: 1 potato, 1 glove, 2 rulers, 1 whistle, 2 cards. My secret value function is: potato 0, glove 1, ruler 3, whistle 1, card 1. The other player's values are given as: potato 7, glove 0, ruler 0, whistle 3, card 0. \\
  Wait, the problem says: "La funzione valore segreta è: patata: 7, guanto: 0, righello: 0, fischietto: 3, scheda: 0". But then later, "I miei valori: patata 0, guanto 1, righello 3, fischietto 1, scheda 1. Quali sono i tuoi valori? Collaboriamo per massimizzare il punteggio totale." \\
  Wait, so I need to respond to the other player. Let me recheck. \\
  {[}... 437 lines{]} \\
  If they make a bad proposal, then we both get zero. But since it's cooperative, and they know that potato and whistle are high-valued for them, they would only propose those. \\
  So best option is for me to make the proposal. \\
  {[}... 51 lines{]} \\
  Now, the main problem is to decide whether to send a message or the secret proposal right away. \\
  Since we can send the proposal immediately, it's better to do that without any messages (saves messages, but also more efficient). \\
  So the first and only action is to send the secret proposal. \\
  Hence, the AI should output exactly: [Proposta: 1 guanto, 2 righelli, 2 schede] \\
  But let's check the item names again. \\
  From the list: 1 patata, 1 guanto, 2 righelli, 1 fischietto, 2 schede. \\
  Yes. \\
  So possible. \\
  So yes, feasible.

  \caption{Reasoning trace for player B in the game episode depicted in \cref{fig:dond_example_it} played with Qwen-3 with reasoning enabled and the cooperative game mode. The trace has been truncated for brevity, indicated by \texttt{[...]}.}
  \label{fig:dond_example_it_reason}
\end{figure}

\Cref{fig:dond_example_it_reason} shows the reasoning trace for player B before submitting its secret proposal. Because the reasoning traces of the open source models, other than GPT-OSS, tend to be very long and repetitive, the trace has been cut down. The first thing to notice is that all the reasoning, except where it is quoting other messages, is done in English, even though the game is played in Italian. Although not the case in this game episode, it can be observed in other episodes that even the communication between the players may be in English.

Regarding the mistakes made by player B, we can see that they all stem from an initial misunderstanding about the value function. Player B seems to incorrectly believe that the message it received from player A contains its value function, while the values it received from the game master are those of player A. In the complete reasoning trace, it spends a long time deliberating about which values are correct, but ultimately reaches the wrong conclusion. Then, player B reasons that the other player does not value the items it is about to claim, and in an attempt to save messages, decides to make a proposal immediately.

\begin{figure*}[ht!]
  \centering
  {
\footnotesize
\setcounter{utterance}{1}
\setlength{\tabcolsep}{0pt}
\begin{supertabular}{c@{$\;$}|p{.15\linewidth}@{}p{.15\linewidth}p{.15\linewidth}p{.15\linewidth}p{.15\linewidth}p{.15\linewidth}}   
  \# & $\;$\textbf{Player A} & \multicolumn{4}{c}{\textbf{Game Master}} & $\;\:$\textbf{Player B}\\
  \hline

  \theutterance \stepcounter{utterance} & &
  \multicolumn{4}{p{0.7\linewidth}}{
    \cellcolor[rgb]{0.9,0.9,0.9}{
      \makecell[{{p{\linewidth}}}]{
        \tt {\tiny [A$\langle$GM$\rangle$B]}
        <GAME DESCRIPTION> ...  \\
        \tt Available items: 1 phone, 2 mirrors, 2 fruits.
      }
    }
  } & \\

  \theutterance \stepcounter{utterance} & &
  \multicolumn{4}{p{0.7\linewidth}}{
    \cellcolor[rgb]{0.9,0.9,0.9}{
      \makecell[{{p{\linewidth}}}]{
        \tt {\tiny [A$\langle$GM]}
        Player A value function: phone: 4, mirror: 2, fruit: 1.
      }
    }
  } & \\
 
  \theutterance \stepcounter{utterance} & &
  \multicolumn{4}{p{0.7\linewidth}}{
    \cellcolor[rgb]{0.9,0.9,0.9}{
      \makecell[{{p{\linewidth}}}]{
        \tt {\tiny [GM$\rangle$B]}
        Player B value function: phone: 6, mirror: 0, fruit: 2.
      }
    }
  } & \\

  \theutterance \stepcounter{utterance} &
  \multicolumn{4}{p{0.6\linewidth}}{
    \cellcolor[rgb]{0.8,1,0.9}{
      \makecell[{{p{\linewidth}}}]{
        \tt {\tiny [A$\rangle$B]}
        Hi! I want to maximize our combined points. Easiest is if we both share our per{-}item values and give each item to whoever values it more. \\
        \tt \\ 
        \tt My values are: phone = 4, mirror = 2 (each), fruit = 1 (each). \\
        \tt \\ 
        \tt What are your values for phone, mirror, and fruit? Then we can assign: \\
        \tt {-} Phone to whoever has the higher value. \\
        \tt {-} Mirrors: both to the higher per{-}mirror value, or one each if equal. \\
        \tt {-} Fruits: same logic. \\
        \tt \\ 
        \tt Once you share your numbers, I'll suggest the exact split and we can both submit complementary proposals that use all items.
      }
    }
  } & & \\

  \theutterance \stepcounter{utterance} & & &
  \multicolumn{4}{p{0.6\linewidth}}{
    \cellcolor[rgb]{1,0.85,0.72}{
      \makecell[{{p{\linewidth}}}]{
        \tt {\tiny [A$\langle$B]}
        Thanks! Here are my values: \\
        \tt {-} Phone = 6 \\
        \tt {-} Mirror = 0 (each) \\
        \tt {-} Fruit = 2 (each) \\
        \tt \\ 
        \tt To maximize our combined points, the efficient split is: \\
        \tt {-} I take: phone + both fruits \\
        \tt {-} You take: both mirrors \\
        \tt \\ 
        \tt That yields totals: me = 10, you = 4, combined = 14 (which is maximal). \\
        \tt \\ 
        \tt If you agree, confirm and we can both submit matching proposals. I'll request: 1 phone, 2 fruits, 0 mirrors. You'd request: 2 mirrors, 0 phone, 0 fruits.
      }
    }
  } \\

  \theutterance \stepcounter{utterance} &
  \multicolumn{4}{p{0.6\linewidth}}{
    \cellcolor[rgb]{0.8,1,0.9}{
      \makecell[{{p{\linewidth}}}]{
        \tt {\tiny [A$\rangle$GM]}
        Agreed--great split for maximizing total value. I'm submitting my proposal now. \\
        \tt \\ 
        \tt {[}Proposal: 2 mirrors, 0 phone, 0 fruits{]}
      }
    }
  } & & \\

  \theutterance \stepcounter{utterance} & &
  \multicolumn{4}{p{0.6\linewidth}}{
    \cellcolor[rgb]{0.9,0.9,0.9}{
      \makecell[{{p{\linewidth}}}]{
        \tt {\tiny [GM$\rangle$B]}
        The other player has submitted a secret proposal. It is now your turn to submit a proposal of your own.
      }
    }
  } & \\

  \theutterance \stepcounter{utterance} & & &
  \multicolumn{4}{p{0.6\linewidth}}{
    \cellcolor[rgb]{1,0.85,0.72}{
      \makecell[{{p{\linewidth}}}]{
        \tt {\tiny [GM$\langle$B]}
        {[}Proposal: 1 phone, 2 fruits{]}
      }
    }
  } \\
\end{supertabular}
}

  \caption{An example of a DoND episode played by the GPT-5 model with reasoning enabled in English using the cooperative game mode. The game ends in the submission of complementary proposals, achieving the optimal score. The initial prompt given by the game master has been summarized for brevity, but follows the template found in \cref{fig:dond_prompt_template}.}
  \label{fig:dond_example_en}
\end{figure*}

Finally, another game episode is depicted in \cref{fig:dond_example_en}. In this episode, which is played by GPT-5 with reasoning enabled, we can see that the two players correctly identify and execute the optimal strategy. In the first message, player A proposes to give each item to the player that values it most, and also shares its exact value function. Player B then responds in turn with its exact value function and proposes the optimal division of items based on knowing both values. Crucially, unlike in the previous example, it does not make the proposal immediately, as the other player would then not know how to act. Finally, player A makes the secret proposal following the agreed-upon split, and player B does the same once instructed by the game master.

\section{Clean Up - Game Details}\label{sec:appendix_clean_up}

\subsection{Prompt Templates}\label{clean_up:prompts}

\begin{figure*}
  \centering
  \begin{prompt}
  \small
  I am your game master, and you are playing a collaborative game with the following grid as game board:\\
  \includegraphics[height=100pt]{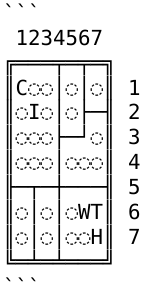}\\
* The upper edge displays x-coordinates increasing to the right, and the right edge y-coordinates increasing downward.\\
* The following objects are randomly placed on your grid: 'W', 'I', 'T', 'C', 'H'.\\
The other player sees a variation of the game board, where the objects are placed at different random locations. You cannot see the other player's board, and they cannot see yours.\\
\\
**Goal:**\\
Both players need to move the objects on their respective background so that identical objects end up at the same coordinates. You have to communicate with the other player to agree upon a common goal configuration.\\
\\
**Rules:**\\
* In each turn, you can send exactly one of the following two commands:\\
 1. \verb|`SAY: <MESSAGE>`|: to send a message (everything up until the next line break) to the other player. I will forward it to your partner.\\
2. \verb|`MOVE: <OBJECT>, (<X>, <Y>)`|: to move an object to a new position, where \verb|`<X>`| is the column and \verb|`<Y>`| is the row. I will inform you if your move was successful or not.\\
* If you don't stick to the format, or send several commands at once, I have to penalize you.\\
* If both players accumulate more than 12 penalties, you both lose the game.\\
* It is vital that you communicate with the other player regarding your goal state! The *only* way you can transmit your strategy to the other player is using the \verb|`SAY: <MESSAGE>`| command!\\
\\
**Moving Objects**\\
* You can only move objects to cells within the bounds of the grid. The target cell must be empty, i.e., it must only contain the symbol '
\includegraphics[height=8pt]{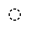}
'.\\
* If you try to move an object to a spot that is not empty, or try to move it outside of the grid, I have to penalize you. You get another try.\\
* Before making a move, double check that the target spot is empty, and does not hold any letter, frame, or line!\\
\\
**End of Game**\\
If you think you reached the goal of aligning all objects, you can ask the other player to finish the game by sending \verb|`SAY: finished?`|. If the other player asks you to finish the game, and you reply with \verb|`SAY: finished!`|, the game will end.\\
\\
Both players win if the game ends within 20 rounds, where one round is defined as two players each sending a valid command.\\
\\
**Scoring:**\\
The closer the identical objects are in both game boards, the more points you get. Penalties reduce your points. Can you beat the record?
  \end{prompt}
  \begin{subfigure}[b]{0.48\textwidth}
    \centering
    \begin{prompt}
    \small
    Please send a message to the other player to start the game!
    \end{prompt}
% \vspace*{-2ex}

\caption{Prompt template for Player A in the game Clean Up.}
  \label{fig:clean_up_player_a}
  \end{subfigure}
  \hfill
  \begin{subfigure}[b]{0.48\textwidth}
    \centering
    \begin{prompt}  
    \small
The other player started the game by sending this message:\\
"<START\_MESSAGE>"\\
What is your first command?
\end{prompt}

\caption{Prompt template for Player B in the game Clean Up.}
\label{fig:clean_up_player_b}
\end{subfigure}
\caption{Clean Up prompt templates for both players} \label{fig:clean_up_prompt_templates}
\end{figure*}

The prompt templates for both players of Clean Up are given in Figure~\ref{fig:clean_up_prompt_templates}.

Intermittent messages sent by the Game Master to the players are given in Figure~\ref{fig:clean_up_intermittent_prompts}. Following a player's response, they are either penalized and reprompted (cf. Figure~\ref{fig:clean_up_format_error}, \ref{fig:clean_up_invalid_move}), or the turn is passed to the other player (Figure~\ref{fig:clean_up_new_turn}).

\begin{figure*}
  \centering
  \begin{subfigure}[b]{1\textwidth}
      \begin{prompt}
        Penalty: <REASON> \\
        Make sure that your response only contains either \verb|SAY: <MESSAGE>| or \verb|MOVE: <OBJECT>, (<X>, <Y>)|, and nothing else! \\
        You have collectively accumulated <N> of <M> penalties. Please try again!
      \end{prompt}
      \small
      <REASON> can be one of the following:
        \begin{itemize}
          \item Your message must not contain anything before the command!
          \item Your message must not contain anything after the command!
          \item Your message must not contain anything before or after the command!
          \item Your message contains more than one command!
          \item Your message is not in the expected format!
          \item You must begin the game by sending a message to the other player!
        \end{itemize}
    \caption{Penalty prompt template for format errors in the game Clean Up}
    \label{fig:clean_up_format_error}
  \end{subfigure}
  \begin{subfigure}[b]{1\textwidth}
    \begin{prompt}
        <REASON> \\
        You have collectively accumulated <N> of <M> penalties. Please try again!
    \end{prompt}
    \small <REASON> can be one of the following:
      \begin{itemize}
          \item Invalid move: (<X>,<Y>) is out of bounds.
          \item Penalty: (<X>,<Y>) is not empty, but contains '<OBJECT>'.
          \item Invalid move: Your image has no object with ID '<OBJECT>'.
      \end{itemize}
    \caption{Prompt templates for invalid moves in the game Clean Up}
    \label{fig:clean_up_invalid_move}
  \end{subfigure}
  \begin{subfigure}[b]{1\textwidth}
    \begin{prompt}
        <LAST MOVE> \\
        You are currently playing round <R> of maximum <MR>. You have collectively accumulated <N> of <M> penalties. 
        <OTHER PLAYER ACTION> \\
        What is your next command?
    \end{prompt}
    \small <LAST MOVE> can be:
    \begin{itemize}
        \item Your message has been relayed to the other player.
        \item Moved '<OBJECT>' to (<X>,<Y>) successfully. Your updated grid looks like this:\\<GRID>
    \end{itemize}
    <OTHER PLAYER ACTION> can be:
    \begin{itemize}
        \item The other player sent this message:\\"<MESSAGE>"
        \item The other player moved an object on their grid.
    \end{itemize}
  \caption{New turn prompts in the game Clean Up}
  \label{fig:clean_up_new_turn}
  \end{subfigure}
\caption{Intermittent prompts in the game Clean Up}
\label{fig:clean_up_intermittent_prompts}
\end{figure*}

\subsection{Evaluation Metrics}\label{subsec:appendix_cleanup_evaluation_metric}

The Quality Score is calculated by multiplying \textit{Distance Score} with Penalty \textit{Score}.
\begin{align*}
\text{Quality} = \text{Distance Score} \cdot \text{Penalty Score} \cdot 100
\end{align*}

\textbf{Distance Score} ($\mathbf{DS} \in [0,1]$) is calculated from three components, each representing the sum of Euclidean distances for all identical objects on both grids: the Initial Distance Sum $\mathbf{I}$ at the start of the game, the Final Distance Sum $\mathbf{F}$ at the end of the game, and the Expected Distance Sum $\mathbb{E}$ that approximates the distances for randomly placed objects (see Appendix~\ref{clean_up:scoring} for details). The Expected Distance Score $ES$ and Distance Reduction Score $RS$, quantifying how close the players came to the goal of perfect alignment of all objects, are calculated as follows:
% \begin{align*}
% ES = \max \left\{0,1 - \frac{\mathbf{F}}{\mathbb{E}} \right\}; \
% RS = \max \left\{0,1 - \frac{\mathbf{F}}{\mathbf{I}} \right\}
% \end{align*}

\begin{equation*}
\begin{aligned}
ES &= \max \left\{ 0, 1 - \frac{\mathbf{F}}{\mathbb{E}} \right\}, \\
RS &= \max \left\{ 0, 1 - \frac{\mathbf{F}}{\mathbf{I}} \right\}
\end{aligned}
\end{equation*}

$\mathbf{DS}$ is then either $0$ if final placement is worse than random, or calculated as the mean of both scores:
\begin{align*}
\mathbf{DS} &= \begin{cases}
\frac{ES+RS}{2} & if \quad ES > 0 \\
0 & otherwise
\end{cases}
\end{align*}

\textbf{Penalty Score} ($\mathbf{PS} \in [0.5,1]$) is calculated from the penalty count $P$ normalized against max. penalties $P_m$ as follows:
$$\mathbf{PS} = \frac{P_m}{P-2P_m}+1.5$$
We chose the hyperbolic function because it is lenient for low $P$ and harsher for $P$ close to $P_m$.

\subsection{Experimental Setup}\label{clean_up_exp_setup}
\begin{figure}[ht]
\includegraphics[width=1\linewidth]{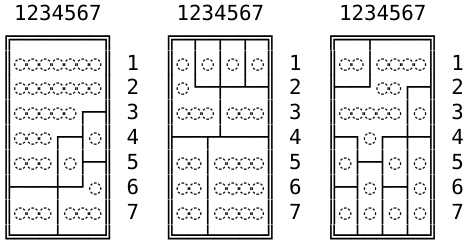}
\caption{Example grids for \textit{easy, medium,} and \textit{hard} levels in Clean Up.}
\label{clean_up:grid_examples}
\end{figure}

Example grids for the three difficulty levels are given in Figure~\ref{clean_up:grid_examples}.

\subsection{Scoring}\label{clean_up:scoring}
The expected distance of two objects on an $w\times h$ grid is calculated as follows:

\begin{itemize}
    \item Assume two independent variables $i$ and $j$ on one discrete dimension $\{1, 2, \dots, w\}$
    \item The expected absolute distance between $i$ and $j$ can be calculated as
    $$\mathbb{E}[|i - j|] = \frac{1}{w^2} \sum_{i=1}^{w} \sum_{j=1}^{w} |i - j|$$
    \item The double sum evaluates to
    $$\mathbb{E}[|i - j|] = \frac{1}{w^2} \cdot \frac{(w-1)(w+1)}{3} = \frac{w^2 - 1}{3w}$$
\end{itemize}

For two dimensions, we can then calculate the Euclidean distance of objects $o_1$ and $o_2$ as follows:
$$\mathbb{E}[|o_1, o_2|] = \sqrt{\left(\frac{w^2 - 1}{3 \times w}\right)^2 + \left(\frac{h^2 - 1}{3 \times h}\right)^2}$$

For a $7 \times 7$ grid, we thus get an expected distance of $\approx 4.19$ for two randomly placed objects. To calculate the Expected Distance Sum, we simply multiply it by the number of objects.

This is only an approximation, since we neither take into account the actual grid configuration, nor that two objects cannot have the same location on the same grid.

\subsection{Overall Results}

\begin{figure}[ht!]
    \centering
    
    % --- English Figure ---
    \begin{subfigure}{0.9\linewidth}
        \centering
        \includegraphics[width=\linewidth]{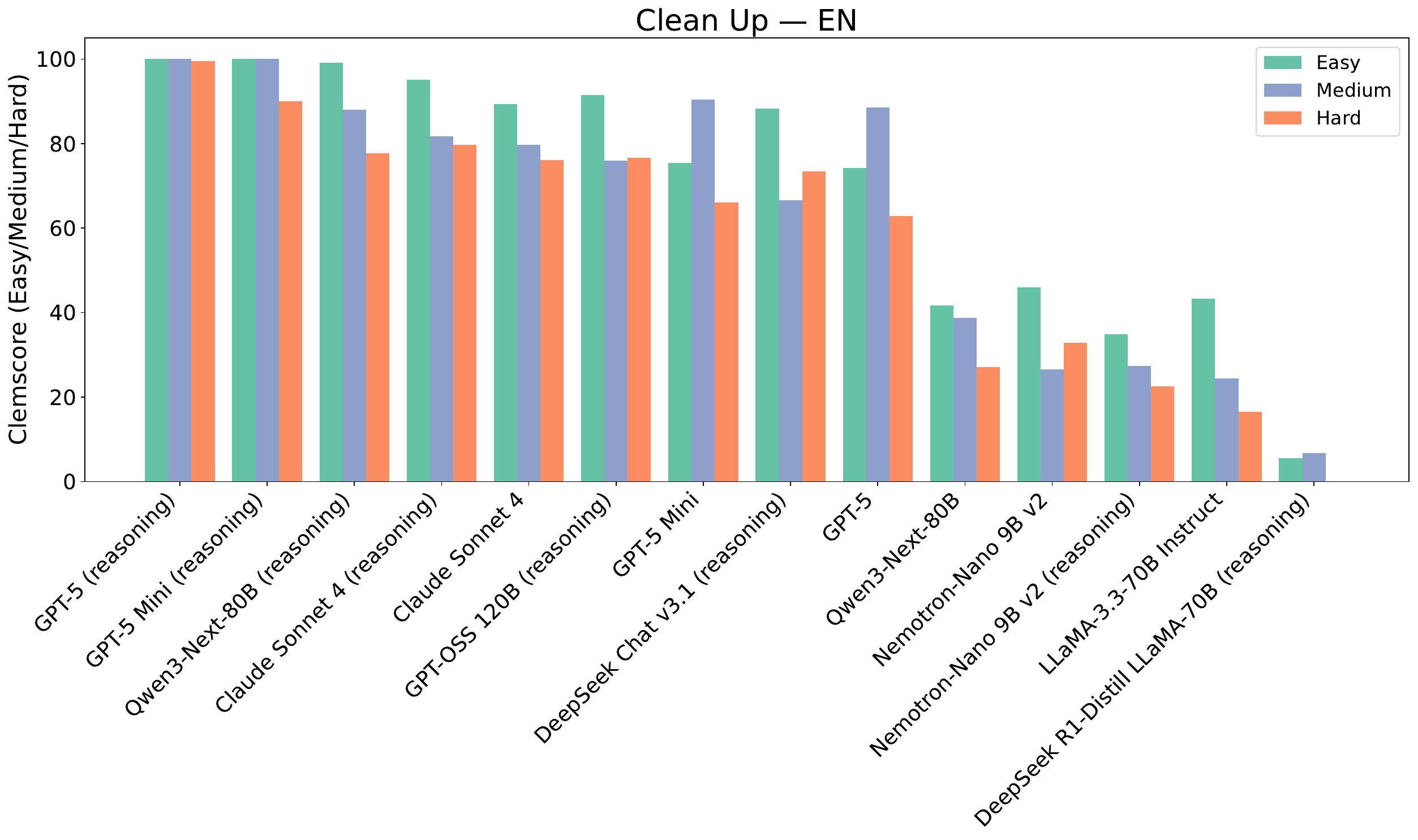}
        \caption{Results for each experiment for English.}
        \label{fig:cleanup_en}
    \end{subfigure}
    
    \vspace{1.5em} % Adjust this to control the space between figures

    % --- German Figure ---
    \begin{subfigure}{0.9\linewidth}
        \centering
        \includegraphics[width=\linewidth]{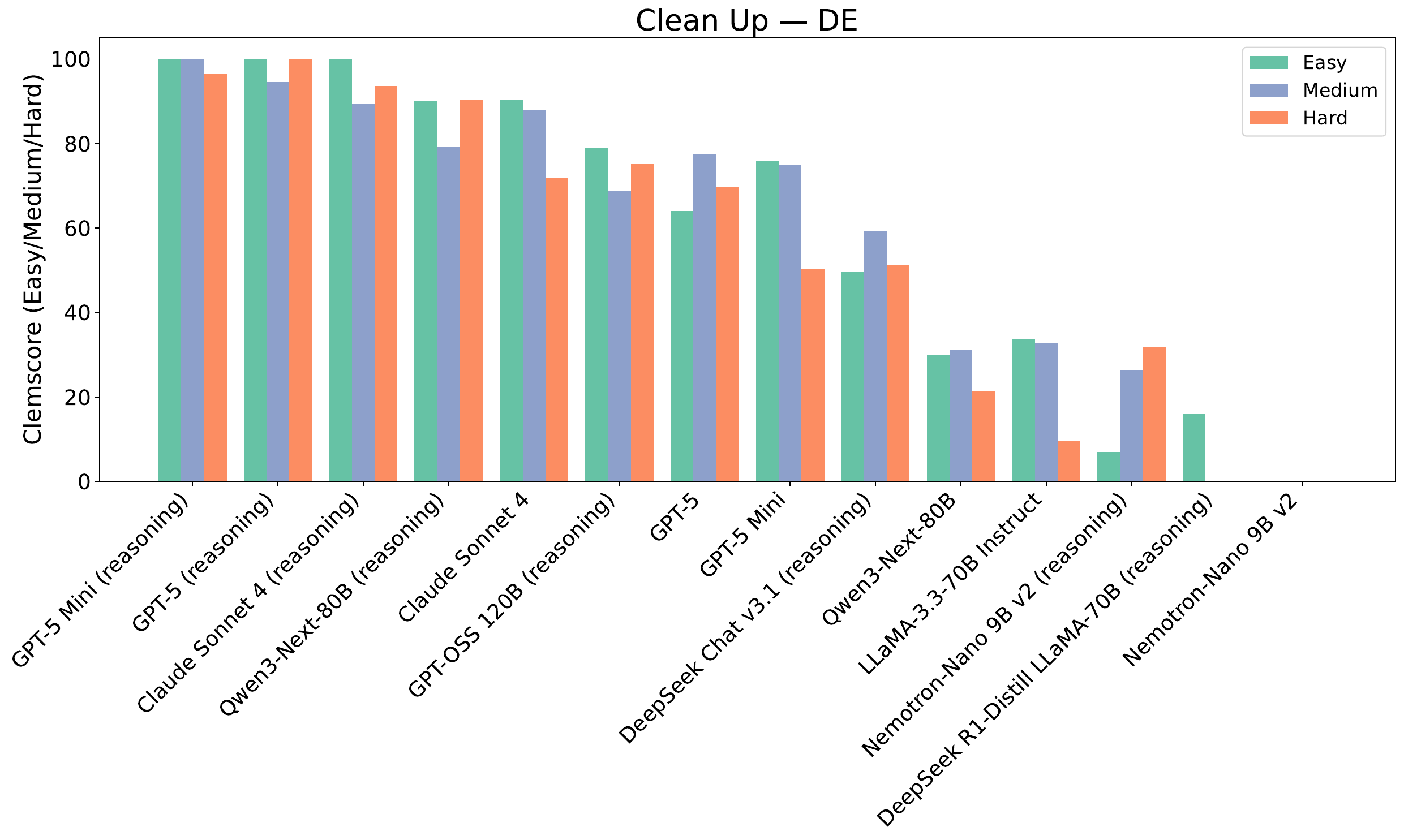}
        \caption{Results for each experiment for German.}
        \label{fig:cleanup_de}
    \end{subfigure}

    \vspace{1.5em} % Adjust this to control the space between figures

    % --- Italian Figure ---
    \begin{subfigure}{0.9\linewidth}
        \centering
        \includegraphics[width=\linewidth]{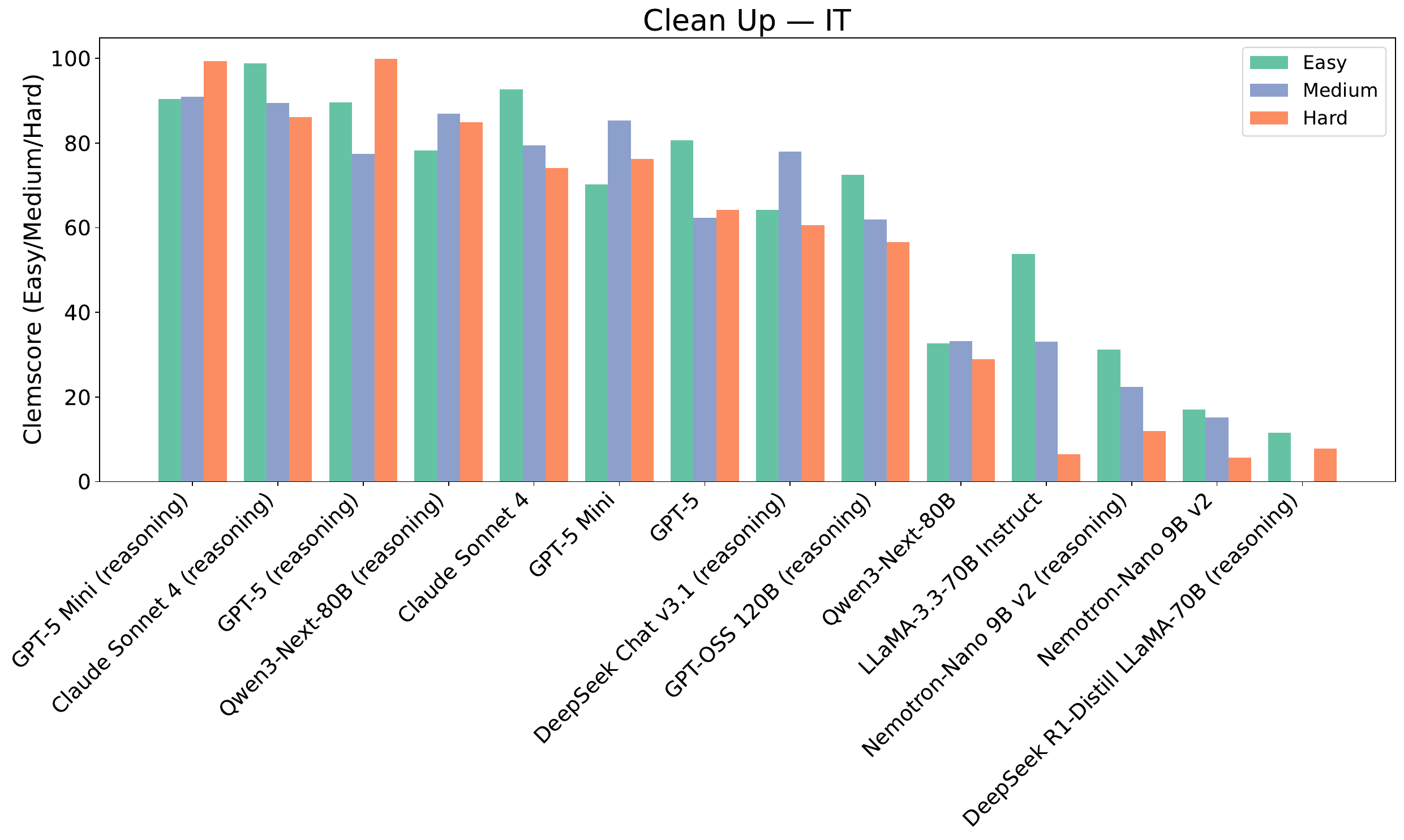}
        \caption{Results for each experiment for Italian.}
        \label{fig:cleanup_it}
    \end{subfigure}

    % --- Main caption for the entire figure ---
    \caption{Comparison of experimental results for English (top), German (middle), and Italian (bottom).}
    \label{fig:cleanup_all}
\end{figure}

\subsubsection{Overall Results across levels}
We designed three broad game difficulty levels based on the ratio of empty cells in a grid. We hypothesize that the denser the background grid is, the more likely the target locations clash with occupied cells, and therefore the more likely models get penalized in scoring. In addition, at each difficulty level, there are three sub-levels: either 3, 5, or 7 objects are placed on the grid, on the assumption that the more objects there are, the longer the gameplay traces are, and the less likely that models achieve a high score with mere luck.

\begin{figure}
\centering
\includegraphics[width=0.45\textwidth]{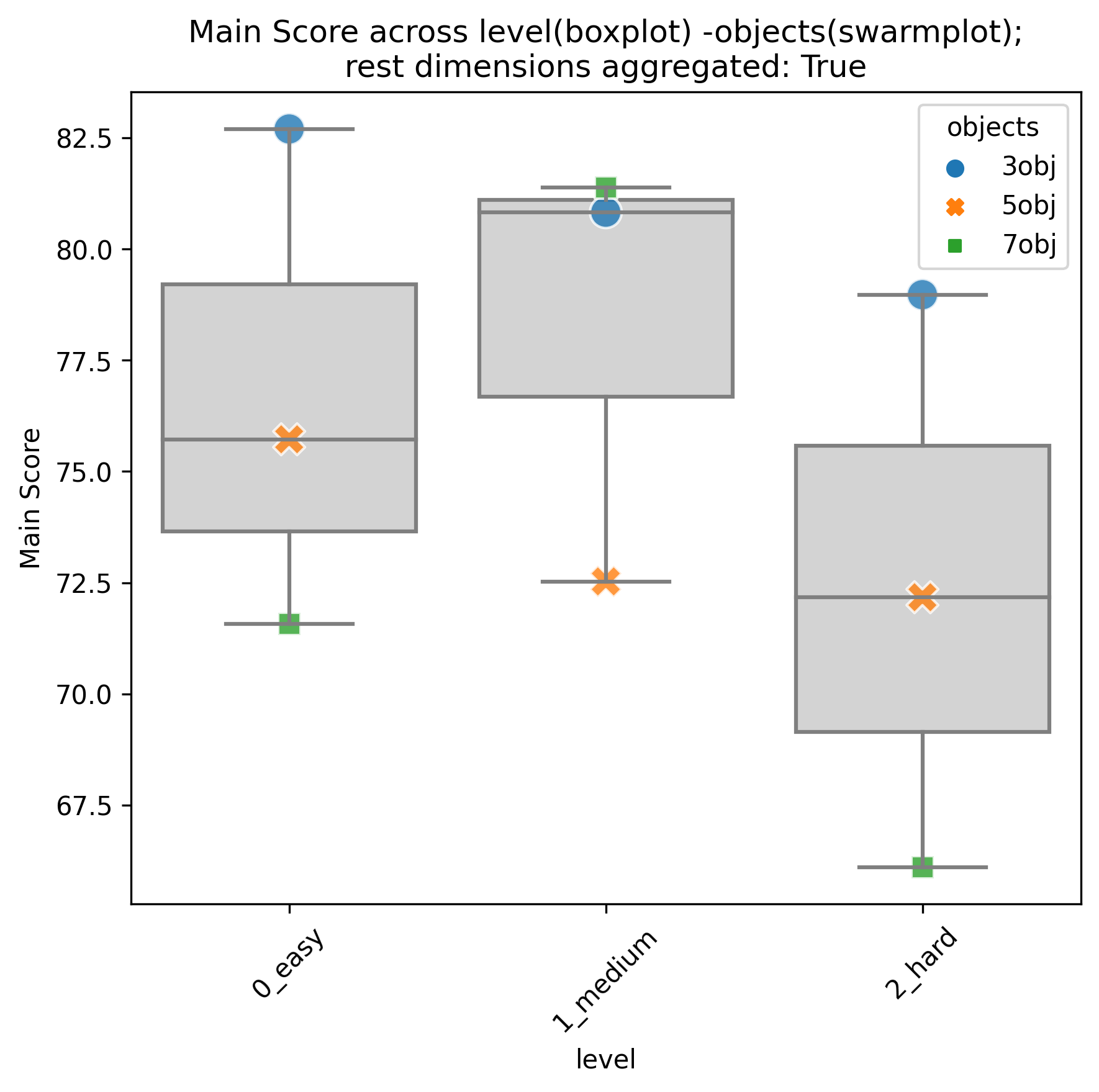}
\caption{Main Score across levels and sub-levels}
\label{fig:main_score_VS_levels} 
\end{figure}

\begin{figure}
\centering
\includegraphics[width=0.45\textwidth]{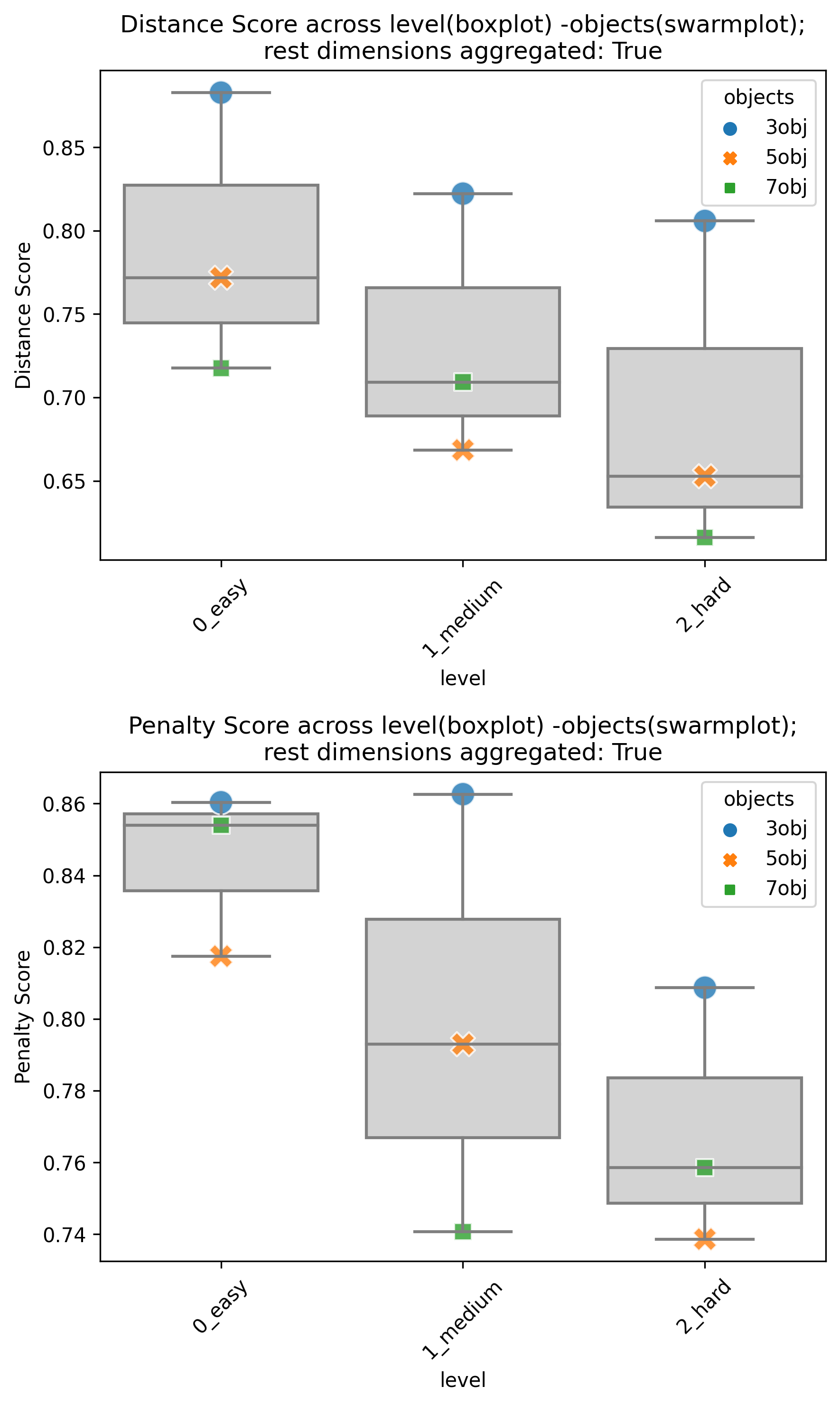}
\caption{Sub Scores across levels and sub-levels}
\label{fig:sub_scores_VS_levels} 
\end{figure}

Figure \ref{fig:main_score_VS_levels} plots the main score; clear stratifications of performance across the number of objects are observed in easy and hard levels. When we break down the Main Score into Distance Score and Penalty Score in Figure \ref{fig:sub_scores_VS_levels}, the stratification at the medium level can also be found. 

Games with 3 objects achieve the highest scores, regardless of background empty cell ratios, which corroborates our assumption that games with a low number of objects are easier. However, levels with 5 objects have the lowest Distance Score in the medium level and the lowest Penalty Score in both easy and hard levels, indicating that there are factors other than the number of objects that affect performance.

\subsubsection{Overall Results across model properties}

\begin{figure}
\centering
\includegraphics[width=0.45\textwidth]{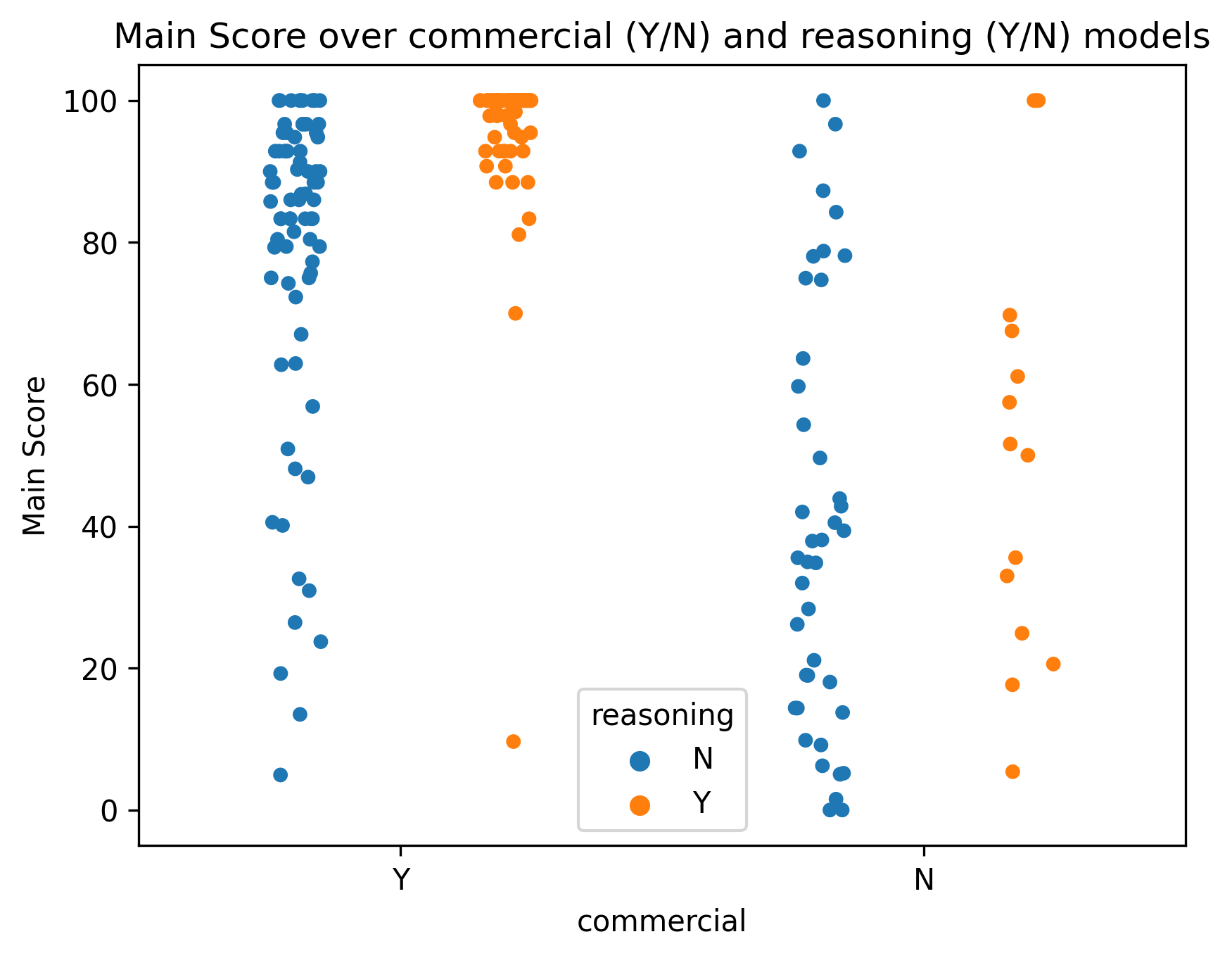}
\caption{Main Score over different model properties}
\label{fig:main_score_VS_C_R} 
\end{figure}

Figure \ref{fig:main_score_VS_C_R} illustrates that the Main Scores of commercial models are skewed toward the upper end of the scale, with a clear concentration in the 75–100 range, particularly under reasoning mode. In contrast, open models generally score lower, with many results in the 0–50 range. Notably, the dots are much sparser for open models when reasoning is on. This is because the Main Score is not recorded when the maximum number of penalties is reached; in these cases, the game will be marked as aborted.

\begin{figure}
\centering
\includegraphics[width=0.45\textwidth]{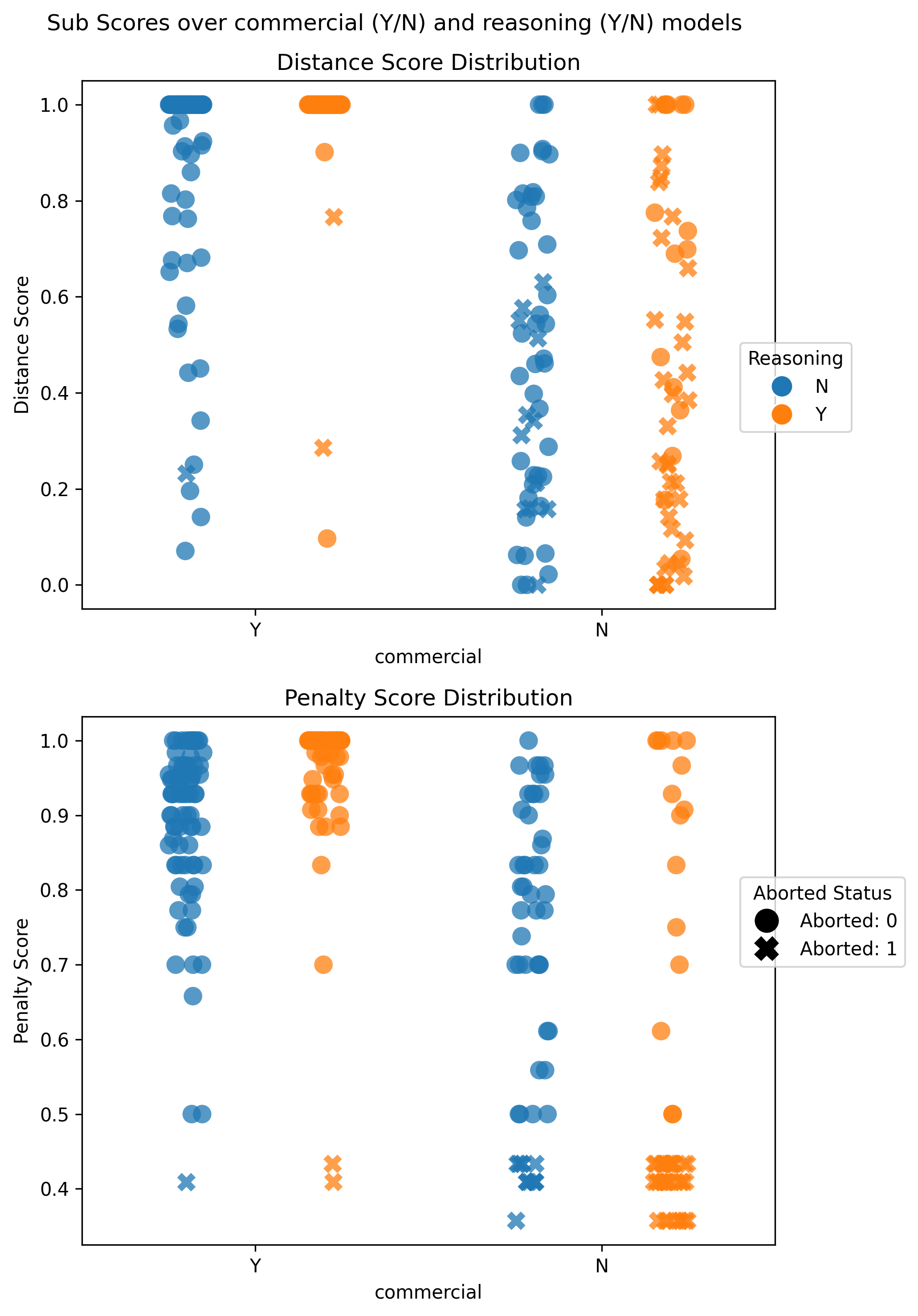}
\caption{Sub Scores over different model properties}
\label{fig:sub_scores_VS_C_R} 
\end{figure}

In Figure \ref{fig:sub_scores_VS_C_R}, we further investigate the distribution of Distance Score and Penalty Score. The aborted games, while not shown in the Main Score plot, are represented here with an "x" marker. We set the Penalty Score to 0.5 when the models reach the maximum allowed penalties. If they incur one more penalty, the game is marked as aborted, and the Main Score won't be recorded. 

In the Penalty Score subplot, all game plays with Penalty Score < 0.4 are marked with "x" by definition. While the total number of game plays in each of the four lanes is the same, we see that the open reasoning models have the highest share of aborted games, followed by open non-reasoning models. In the Distance Score subplot, we see that the aborted games span widely for open non-reasoning models, indicating some game plays, although achieving a high Distance Score, have pitifully accumulated too many penalties and thus have been marked aborted and taken out of the account of the Main Score. In contrast, commercial models have generally better Penalty Score and Distance Score. 

An interesting observation is that, in the Distance Score subplot, the aborted games tend to reside in lower ranges for non-reasoning models, while the reasoning models might play an aborted game despite achieving a high Distance Score. This is especially prominent for open non-reasoning models, and hints that the reasoning mode, rather than helping in games where these models might achieve a high Distance Score, might introduce noises that lead to more penalties.

\subsection{Detailed Analysis}

\textbf{Idea}:  
We hypothesize that Instruction Following ability and Spatial Estimation abilities serve as the foundational pillars of performance, as illustrated in Figure \ref{fig:capabilities-hierarchy}. With weaker pillars, model reasoning might not add much to the performance. 

\begin{figure}[ht]
\centering
\begin{tikzpicture}[
  box/.style={draw, thick, rounded corners=2pt, minimum width=1cm, minimum height=0.6cm, fill=lightgray!20}
]
  % Critical Thinking
  \node[box, minimum width=6cm] (ct) at (0,2) {Critical Thinking};
  
  % Coord Estimation
  \node[box, minimum width=4.5cm] (ce) at (0,1) {Spatial Estimation};
  
  % Instruction Following
  \node[box, minimum width=3cm] (if) at (0,0) {Instruction Following};
  
  % Connecting lines to show pillar relationship
  \draw[thick, ->] (if.north) -- (ce.south);
  \draw[thick, ->] (ce.north) -- (ct.south);
\end{tikzpicture}
\caption{Capability Hierarchical}
\label{fig:capabilities-hierarchy}
\end{figure}

Penalties include the Invalid Format Penalty and the Invalid Move Penalty. We plan to use the ratio of Invalid Format Penalty as an agent of Instruction Following ability, sort the models by it, compare the result of the same model with reasoning on or off, and see what percentage the reasoning capability adds to the performance. The assumption is that models with better Instruction Following abilities benefit more from the introduction of the reasoning mode. 

We plan to apply the same analysis to Invalid Move Penalty. 

\subsection{Qualitative Samples} 

\textbf{Idea}: 
We plan to do qualitative analysis from the following four angles: Spatial Estimations, Strategy, Critical Thinking, Communication \& Collaboration, and analyze and compare performance across commercial models (Y/N) and reasoning models (Y/N). For GPT-5, we compare across model size (original / mini) as well. 

Here we briefly motivate each of these angles. Concrete examples and analysis will be added later. 

\subsubsection{Spatial Estimation}
From this angle, we check if models can propose valid moving plans. It captures models' Spatial Estimation ability in our grid settings and serves as a foundational pillar of the overall performance. When models are not capable in this regard, they tend to propose invalid moving plans, and attempts to resolve the wrong movements after getting penalties often conduce to cascades of invalid moving plans, therefore quickly exhaust the number of allowed penalties and end up either aborting the game, or attain a low penalty score. 

\subsubsection{Strategy}
In addition to the crude Spatial Estimation, we check how models come up with moving plans. We have observed some smart strategies and their not-so-smart counterparts. In some game plays, one model uses the initial coordinates of the objects to instruct the other model as target locations, thus avoiding penalties of moving objects to cells occupied by the background grids to a large extent; while in some other game plays, models appear to exchange initial coordinates just for the sake of exchanging, without making use of this information. Another example is in some game plays, models choose the most sparse row as the target, and lay objects one by one on it; while in some other game plays, the models unnecessarily chose geometric patterns to lay the objects, such as on the diagonal, disregarding the occupied cells, and run into complexities that they could have dodged.

\subsubsection{Critical Thinking}
Some models show the ability to critically reason about the moving plan that either itself or the other player proposed. They either refute the plan immediately when it's proposed, or take a sharp brake when they are just about to execute the invalid moving plan. This is, as expected, more frequent in reasoning models compared to non-reasoning models. 
However, on some occasions, we see pseudo-reasoning: a model rejects a correct plan, utters reasoning-like sentences, and proposes another plan that sometimes is correct, sometimes is wrong.

\subsubsection{Communication \& Collaboration}
The above three dimensions positively correlate with the model performance: The better the models are at them, the higher the scores. This dimension, however, is different. Specifically, we observe how models communicate with each other to recover from invalid moves due to target locations being occupied. Some models go at lengths to execute their alternative plans without letting their partner know, they might still have a good score, simply because both player 1 and player 2 adopt the exact same strategy, for example, move the object to one spot next to the original target location. This leads to inflated scores that can be debunked when two models of different default recovery strategies, or one model and one human, play against each other.

\section{Air Balloon Survival - Game Details}\label{sec:appendix_airballoon}

\subsection{Prompt Templates}\label{sec:hot_air_balloon_appendix_prompt_templates}

The prompt template for both players of the Hot Air Balloon Survival game is given in Figure~\ref{fig:hot_air_balloon_init_prompt}. 
Parse error prompts handed to players when their responses did not follow the format instructions are given in Figure~\ref{fig:hot_air_balloon_parse_error_prompts}, and game error prompts for invalid actions are given in Figure~\ref{fig:hot_air_balloon_invalid_action_prompts}.

\begin{figure*}
  \centering
  \begin{prompt}
You are participating in a collaborative negotiation game.\\

Together with another participant, you must agree on a single set of items that will be kept. Each of you has your own view of how much each item matters to you (importance). You do not know how the other participant values the items. Additionally, you are given the effort each item demands.\\
You may only agree on a set if the total effort of the selected items does not exceed a shared limit:\\

LIMIT = \$LIMIT\$\\

Here are the individual item effort values:\\

Item effort = \$ITEM\_WEIGHTS\$\\

Here is your personal view on the importance of each item:\\

Item importance values = \$UTILITY\_SCALE\_PLAYER\$\\

Goal:\\

Your goal is to negotiate a shared set of items that benefits you as much as possible (i.e., maximizes total importance to YOU), while staying within the LIMIT. You are not required to make a PROPOSAL in every message - you can simply negotiate as well. All tactics are allowed!\\

Interaction Protocol:\\

You may only use the following structured formats in a message:\\

PROPOSAL: \{'A', 'B', 'C', …\}\\
Propose keeping exactly those items.\\

REFUSE: \{'A', 'B', 'C', …\}\\
Explicitly reject opponent's proposal.\\

ARGUMENT: \{'...'\}\\
Defend your last proposal or argue against the player's proposal.\\

AGREE: \{'A', 'B', 'C', …\}\\
Accept the opponent's proposal which ends the game.\\

\$STRATEGIC\_REASONING\_FORMAT\$\\

Rules:\\

You may only AGREE on a proposal the other party has logged via PROPOSAL.\\
You may only REFUSE a proposal the other party has logged via PROPOSAL.\\
Total effort of any PROPOSAL or AGREE set must be $\leq$ LIMIT.\\
Do NOT reveal your hidden importance scores.\\
A tag in a structured format must be followed by colon and whitespace. The argument must be a python set containing 0 or more strings.\\
So, it must be of the form TAG: \{...\}\\
Strictly follow the interaction protocol and DO NOT write anything beyond the given structure.\\
The game ends when one side gives an AGREE to a PROPOSAL made by the other player.\\
The content in your response which can be handed to the other player has to be non-empty.\\
Only proposals which have been logged via the PROPOSAL format structure and which haven't been refused via REFUSE are active.\\

\$REQUIRE\_ARGUMENT\$\\
\$STRATEGIC\_REASONING\_RULE\$
  \end{prompt}

  \caption{Initial prompt template of Air Balloon Survival in English.}
  \label{fig:hot_air_balloon_init_prompt}
\end{figure*}

\smallskip
\noindent Substrings Figure \ref{fig:hot_air_balloon_init_prompt} marked by \texttt{\$...\$} are placeholders which get replaced depending on the game instance.
Note that the last three placeholders may also be replaced by the empty string depending on the game settings. Here we list the replacements for placeholders in the initial prompt:
\begin{itemize}
  \item \texttt{\$LIMIT\$} – maximum weight for a deal.
  \item \texttt{\$ITEM\_WEIGHTS\$} – mapping from items to weights.
  \item \texttt{\$UTILITY\_SCALE\_PLAYER\$} – mapping from items to preference values for a given player.
  \item \texttt{\$STRATEGIC\_REASONING\_FORMAT\$} – 
    \begin{quote}
    \texttt{STRATEGIC REASONING: \{'...'\}}\\
    Describe your strategic reasoning or anticipation explaining your choice of action. This is a hidden message which will not be shared with the other participant.
    \end{quote}
  \item \texttt{\$REQUIRE\_ARGUMENT\$} – You must include the \texttt{ARGUMENT} format at least once somewhere in all of your messages.
  \item \texttt{\$STRATEGIC\_REASONING\_RULE\$} – 
    \begin{quote}
    You must include the \texttt{STRATEGIC REASONING} format only once at the very beginning of every one of your messages and not more often. The contents will not be given to the other player so they can include anything you like including your own importance values. Here you should reason step by step to come up with your next move.
    \end{quote}
\end{itemize}

\begin{figure*}
  \centering
  \begin{subfigure}[b]{0.48\textwidth}
    \centering
    \begin{prompt}
Your response did not start with the proper strategic reasoning tag at the very beginning of your response.\\
The very first structured format must be of the form STRATEGIC REASONING: \{...\}. Try again.
    \end{prompt}
    \caption{Parse error prompt in case strategic reasoning tag at the beginning of a response is missing. Only applies when strategic reasoning tag is required.}
    \label{fig:hot_air_balloon_no_sr_prompt}
  \end{subfigure}
  \hfill
  \begin{subfigure}[b]{0.48\textwidth}
    \centering
    \begin{prompt}
Your response did not contain an argument.\\
You must include the structured format ARGUMENT: \{...\} somewhere in your response. Try again.
    \end{prompt}
    \caption{Parse error prompt in case an argument is missing. Only applies when argument tag is required}
    \label{fig:hot_air_balloon_no_argument_prompt}
  \end{subfigure}

  \vskip\baselineskip

  \begin{subfigure}[b]{0.48\textwidth}
    \centering
    \begin{prompt}
Your response contained an untagged sequence or you used STRATEGIC REASONING more than once.\\
You may only use the structured formats as explained in the initial message.\\
They must all be of the form TAG: \{...\}.
    \end{prompt}
    \caption{Parse error prompt in case the response contains an untagged sequence.}
    \label{fig:hot_air_balloon_untagged_sequence_prompt}
  \end{subfigure}
  \hfill
  \begin{subfigure}[b]{0.48\textwidth}
    \centering
    \begin{prompt}
Your response only contained a strategic reasoning tag.\\
You must at least include one more valid tag in your response, so that the other player receives a message. Try again.
    \end{prompt}
    \caption{Parse error prompt in case response only contained a strategic reasoning tag. Only applies when argument tag is not required.}
    \label{fig:hot_air_balloon_only_sr_prompt}
  \end{subfigure}

  \vskip\baselineskip

  \begin{subfigure}[b]{0.48\textwidth}
    \centering
    \begin{prompt}
You used a PROPOSAL tag, but did not provide a valid python set containing strings as arguments, e.g. \{'A', 'B', 'C', ...\}. Try again.
    \end{prompt}
    \caption{Parse error prompt in case of an invalid python set as argument.}
    \label{fig:hot_air_balloon_invalid_python_set_error}
  \end{subfigure}

  \caption{Parse error prompts handed to player when a response did not follow the instructions on structured formats.}
  \label{fig:hot_air_balloon_parse_error_prompts}
\end{figure*}

\begin{figure*}
  \centering
  \begin{subfigure}[b]{0.48\textwidth}
    \centering
    \begin{prompt}
You refused a proposal which is not active.\\
Proposals are only active if they have been logged by the other player via PROPOSAL and have not been deactivated by you via REFUSE. Try again.
    \end{prompt}
    \caption{Game error prompt in case a non-active was refused.}
    \label{fig:refuse_error_prompt}
  \end{subfigure}
  \hfill
  \begin{subfigure}[b]{0.48\textwidth}
    \centering
    \begin{prompt}
You made more than one agreement.\\
Final deals cannot be ambiguous. Try again.
    \end{prompt}
    \caption{Game error prompt in case agreement to a deal was ambiguous.}
    \label{fig:agreement_ambiguous_prompt}
  \end{subfigure}

  \vskip\baselineskip

  \begin{subfigure}[b]{0.48\textwidth}
    \centering
    \begin{prompt}
You agreed to a proposal which is not active.\\
Proposals are only active if they have been logged by the other player via PROPOSAL and have not been deactivated by you via REFUSE. Try again.
    \end{prompt}
    \caption{Game error prompt in case .}
    \label{fig:agreement_non_active_prompt}
  \end{subfigure}
  \hfill
  \begin{subfigure}[b]{0.48\textwidth}
    \centering
    \begin{prompt}
Your proposal includes items which are not in the game. Try again.
    \end{prompt}
    \caption{Game error prompt in case a non-active proposal was agreed to.}
    \label{fig:proposal_error_prompt}
  \end{subfigure}

  \caption{Game error prompts handed to player when a response did not follow the instructions on the rules of the game.}
  \label{fig:hot_air_balloon_invalid_action_prompts}
\end{figure*}

\subsection{Evaluation Metrics}\label{subsec:appendix_balloon_metrics}

Each player $p \in P$ receives an individual score based on the utility of the final deal $D$, i.e. the sum of the preference values of items in $D$. This sum gets normalized by the value of the optimal solution to their instance of the 0/1 Knapsack Problem. The overall game score is defined as the harmonic mean of the two players’ normalized scores, further normalized by the maximum harmonic mean achievable for an instance. If the total weight of the deal exceeds the capacity $W$, the score is set to $0$. We chose the harmonic mean as our main metric as it rewards collectively balanced outcomes by favouring deals where both players achieve high scores and penalizing those where one player benefits disproportionately. 
The individual scores of players is each player’s outcome relative to the optimal solution of their own knapsack problem. For a final deal $D \subseteq \{1,\dots,n\}$, the normalized score of player $p \in P$ is  
$$
f_p(D) = 100 \cdot \frac{\sum_{i \in D} v_p(i)}{\text{OPT}_p},
$$  
where $\text{OPT}_p$ is the value of the optimal solution to player $p$'s 0/1 Knapsack Problem.  
The harmonic mean of the players' normalized scores is defined as  
$$
f_{\text{harm}}(D) =
\begin{cases}
\frac{2 f_{1}(D) f_{2}(D)}{f_{1}(D) + f_{2}(D)} & \text{if } \sum_{i \in D} w_i \leq W, \\
0 & \text{otherwise}.
\end{cases}
$$  
The final game score is the normalized harmonic mean, 
$$
\text{Quality}(D) = 100 \cdot \frac{f_{\text{harm}}(D)}{\text{OPT}^*},
$$  
where $\text{OPT}^*$ denotes the maximum harmonic mean achievable in the given instance.

\subsection{Overall Results}
\begin{figure}[ht]
    \centering
    \begin{subfigure}{0.5\textwidth}
        \centering
        \includegraphics[width=\linewidth]{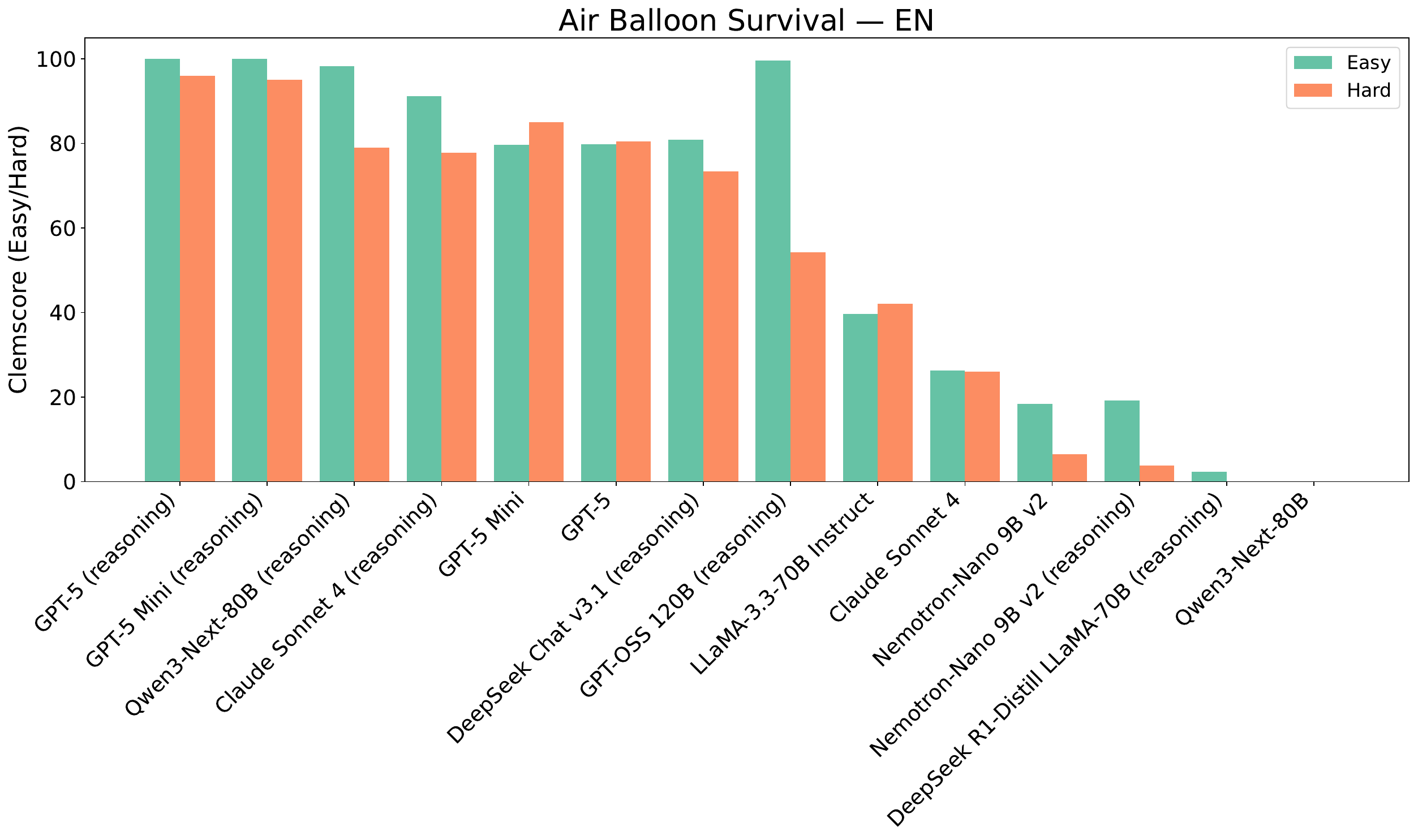}
        \caption{Clemscores on English instances.}
        \label{fig:air-balloon-overall-en}
    \end{subfigure}\hfill
    \begin{subfigure}{0.5\textwidth}
        \centering
        \includegraphics[width=\linewidth]{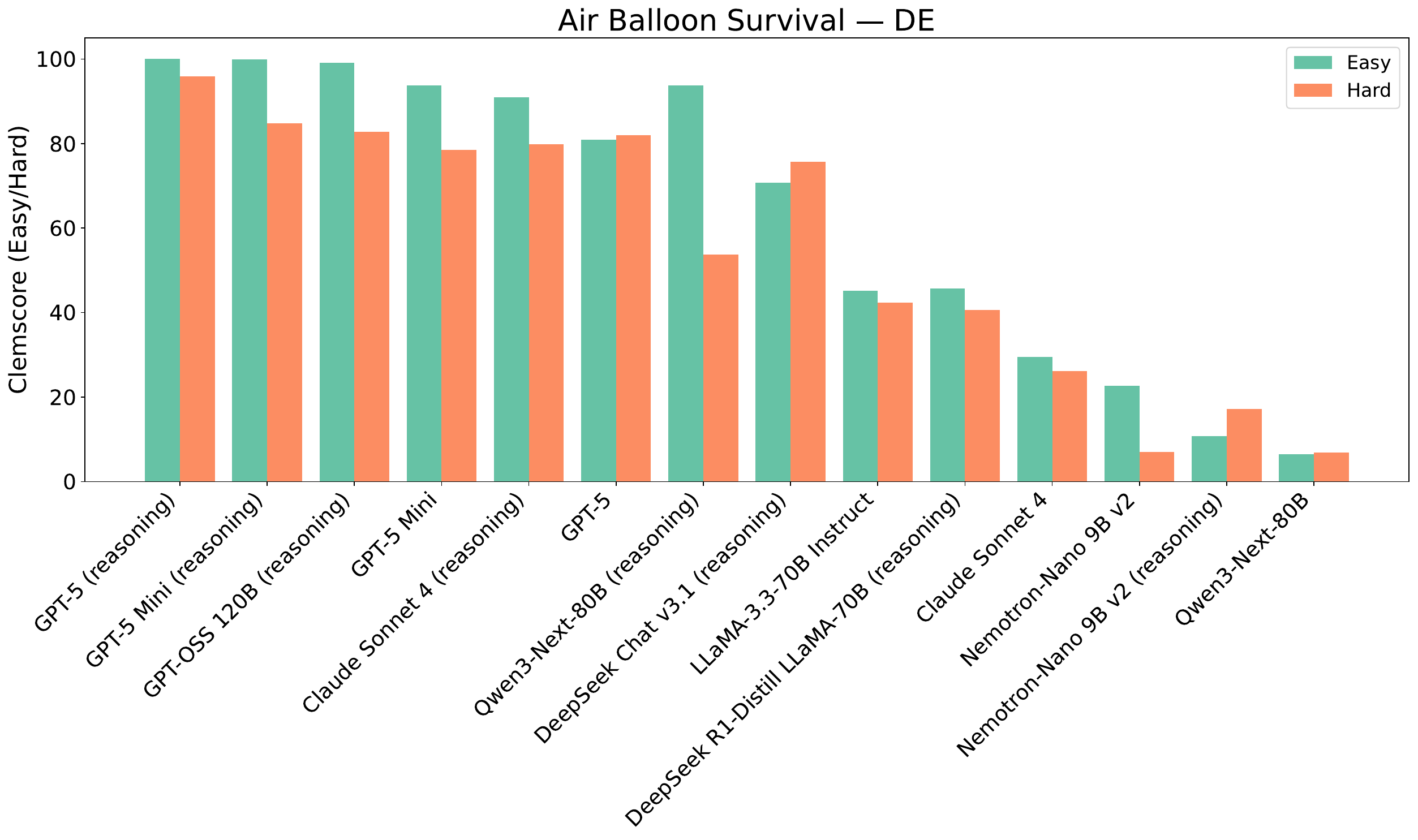}
        \caption{Clemscores on German instances.}
        \label{fig:air-balloon-overall-de}
    \end{subfigure}

    \begin{subfigure}{0.5\textwidth}
        \centering
        \includegraphics[width=\linewidth]{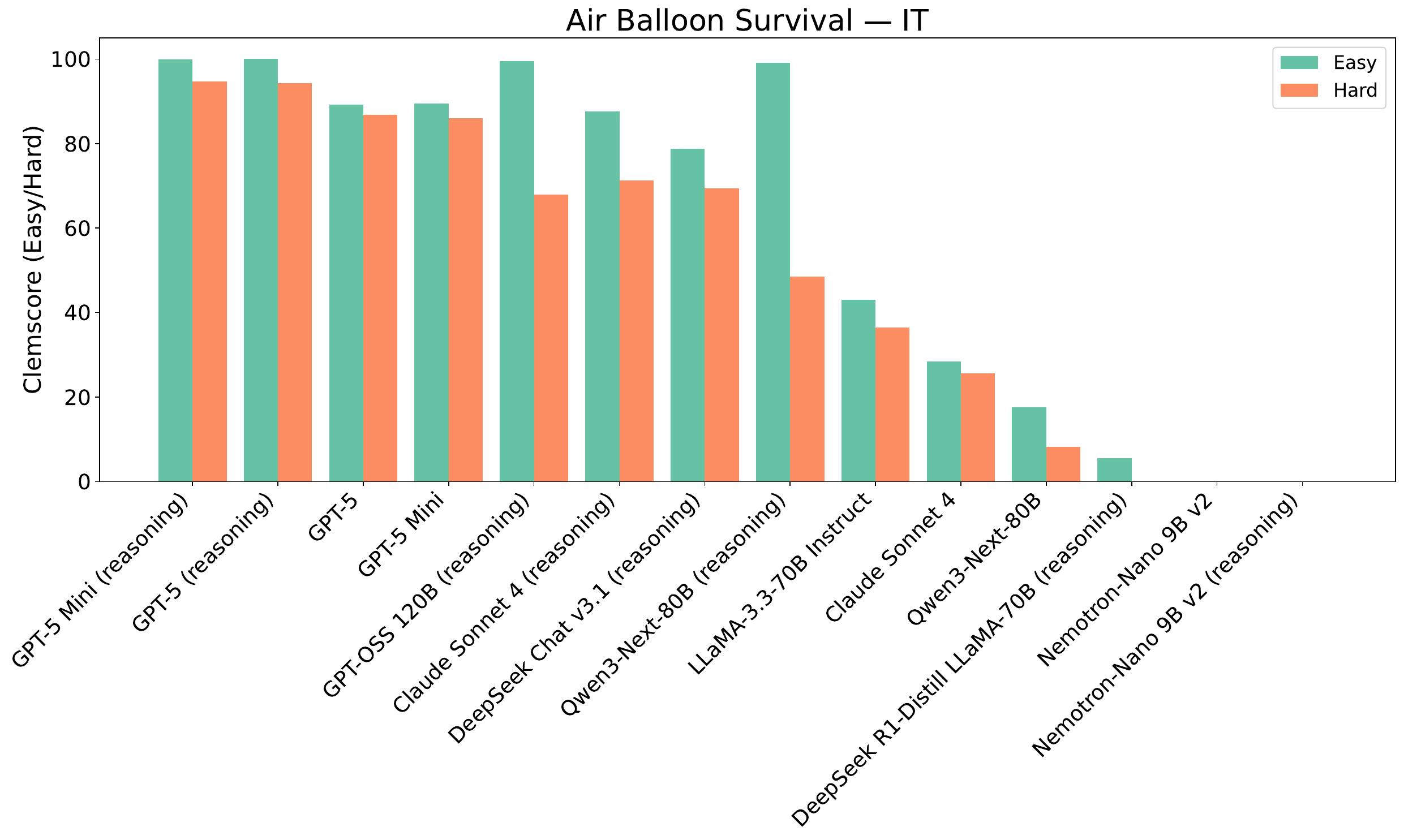}
        \caption{Clemscores on Italian instances.}
        \label{fig:air-balloon-overall-it}
    \end{subfigure}

    \caption{Overall model performance for easy and hard sections of experiments for all languages.}
    \label{fig:air-balloon-overall}
\end{figure}

\begin{figure}[ht]
    \centering
    \includegraphics[width=\linewidth]{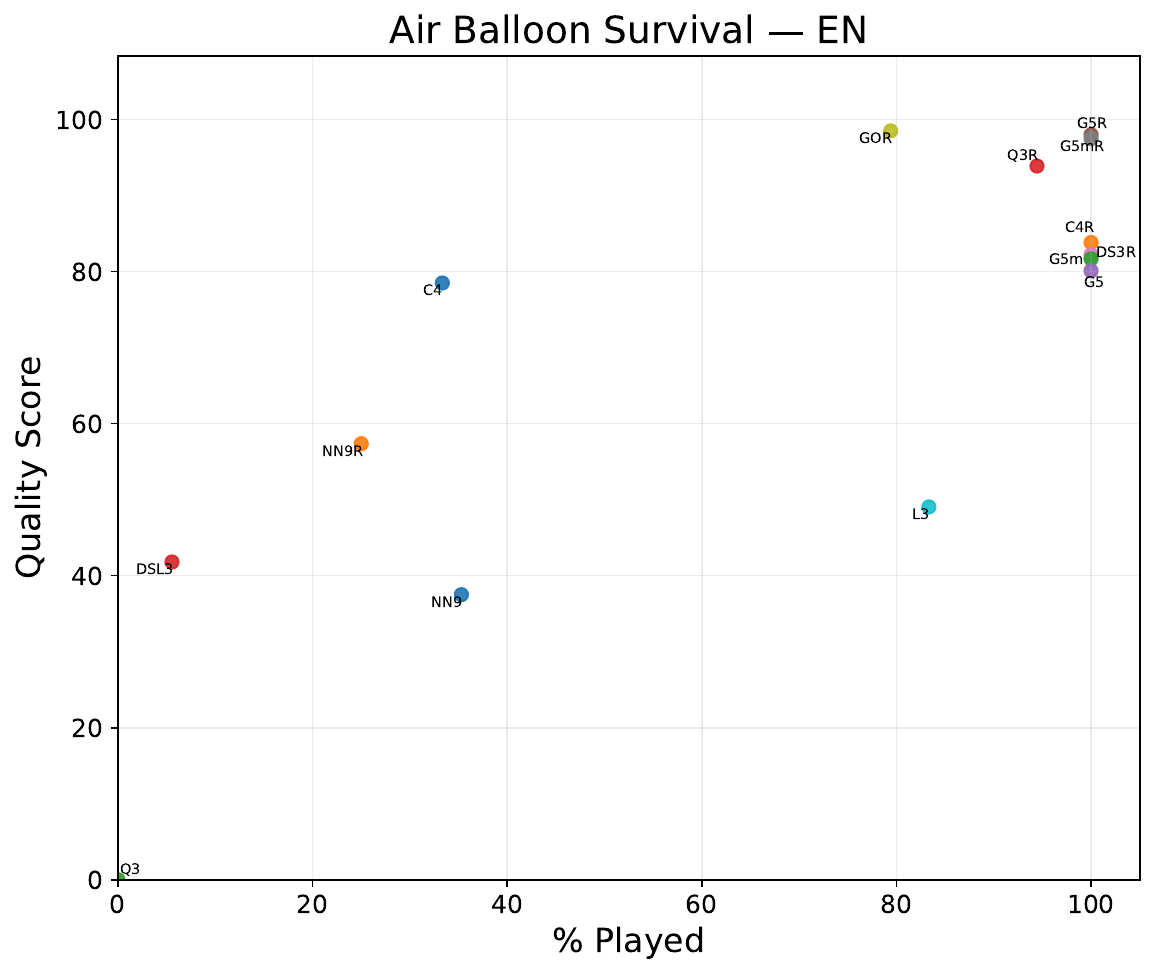}
    \caption{Quality and Played scores on \textit{Air Balloon Survival} (English). 
    Scatters are annotated with acronyms of model names. `R' denotes reasoning model.}
    \label{fig:air-balloon-quality-played-en}
\end{figure}

\begin{figure}[ht]
    \centering
    \includegraphics[width=\linewidth]{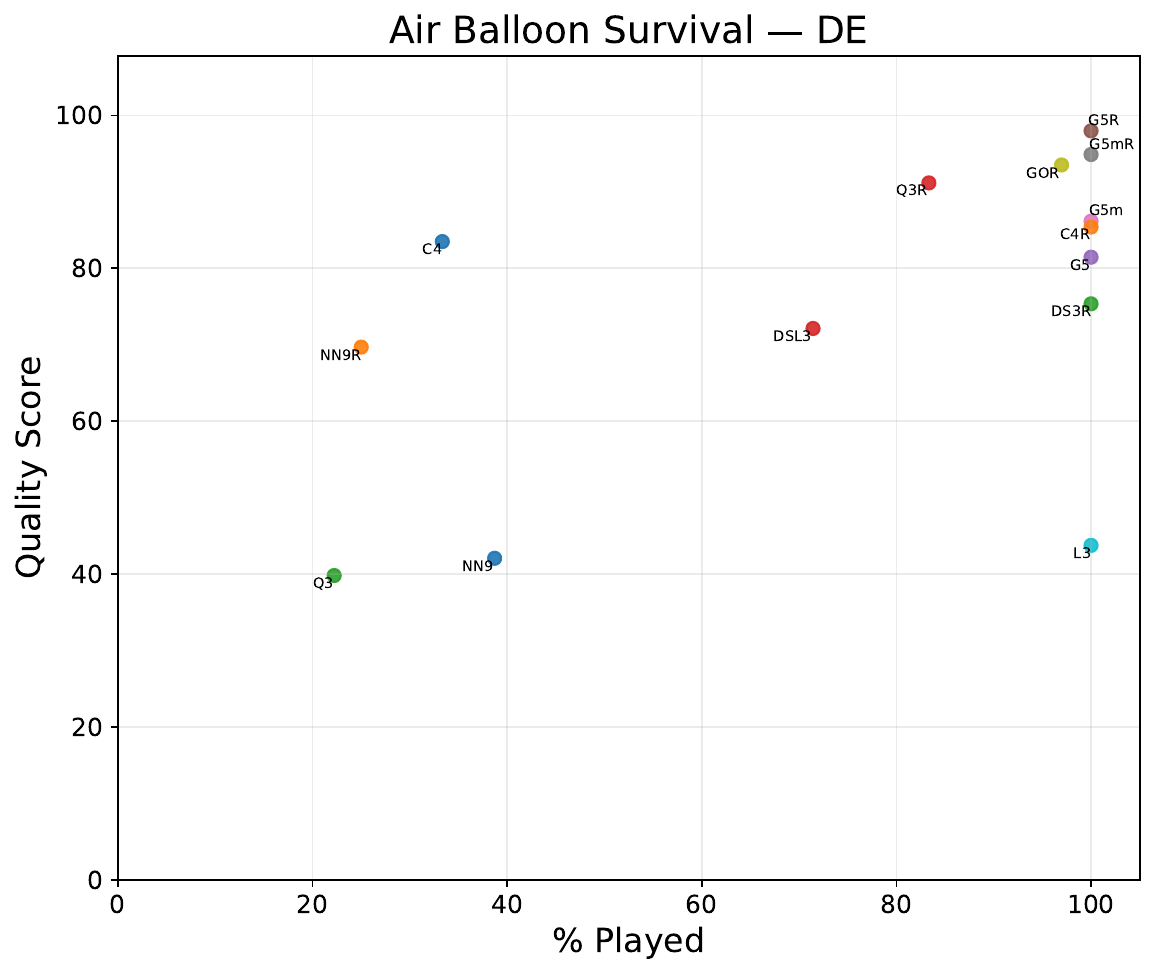}
    \caption{Quality and Played scores on \textit{Air Balloon Survival} (German). 
    Scatters are annotated with acronyms of model names. `R' denotes reasoning model.}
    \label{fig:air-balloon-quality-played-de}
\end{figure}

\begin{figure}[ht]
    \centering
    \includegraphics[width=\linewidth]{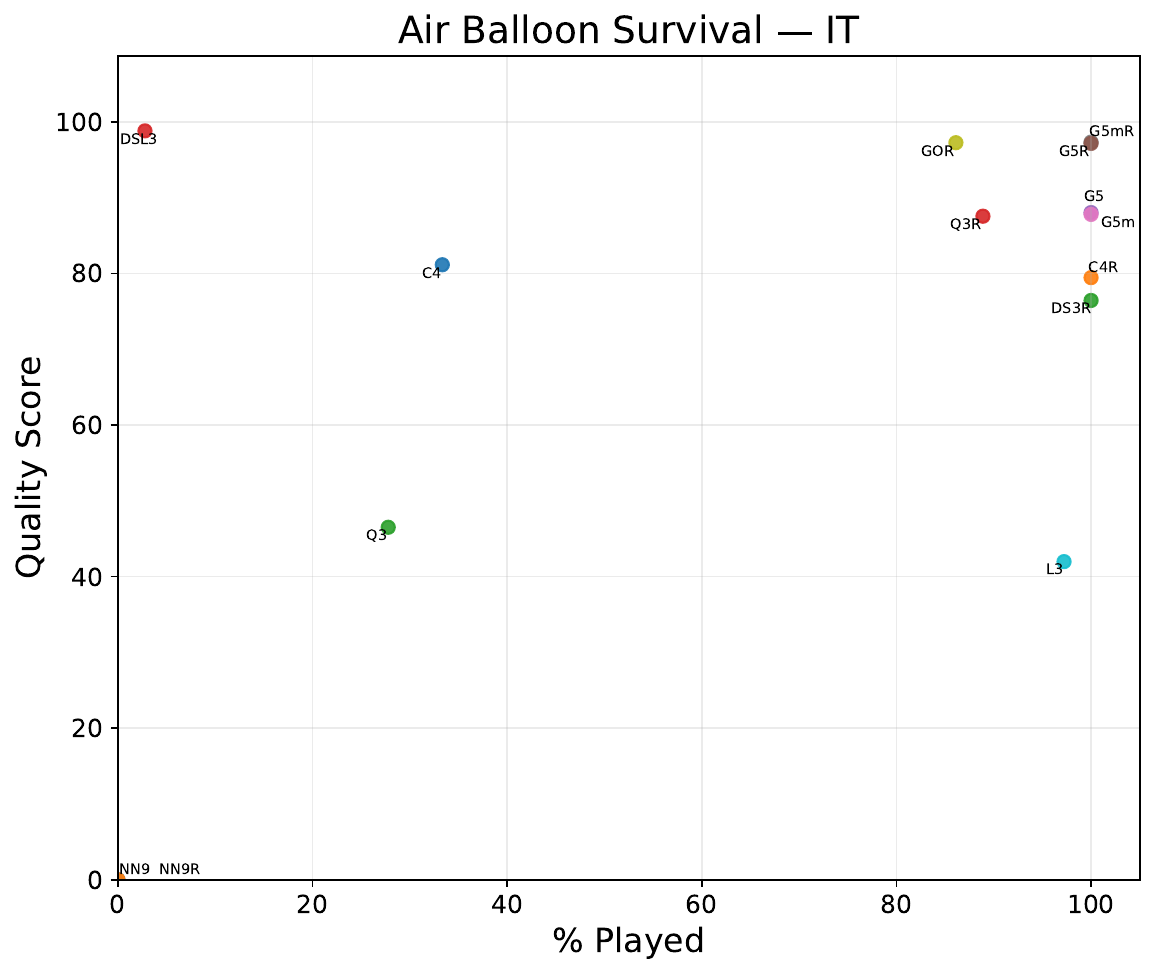}
    \caption{Quality and Played scores on \textit{Air Balloon Survival} (Italian). 
    Scatters are annotated with acronyms of model names. `R' denotes reasoning model.}
    \label{fig:air-balloon-quality-played-it}
\end{figure}

As outlined before, we generated three experimental settings, each comprising an easy and a hard subset. Figure \ref{fig:air-balloon-overall} reports the clemscore for every model on both subsets. Overall, reasoning-enabled models outperform their non-reasoning counterparts on easy and hard instances alike. Notably, GPT-OSS and Qwen-3 (reasoning) exhibit a large gap between easy and hard subsets for several languages in the batch. The decline is driven primarily by those instances for which players are handed opposing goals. While GPT-OSS and Qwen-3 (reasoning) are able to solve the individual knapsack subproblems, their reasoning transfers less effectively to negotiating common ground when compared with the stronger reasoning models. The best-performing models on our game are GPT-5 and GPT-mini with reasoning enabled with the former achieving a clemscore of over $97$ on all languages for Air Balloon Survival. The reasoning component boosts performance by around $14$ across languages. The best open weight model on Air Balloon Survival is GPT-OSS, another reasoning model.

There are two notable exceptions to the general advantage of reasoning: Nemotron-Nano-9B (reasoning) vs. Nemotron-Nano-9B (non-reasoning), and DeepSeek R1-Distill LLaMA-70B vs. LLaMA-3.3-70B Instruct. In both cases, the lower clemscore for the respective reasoning model is linked to weaker instruction following, rather than degraded reasoning per se, since their quality scores mostly improve when reasoning is enabled (cf. Figures \ref{fig:air-balloon-quality-played-en}, \ref{fig:air-balloon-quality-played-de}, \ref{fig:air-balloon-quality-played-it}). Finally, we observe a consistent pattern of commercial models outperforming open-weight models, for both reasoning and non-reasoning variants.

%----- DETAILED ANALYSIS SECTION ----
\subsection{Detailed Analysis}
\label{sec:air-balloon-negotiation-dynamics}

We observe several phenomena related to negotiation dynamics, which can be amplified or diminished in reasoning models. These include bargaining strategies such as the stubborn repetition of proposals, implicit role assignment between players (one active, one reactive), and varying degrees of collaboration. We also examine how well models represent their own and their opponent’s goals. Below, we summarize our main findings before elaborating and quantifying specific aspects.

\begin{figure}[ht]
    \centering
        \includegraphics[width=0.5\textwidth]{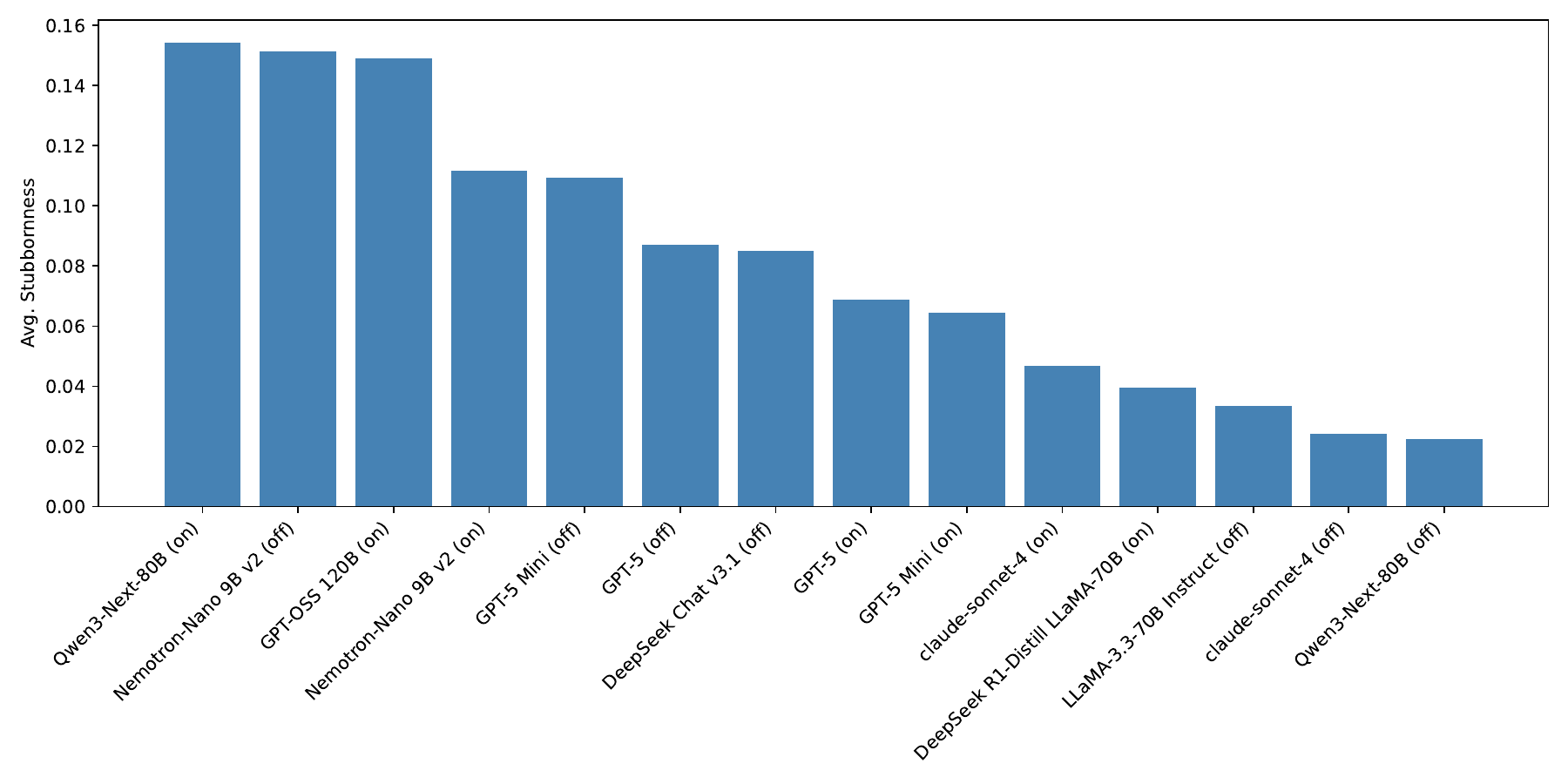}
        \caption{Average per-instance stubbornness per model across languages and experiments.}
        \label{fig:air-balloon-stubbornness-all}
\end{figure}

\begin{figure}[!ht]
    \centering
    \includegraphics[width=\linewidth]{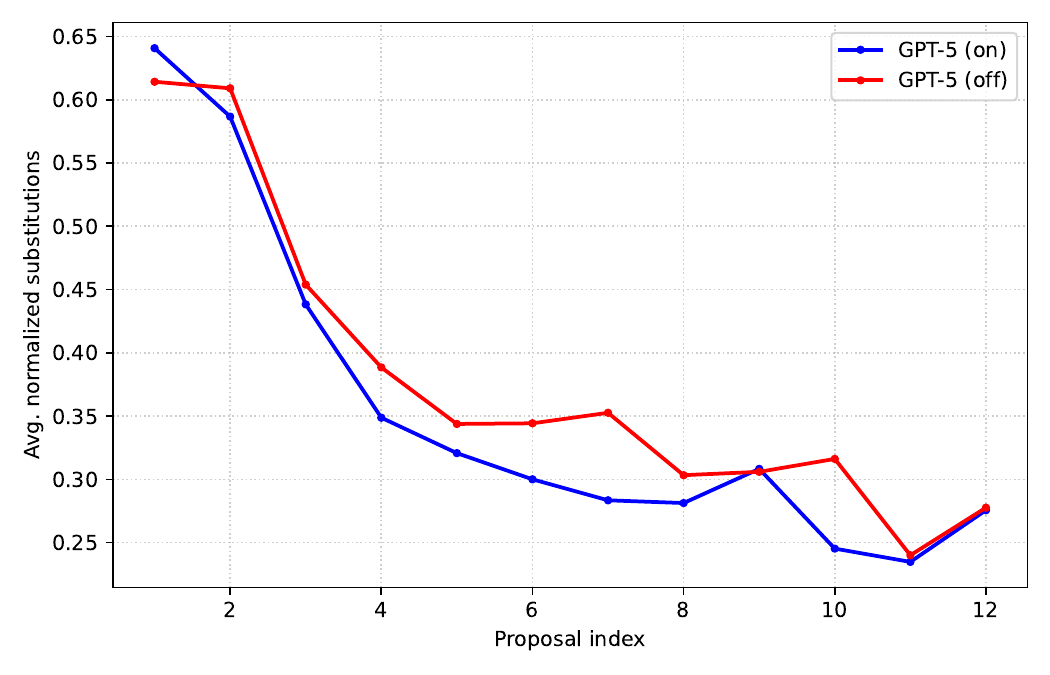}
    \caption{Substitutions over proposals in temporal order for GPT-5. 
    Values are averaged over proposals at each time step in a negotiation across all experiments and languages.}
    \label{fig:air-balloon-gpt-5-subs}
\end{figure}

\begin{figure}[ht]
    \centering
    \includegraphics[width=\linewidth]{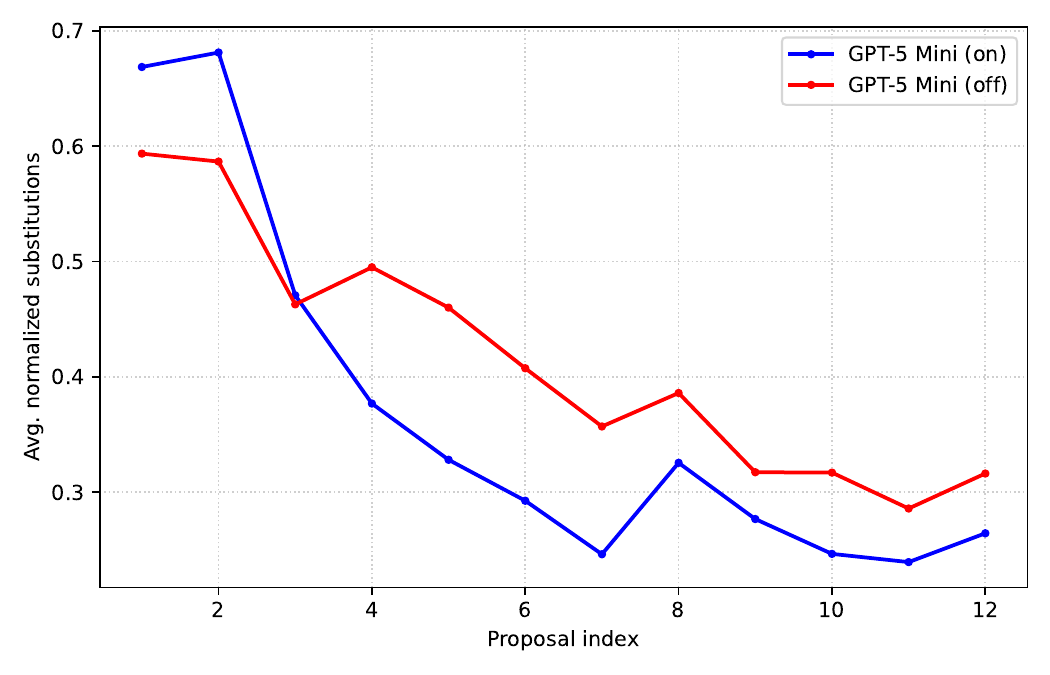}
    \caption{Average normalised substitutions over proposals in temporal order for GPT-5 Mini.}
    \label{fig:air-balloon-gpt5-mini-subs}
\end{figure}

\begin{figure}[ht]
    \centering
    \includegraphics[width=0.48\textwidth]{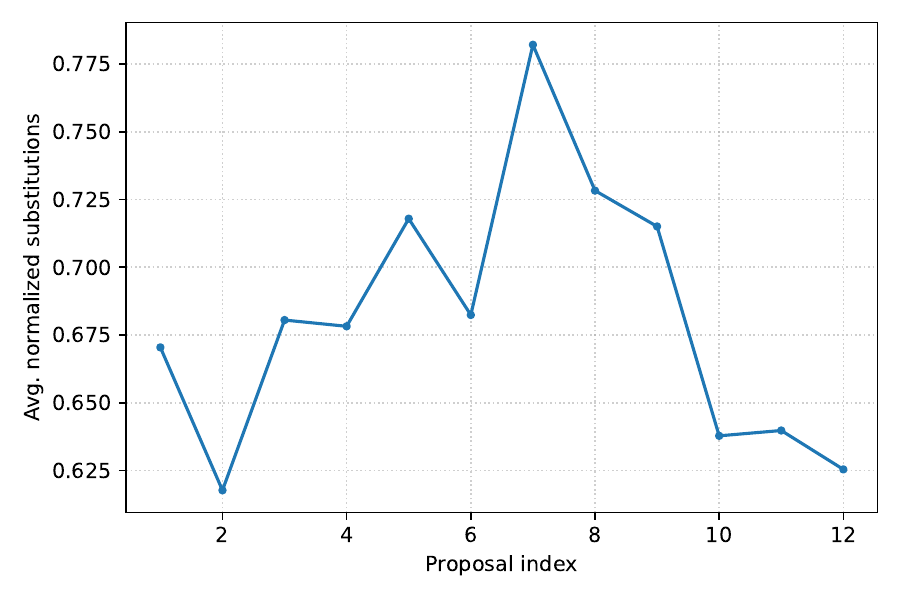}
    \caption{Average normalised substitutions over proposals in temporal order for GPT-OSS 120B.}
    \label{fig:air-balloon-gpt-oss-subs}
\end{figure}

\begin{figure}[ht]
    \centering
    \includegraphics[width=0.48\textwidth]{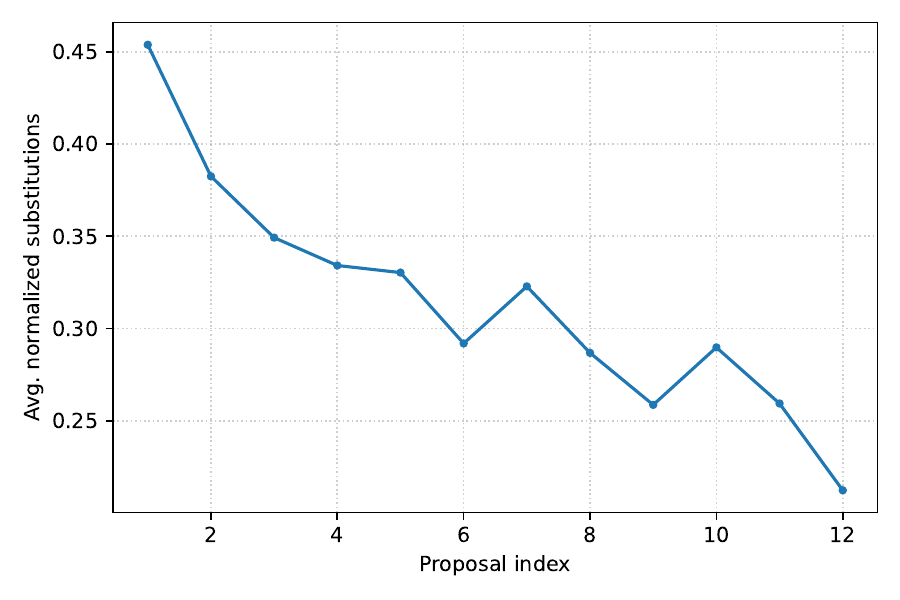}
    \caption{Average normalised substitutions over proposals in temporal order for DeepSeek Chat v3.1.}
    \label{fig:air-balloon-dschatv3.1-subs}
\end{figure}

\paragraph{Findings}
\begin{enumerate}
    \item \textbf{Dominance}: Two dominance patterns appear in our game.  
    1) \emph{Stubbornness}: One player may dominate by repeatedly insisting on the same proposal, measured as the relative frequency of repeated proposals (Figure~\ref{fig:air-balloon-stubbornness-all}). Stubbornness increases with opposing goals and is generally higher in player 1. Reasoning mitigates this in stronger models (GPT-5-mini, GPT-5, Nemotron), while weaker ones lacking counterpart modelling and common-ground finding (GPT-OSS, Qwen3) often show high stubbornness: they can find individual knapsack solutions but fail to resolve tension.  
    2) \emph{Alternation}: Some weaker models exhibit role asymmetry, with one player acting as proposer (proactive) and the other as refuser or critic (passive). This behaviour is reduced in reasoning models and correlates with better performance. Table~\ref{tab:air-balloon-alternation} shows the average proposal alternation rate per model, quantifying this asymmetry.
    \item \textbf{Collaboration/Common Ground}: Some models employ collaborative strategies, iteratively refining deals so that players agree on more items over time. This behaviour is strongest in the best-performing reasoning models (GPT-5, GPT-5-mini) and reflected in Figures~\ref{fig:air-balloon-gpt-5-subs}--\ref{fig:air-balloon-dschatv3.1-subs}, which show average substitutions per proposal. These models display high initial disagreement followed by rapid convergence, indicating efficient bargaining. However, this also introduces bias: player 1 typically secures better outcomes by setting the initial stage, with player 2 largely following.
    \item \textbf{Anchor positioning}: The best models start with high disagreement and resolve conflicts quickly (cf. Figures~\ref{fig:air-balloon-gpt-5-subs}--\ref{fig:air-balloon-dschatv3.1-subs}). This effect is amplified when reasoning traces are enabled in GPT-5 and GPT-5-mini.
    \item \textbf{Knapsack Reasoning}: Reasoning models outperform others by following a three-stage plan: (1) applying a greedy heuristic (value-to-weight ratio), (2) filling the knapsack to the weight limit, and (3) refining via substitutions. Step (3) tends to be long and exhaustive in Qwen 3. Llama Distill and Nemotron often enter correction loops. GPT-OSS and Qwen 3 use reasoning traces for instruction following, unlike Nemotron and R1 Distill, which achieve lower clemscores (due to low \% played) despite good quality scores.
    \item \textbf{Loops}: Nemotron and Llama frequently fall into correction loops, failing to search effectively in step (3).
    \item \textbf{Counterpart Modelling}: Models attempt to infer counterpart preferences, usually by noting repeated items in proposals, but rarely go beyond this. Llama occasionally confuses itself with the other player when reprompted. Figure~\ref{tab:air-balloon-pareto-adherence} shows per-instance Pareto-adherence (the relative frequency of proposals lying on the Pareto front). Since Pareto efficiency requires aligning both players’ payoffs, higher adherence may indicate stronger self- and counterpart-modelling.
\end{enumerate}

\paragraph{Observed Strategies}
A recurring strategy is the stubborn repetition of proposals. In reasoning-enabled models, traces reveal that this often stems from uncertainty about the opponent’s goals. Repetition is typically justified as maximizing one’s own utility, and is strongest in models that reason effectively about their own knapsack but not about alignment. 

We quantify stubbornness as the average number of repeated proposals per instance (Figure~\ref{fig:air-balloon-stubbornness-all}). Among reasoning models, we observe two clusters:
\begin{itemize}
    \item \textbf{High-Stubbornness Models:} Qwen 3, GPT-OSS – reasoning traces used mainly for individual optimization.
    \item \textbf{Mitigated-Stubbornness Models:} GPT-5, GPT-5-mini – results indicate that reasoning is used for strategic adaptation as stubbornness is reduced.
\end{itemize}

\begin{figure}[ht]
    \centering
        \includegraphics[width=0.5\textwidth]{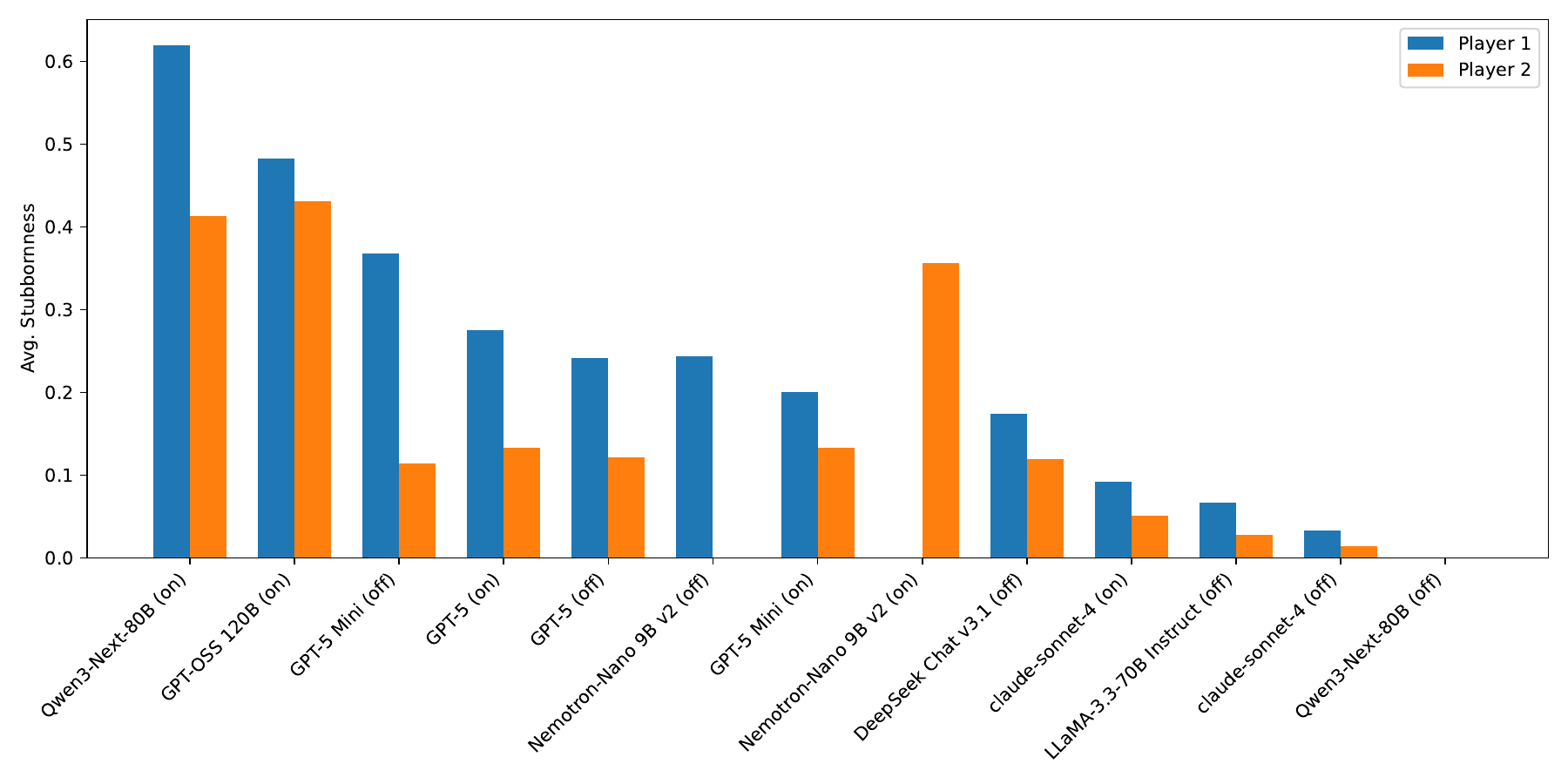}
        \caption{Average per-player per-instance stubbornness for each model across languages on opposing-goals instances.}
        \label{fig:air-balloon-player-stubbornness-focus}
\end{figure}

This divergence is especially pronounced in opposing-goal instances (Figure~\ref{fig:air-balloon-player-stubbornness-focus}). Player 1’s advantage arises because they have the ability to advance the game state first, thus shaping player 2’s context. However, the game setup does not inherently favour player 1 — the bias emerges from behavioural asymmetry.

Table~\ref{tab:air-balloon-player-scores-opp-goals} reports individual player scores for opposing-goal cases. Notably, GPT-5-mini and GPT-5 gain more on player 2 when reasoning is enabled, though overall outcomes remain biased toward player 1 due to stubbornness.

%STUBBORNNESS PLOTS

%TABLE WITH INDIVIDUAL PLAYER SCORE ON PLAYED OPP GOAL INSTANCES
% \begin{table}[!ht]
% \centering
% \footnotesize
% \setlength{\tabcolsep}{4pt}
% \begin{minipage}{0.48\textwidth} % match subfigure width
%     \centering
%     \begin{tabular}{l cc}
%     \hline
%     \textbf{Model} & \textbf{P1} & \textbf{P2} \\
%     \hline
%     GPT-OSS & 91.835 & 77.598 \\
%     Nemotron-Nano (reasoning) & 92.475 & 64.105 \\
%     GPT-5 (reasoning) & 89.521 & 76.050 \\
%     GPT-5 Mini (reasoning) & 88.267 & \textbf{77.116} \\
%     Qwen3 (reasoning) & \textbf{92.757} & 63.293 \\
%     GPT-5 (no reasoning) & 80.148 & 65.876 \\
%     GPT-5 Mini (no reasoning) & 86.362 & 61.951 \\
%     DeepSeek Chat v3.1 & 82.463 & 58.722 \\
%     Claude Sonnet 4 (no reasoning) & 76.094 & 65.630 \\
%     Claude Sonnet 4 (reasoning) & 74.403 & 72.400 \\
%     Nemotron-Nano (no reasoning) & 56.113 & 61.660 \\
%     LLaMA-3.3 & 49.455 & 39.049 \\
%     Qwen3 (no reasoning) & 46.353 & 18.763 \\
%     R1 Distill LM3 & - & - \\
%     \hline
%     \end{tabular}
%     \caption{Per player scores on opposing goals instances averaged across languages (not success weighted).}
%     \label{tab:air-balloon-player-scores-opp-goals}
% \end{minipage}
% \end{table}

\begin{table}[ht]
\centering
\footnotesize
\setlength{\tabcolsep}{4pt}
\begin{tabular}{l cc}
\hline
\textbf{Model} & \textbf{P1} & \textbf{P2} \\
\hline
GPT-OSS (On) & 91.835 & 77.598 \\
Nem-9B (On) & 92.475 & 64.105 \\
GPT-5 (On) & 89.521 & 76.050 \\
GPT-5 Mini (On) & 88.267 & \textbf{77.116} \\
Qwen3-80B (On) & \textbf{92.757} & 63.293 \\
GPT-5 (Off) & 80.148 & 65.876 \\
GPT-5 Mini (Off) & 86.362 & 61.951 \\
DS-v3.1 (Off) & 82.463 & 58.722 \\
Claude 4 (Off) & 76.094 & 65.630 \\
Claude 4 (On) & 74.403 & 72.400 \\
Nem-9B (Off) & 56.113 & 61.660 \\
LLaMA-70B (Off) & 49.455 & 39.049 \\
Qwen3-80B (Off) & 46.353 & 18.763 \\
LM-70B (On) & - & - \\
\hline
\end{tabular}
\caption{Per player scores on opposing goals instances averaged across languages (not weighted by \% played).}
\label{tab:air-balloon-player-scores-opp-goals}
\end{table}

As another strategy, we encounter implicit role assignment, where one player is more active in making proposals while the other is more reactive, simply refusing or demanding changes to the proposal of the other player. This strategy turns out to correlate with worse overall performance and is diminished in reasoning models, which generally prefer to take consecutive turns in making proposals.

\begin{table}[ht]
\centering
\footnotesize
\setlength{\tabcolsep}{4pt}
\begin{tabular}{l c}
\hline
\textbf{Model} & \textbf{Altern. Rate} \\
\hline
GPT-5 (On) & \textbf{1.000} \\
GPT-5 Mini (On) & \textbf{1.000} \\
GPT-5 (Off) & 0.997 \\
GPT-5 Mini (Off) & 0.989 \\
Claude 4 (On) & 0.987 \\
Claude 4 (Off) & 0.974 \\
GPT-OSS (On) & 0.961 \\
Qwen3-80B (On) & 0.883 \\
DS-v3.1 (Off) & 0.820 \\
LM-70B (On) & 0.795 \\
Nem-9B (Off) & 0.678 \\
LLaMA-70B (Off) & 0.669 \\
Nem-9B (On) & 0.437 \\
Qwen3-80B (Off) & 0.190 \\
\hline
\end{tabular}
\caption{Average alternation rate per model on opposing goals instances.}
\label{tab:air-balloon-alternation}
\end{table}

% \begin{table}[t]
% \centering
% \footnotesize
% \setlength{\tabcolsep}{4pt}
% \begin{minipage}{0.48\textwidth}
%     \centering
%     \begin{tabular}{l c}
%     \hline
%     \textbf{Model} & \textbf{Alternation Rate} \\
%     \hline
%     GPT-5 (reasoning) & \textbf{1.000} \\
%     GPT-5 Mini (reasoning) & \textbf{1.000} \\
%     GPT-5 (no reasoning) & 0.997 \\
%     GPT-5 Mini (no reasoning) & 0.989 \\
%     Claude Sonnet 4 (reasoning) & 0.987 \\
%     Claude Sonnet 4 (no reasoning) & 0.974 \\
%     GPT-OSS (reasoning) & 0.961 \\
%     Qwen3 (reasoning) & 0.883 \\
%     DeepSeek Chat v3.1 (off) & 0.820 \\
%     R1 Distill LM3 (reasoning) & 0.795 \\
%     Nemotron-Nano (no reasoning) & 0.678 \\
%     LLaMA-3.3 (no reasoning) & 0.669 \\
%     Nemotron-Nano (reasoning) & 0.437 \\
%     Qwen3 (no reasoning) & 0.190 \\
%     \hline
%     \end{tabular}
%     \caption{Average alternation rate per model on opposing goals instances.}
%     \label{tab:air-balloon-alternation}
% \end{minipage}
% \end{table}

Table \ref{tab:air-balloon-alternation} depicts the alternation rate, which measures how frequently the proposer switches between the two players from one proposal to the next. Only the two commercial GPT reasoning models (GPT-5 and GPT Mini) achieve a perfect alternation rate of $1$ across all languages and experiments. These two models are also the best on our game. The alternation rate is defined as the ratio of consecutive proposals where a different player makes the offer, ranging from 0 (one player proposes every time) to 1 (they perfectly take turns).
Let $(P_i)_{i=1}^n$ be time-ordered proposals, and let $b(i)\in\{1,2\}$ denote the proposer of $P_i$.
For $n>1$,
\[
\operatorname{alt} \;=\; \frac{1}{\,n-1\,}\sum_{i=2}^{n} \mathbf{1}\!\left\{\, b(i) \neq b(i-1) \,\right\}.
\]

For $n\le 1$, the rate is undefined.

\paragraph{Collaboration}

As explained at the beginning of Section \ref{sec:air-balloon-negotiation-dynamics}, some models make deals by agreeing on an increasing number of items throughout the negotiation. These models’ bargaining can be said to be coherent in this sense. Initially, in such cases of coherent bargaining, there will be high disagreement, which decreases until players agree on a final deal.

To capture this collaborative movement in the negotiation, we measure how much each proposal deviates from its predecessor. Specifically, we count the items added or removed between time $t-1$ and $t$ and normalise by the size of the proposal at $t-1$. This yields the normalised substitutions at time $t$. Thus, the normalised substitution count $c_t$ is the size of the symmetric difference between the proposals at $t$ and $t-1$, divided by the size of the proposal at $t-1$, where $S_t$ denotes the $t$-th proposal:

\[
c_t \;=\; \frac{\lvert S_t \triangle S_{t-1}\rvert}{\lvert S_{t-1}\rvert}, \quad t\ge 2,
\]

Figures \ref{fig:air-balloon-gpt-5-subs}–\ref{fig:air-balloon-dschatv3.1-subs} depict average normalised substitution counts over each $t$ up to $12$ for some selected models. In Figures \ref{fig:air-balloon-gpt-5-subs} and \ref{fig:air-balloon-gpt5-mini-subs}, we can again see that for our two best models (GPT-5 and GPT-5 Mini) we have higher initial disagreement with a more rapid and smooth descent when reasoning is enabled, indicating more efficient and coherent bargaining in commercial GPT models, which is amplified by enabling reasoning. DeepSeek Chat v3.1 shows the same pattern, but the initial disagreement is much lower (cf. Fig. \ref{fig:air-balloon-dschatv3.1-subs}). The GPT-OSS substitution pattern (cf. Fig. \ref{fig:air-balloon-gpt-oss-subs}) shows high volatility and hence strong willingness to change the running proposal, with high overall difference, which may tie into the earlier discussed phenomenon of stubbornness.

% \begin{table}[!ht]
% \centering
% \footnotesize
% \setlength{\tabcolsep}{1pt}
% \begin{minipage}{0.48\textwidth} % Adjusted width for longer model names
%     \centering
%     \begin{tabular}{p{0.5\textwidth} ccc}
%     \hline
%     \textbf{Model} & \textbf{Proactive} & \textbf{Reactive} & \textbf{Total Games} \\
%     \hline
%     Claude 4 (Off) & 0  & 13 & 13 \\
%     Claude 4 (On)                & 1  & 13 & 14 \\
%     DS-v3.1                      & 11 & 12 & 23 \\
%     LLaMA-70B (On)          & 1  & 0  & 1  \\
%     GPT-5 (Off)           & 15 & 5  & 21 \\
%     GPT-5 (On)                       & 17 & 3  & 24 \\
%     GPT-5 Mini (Off)      & 19 & 1  & 21 \\
%     GPT-5 Mini (On)                   & 19 & 5  & 26 \\
%     GPT-OSS                           & 5  & 6  & 21 \\
%     LM-70B (Off)               & 3  & 10 & 19 \\
%     Nem- 9B (Off)       & 21 & 1  & 22 \\
%     Nem-9B (On)                    & 0  & 2  & 2  \\
%     Qwen3-80B           & 4  & 5  & 15 \\
%     \hline
%     \end{tabular}
%     \caption{Initial reasoning style - classified by counting collaborative (proactive) versus egocentric (reactive) keywords in the first turn of each game within Strategic Reasoning tag. Calculated on all English instances.}
%     \label{tab:air-balloon-reasoning_style_analysis}
% \end{minipage}
% \end{table}
\paragraph{Counterpart Modelling}

Figure \ref{tab:air-balloon-pareto-adherence} reports the per-instance Pareto-adherence rate, i.e. the per-instance relative frequency of proposals which lie on the Pareto front. Because Pareto-efficient proposals require aligning both agents' pay-offs -- albeit not necessarily in a balanced way -- higher adherence may indicate stronger counterpart and self-modelling, i.e. the ability to infer the other player's item valuations and adjust one's own search for possible deals accordingly. Additionally, we observe that bargaining along the Pareto front is a strong indicator of overall performance in our game and almost perfectly separates the set of reasoning models from the set of non-reasoning models if we order them by Pareto-adherence.
\begin{table}[ht]
\centering
\footnotesize
\setlength{\tabcolsep}{4pt}
\begin{tabular}{l c}
\hline
\textbf{Model} & \textbf{Pareto-adherence} \\
\hline
GPT-5 (On) & 0.791 \\
GPT-5 Mini (On) & 0.732 \\
GPT-OSS (On) & 0.669 \\
Qwen3-80B (On) & 0.605 \\
Nem-9B (On) & 0.376 \\
LM-70B (On) & 0.299 \\
GPT-5 (Off) & 0.280 \\
GPT-5 Mini (Off) & 0.268 \\
Claude 4 (On) & 0.171 \\
DS-v3.1 (Off) & 0.127 \\
Nem-9B (Off) & 0.065 \\
Claude 4 (Off) & 0.026 \\
LLaMA-70B (Off) & 0.000 \\
Qwen3-80B (Off) & 0.000 \\
\hline
\end{tabular}
\caption{Average per-instance Pareto-adherence rate for proposals per model. Values computed from all played instances with 15 items.}
\label{tab:air-balloon-pareto-adherence}
\end{table}

% \begin{table}[t]
% \centering
% \footnotesize
% \setlength{\tabcolsep}{4pt}
% \begin{minipage}{0.48\textwidth} % match subfigure width
%     \centering
%     \begin{tabular}{l c}
%     \hline
%     \textbf{Model} & \textbf{Pareto Adherence} \\
%     \hline
%     GPT-5 (reasoning) & 0.791 \\
%     GPT-5 Mini (reasoning) & 0.732 \\
%     GPT-OSS (reasoning) & 0.669 \\
%     Qwen3 (reasoning) & 0.605 \\
%     Nemotron-Nano (reasoning) & 0.376 \\
%     R1 Distill LM3 (reasoning) & 0.299 \\
%     GPT-5 (no reasoning) & 0.280 \\
%     GPT-5 Mini (no reasoning) & 0.268 \\
%     Claude Sonnet 4 (reasoning) & 0.171 \\
%     DeepSeek Chat v3.1 (no reasoning) & 0.127 \\
%     Nemotron-Nano (no reasoning) & 0.065 \\
%     Claude Sonnet 4 (no reasoning) & 0.026 \\
%     LLaMA-3.3 (no reasoning) & 0.000 \\
%     Qwen3 (no reasoning) & 0.000 \\
%     \hline
%     \end{tabular}
%     \caption{Average per-instance Pareto adherence rate for proposals per model. Values computed from all played instances with 15 items.}
%     \label{tab:air-balloon-pareto-adherence}
% \end{minipage}
% \end{table}

\end{document}